\documentclass{article}

\usepackage{amsmath,amsfonts,bm}

\def\eqref#1{equation~\ref{#1}}

\def\1{\mathbbm{1}}

\DeclareMathAlphabet{\mathsfit}{\encodingdefault}{\sfdefault}{m}{sl}
\SetMathAlphabet{\mathsfit}{bold}{\encodingdefault}{\sfdefault}{bx}{n}

\newcommand{\KL}{D_{\mathrm{KL}}}

\newcommand{\JS}{D_{\mathrm{JS}}}

\usepackage{microtype}
\usepackage{graphicx}
\usepackage{subfigure}
\usepackage{booktabs} 
\usepackage{enumitem}
\usepackage{multirow}
\usepackage{rotating}
\usepackage{wrapfig}
\usepackage{chngcntr}
\usepackage{etoolbox}
\usepackage{threeparttable}
\usepackage{siunitx}
\usepackage{tabularray}
\usepackage{caption}

\makeatletter

\newcommand{\Rmnum}[1]{\expandafter@slowromancap\romannumeral #1@}
\makeatother

\usepackage{hyperref}

\usepackage[accepted]{icml2024}

\usepackage{amsmath}
\usepackage{amssymb}
\usepackage{mathtools}
\usepackage{amsthm}
\usepackage{commath}

\usepackage[capitalize]{cleveref}

\theoremstyle{plain}
\newtheorem{theorem}{Theorem}[section]

\theoremstyle{definition}
\newtheorem{definition}[theorem]{Definition}

\theoremstyle{remark}
\newtheorem{remark}[theorem]{Remark}

\usepackage[textsize=tiny]{todonotes}

\allowdisplaybreaks[4]

\icmltitlerunning{OLLIE: Imitation Learning from Offline Pretraining to Online Finetuning}

\begin{document}

\twocolumn[
\icmltitle{OLLIE: Imitation Learning from Offline Pretraining to Online Finetuning}



\icmlsetsymbol{equal}{*}

\begin{icmlauthorlist}
\icmlauthor{Sheng Yue}{thu}
\icmlauthor{Xingyuan Hua}{thu}
\icmlauthor{Ju Ren}{thu,zgc}
\icmlauthor{Sen Lin}{hu}
\icmlauthor{Junshan Zhang}{ucd}
\icmlauthor{Yaoxue Zhang}{thu,zgc}
\end{icmlauthorlist}

\icmlaffiliation{thu}{Department of Computer Science and Technology, Tsinghua University, Beijing, China}
\icmlaffiliation{zgc}{Zhongguancun Laboratory, Beijing, China}
\icmlaffiliation{hu}{Department of Computer Science, University of Houston, Texas, US}
\icmlaffiliation{ucd}{Department of Electrical and Computer Engineering, University of California, Davis, US}

\icmlcorrespondingauthor{Ju Ren}{renju@tsinghua.edu.cn}

\icmlkeywords{imitation learning, pretraining, finetuning}

\vskip 0.3in
]



\printAffiliationsAndNotice{}  

\begin{abstract}

In this paper, we study offline-to-online Imitation Learning (IL) that pretrains an imitation policy from static demonstration data, followed by fast finetuning with minimal environmental interaction. We find the na\"ive combination of existing offline IL and online IL methods tends to behave poorly in this context, because the initial discriminator (often used in online IL) operates randomly and discordantly against the policy initialization, leading to misguided policy optimization and \textit{unlearning} of pretraining knowledge. To overcome this challenge, we propose a principled offline-to-online IL method, named \texttt{OLLIE}, that simultaneously learns a near-expert policy initialization along with an \textit{aligned discriminator initialization}, which can be seamlessly integrated into online IL, achieving smooth and fast finetuning. Empirically, \texttt{OLLIE} consistently and significantly outperforms the baseline methods in \textbf{20} challenging tasks, from continuous control to vision-based domains, in terms of performance, demonstration efficiency, and convergence speed. This work may serve as a foundation for further exploration of pretraining and finetuning in the context of IL.


\end{abstract}

\section{Introduction}
\label{sec:introduction}

Imitation Learning (IL) provides methods for learning skills from demonstrations, proving particularly promising in domains such as robot control, autonomous driving, and natural language processing, where manually specifying reward functions is challenging but historical human demonstrations are readily accessible~\citep{hussein2017imitation,osa2018algorithmic}. The current IL methods can be broadly categorized into two groups: \emph{(\romannumeral1)} \emph{online IL}~\citep{ho2016generative,finn2016guided,fu2018learning} that learns imitation policies in need of continual interactions with the environment and \emph{(\romannumeral2)} \emph{offline IL}~\citep{pomerleau1988alvinn,kim2022demodice,xu2022discriminator} that extracts policies only from static demonstration data. In general, online IL exhibits superior demonstration efficiency at the cost of interactional expense and potential risk~\citep{jena2021augmenting}. Offline IL, in contrast, is more economical and safe but susceptible to error compounding owing to the covariate shift~\citep{ross2010efficient}.

These two methodological streams have been largely separated thus far, which leads to a natural question: \textit{can we combine online and offline IL to get the better parts of both worlds?} One potential solution here is to learn a reward function from offline demonstrations and subsequently use it to guide online policy optimization~\citep{chang2022mitigating,watson2023coherent,yue2023clare}. 
However, due to the intrinsic covariate shift and reward ambiguity, it is highly challenging to define and learn meaningful reward functions without environmental interaction~\citep{xu2022understanding}.\footnote{The reward ambiguity refers to the existence of a large set of reward functions under which the observed policy is optimal.} As a result, most offline reward learning methods either struggle with reward extrapolation errors~\citep{watson2023coherent} or rely on complex model-based optimization~\citep{chang2022mitigating,yue2023clare,zeng2023demonstrations,cideron2023get}. Besides, they suffer from hyperparameter sensitivity and learning instability, and are not scalable in high-dimensional environments~\citep{garg2021iq,yu2022how}. 

Inspired by the success of the \textit{pretraining and finetuning} paradigm in vision and language~\citep{brown2020language,he2022masked}, another promising way is pretraining with existing offline IL methods and finetuning using online IL, e.g., employing \texttt{GAIL}~\citep{ho2016generative} to refine the policy pretrained from \texttt{BC}~\citep{pomerleau1988alvinn}.
Unfortunately, as shown in our empirical results in \cref{sec:motivation} (also pointed out by \citet{sasaki2019sample,jena2021augmenting,orsini2021matters}), the pretrained policies can hardly help and may even degrade the performance in comparison with online IL training from scratch. We find that it is the consequence of \textit{discriminator misalignment} -- \texttt{GAIL}'s initial discriminator (acting as a local reward function) performs randomly and inconsistently against the policy initialization. It thus steers an erroneous policy optimization and induces the policy to unlearn the pretraining knowledge. 

\textbf{Contributions.} In this paper, we bridge this gap by introducing a principled offline-to-online IL method, namely \textit{OffLine-to-onLine Imitation lEarning} (\texttt{OLLIE}). It not only can provably learn a near-expert offline policy from both expert and imperfect demonstrations but can also derive the \emph{discriminator aligning with the policy with no additional computation}, enabling the pretrained policy to be fast finetuned by \texttt{GAIL} at no cost of performance degradation. Specifically, we first deduce an equivalent surrogate objective for standard IL, allowing for the utilization of imperfect/noisy data. Then, we employ convex conjugate to transform its dual problem into a convex-concave Stochastic Saddle Point (SSP) problem that can be solved with unbiased stochastic gradients in an entirely offline fashion. Importantly, this transformation enables us to extract the optimal policy simply by weighted behavior cloning while deriving the corresponding discriminator directly from the components already obtained. The policy and discriminator can jointly serve as the initialization of \texttt{GAIL}, achieving continual and smooth online finetuning. Notably, \texttt{OLLIE} circumvents intermediate reward inference and operates in a model-free manner, thereby well-suited for high-dimensional environments. In addition, thanks to the effective exploitation of suboptimal demonstrations, \texttt{OLLIE} can remain performant even with very sparse expert demonstrations.


In the experiments, we thoroughly evaluate \texttt{OLLIE} on more than \textbf{20} challenging tasks, from continuous control to vision-based domains. In offline IL, \texttt{OLLIE} achieves consistent and significant improvements over existing methods, often by \textbf{2-4x} along with faster convergence; during finetuning, it avoids the unlearning issue and achieves substantial performance improvement within a small number of interactions (often reaching the expert within \textbf{10} online episodes). 

\section{Related Work}
\label{sec:related_work}

In this section, we briefly introduce related literature due to space limitation. For a detailed discussion and comparative analysis, see \cref{sec:supp_related_work}.

\textbf{Online imitation learning.} IL has a long history, with early efforts using supervised learning to match a policy’s actions to those of the expert~\citep{sammut1992learning}. Among recent advances, \citet{ho2016generative} and the other follow-up Adversarial Imitation Learning (AIL) works~\citep{li2017infogail,fu2018learning,kostrikov2018discriminator,blonde2019sample,sasaki2019sample,wang2019random,barde2020adversarial,ghasemipour2020divergence,ke2021imitation,ni2021f,swamy2021moments,viano2022proximal,al2023ls}, which cast the problem into a game-theoretic optimization, have been proven particularly successful from low-dimensional continuous control to high-dimensional tasks like autonomous driving from pixelated input~\citep{kuefler2017imitating,zou2018understanding,ding2019goal,arora2021survey,jena2021augmenting}. However, these methods typically require substantial environmental interactions, hampering its deployment from scratch in many cost-sensitive or safety-sensitive domains. 

\textbf{Offline imitation learning.} The simplest approach to offline IL is Behavior Cloning (\texttt{BC})~\citep{pomerleau1988alvinn} that directly mimics expert behaviors using regression, whereas it inevitably suffers from error compounding, i.e., the policy is not able to recover expert behaviors when it leads to a state not observed in expert demonstrations~\citep{rajaraman2020toward}. Considerable research has been devoted to developing offline IL methods to remedy this problem, generally divided into two categories: \textit{{1)~direct policy extraction}}~\citep{jarrett2020strictly,kostrikov2020imitation,sasaki2021behavioral,garg2021iq,swamy2021moments,florence2022implicit,kim2022demodice,xu2022discriminator,li2023imitation} and \textit{{2)~offline Inverse Reinforcement Learning (IRL)}} ~\citep{reddy2019sqil,wang2019random,brantley2020disagreement,zolna2020offline,chan2021scalable,dadashi2021primal,chang2022mitigating,watson2023coherent,yue2023clare,zeng2023demonstrations}. Yet, due to the dynamics property of decision-making problems and limited state-action coverage of offline data, purely offline learning does not suffice to ensure satisfactory performance in practical and high-dimensional scenarios. 

\textbf{Offline-to-online reinforcement learning.} The recipe of pretraining and finetuning has led to great success in modern machine learning~\citep{brown2020language,he2022masked}.  Very recently, numerous efforts sought to translate such a recipe to decision-making problems, which use offline RL for initializing value functions and policies and subsequently employ online RL to improve the policy~\cite{lee2022offline,mark2022finetuning,song2022hybrid,ball2023efficient,li2023accelerating,nakamoto2023cal,wagenmaker2023leveraging,wang2023warm,wang2023train,yang2023hybrid,zhang2023policy,yue2024federated}. In light of these advances, a potential solution to offline-to-online IL could be abstracting a reward function by offline IRL and resorting to RL for tuning the policy. However, it is highly indirect, and the reward extrapolation would largely aggravate variance and bias in both offline and online IL~\citep{chang2022mitigating,watson2023coherent,yue2023clare}.

To the best of our knowledge, we are the first to study offline-to-online IL bypassing intermediate IRL processes.


\section{Preliminaries}
\label{sec:preliminaries}


\textbf{Markov Decision Process.} MDP can be specified by tuple $M\doteq\langle\mathcal{S},\mathcal{A},T,R,\mu,\gamma\rangle$, with state space $\mathcal{S}$, action space $\mathcal{A}$, transition dynamics $T:\mathcal{S}\times\mathcal{A}\rightarrow\mathcal{P}(\mathcal{S})$, reward function ${R}:\mathcal{S}\times\mathcal{A}\rightarrow\mathbb{R}$, initial state distribution $\mu\in\mathcal{P}(\mathcal{S})$, and discount factor $\gamma\in(0,1)$, where $\mathcal{P}(\mathcal{S})$ denotes the set of distributions over $\mathcal{S}$. A stationary stochastic policy maps states to distributions over actions as $\pi:\mathcal{S}\rightarrow\mathcal{P}(\mathcal{A})$. The objective of RL can be expressed as maximizing expected cumulative rewards~\citep{puterman2014markov}: 
\begin{align}
	\max_{\pi}\mathbb{E}_{(s,a)\sim\rho^{\pi}}\left[{R}(s,a)\right]
\end{align}
where $\rho^\pi$ is the normalized stationary state-action distribution (abbreviated as stationary distribution) of policy $\pi$:
\begin{align*}
	\rho^{\pi}(s,a)\doteq(1-\gamma)\sum^{\infty}_{h=0}\gamma^h\Pr(s_h=s,a_h=a\mid T,\pi,\mu).
\end{align*}



\textbf{Imitation learning.} IL is the setting where underlying reward signals are not available. Instead, it has access to an expert dataset, denoted as $\mathcal{D}_e\doteq\{(s_i,a_i,s'_i))\}^{n_e}_{i=1}$, where $(s_i,a_i,s'_i)$ is the transition collected from an unknown expert policy (often assumed to be optimal). Denote the state-action distribution in $\mathcal{D}_e$ as $\tilde\rho^e$. The IL problem is typically cast as divergence minimization between the learning policy and expert policy, $\min_\pi D(\tilde\rho^e\|\rho^\pi)$, where $D$ is a divergence measure such as $f$-divergence~\citep{ghasemipour2020divergence}.

\textbf{Online imitation learning.} Online IL is the setting where the IL algorithms are allowed to interact with the MDP. \textit{Generative Adversarial Imitation Learning} (\texttt{GAIL}) is a well-established online IL approach~\citep{ho2016generative} building on Generative Adversarial Networks (GANs)~\cite{goodfello2016generative}. \texttt{GAIL} learns a discriminator $D:\mathcal{S}\times\mathcal{A}\rightarrow(0,1)$ to recognize whether a state-action pair comes from the expert demonstrations, while a generator $\pi$ mimics the expert policy via maximizing the local rewards given by the discriminator. Specifically, the \texttt{GAIL}'s learning objective is 
\begin{align}
\label{eq:gail}
\min_\pi\max_D \mathbb{E}_{\rho^\pi}[\log D(s,a)] + \mathbb{E}_{\tilde\rho^e}[\log(1-D(s,a))].
\end{align}
Given $\pi$, the optimal discriminator can be expressed by
\begin{align}
\label{eq:learned_discriminator}
D^*(s,a)=\frac{\rho^\pi(s,a)}{\rho^\pi(s,a) + \tilde\rho^e(s,a)}.
\end{align}
In fact, \texttt{GAIL} is minimizing the Jensen-Shannon (JS) divergence between the stationary distributions of the expert and imitating policies: $\min_\pi \JS(\rho^\pi\|\tilde\rho^e)$. 

\textbf{Offline imitation learning.} Offline IL refers to the problem where IL algorithms cannot get access to the environments, and they have to extract policies only from demonstrations. \texttt{BC} is a classical and commonly used offline IL method, whose objective is maximizing the negative log-likelihood over expert data: $\max_{\pi} \mathbb{E}_{(s,a) \sim \mathcal{D}_e}[\log\pi(a|s)]$. 
Due to the limited state coverage of $\mathcal{D}_e$, the learned policy from offline IL would suffer from severe compounding errors. Accordingly, many recent methods assume an additional yet imperfect dataset to supplement the expert data, represented as $\mathcal{D}_s\doteq\{(s_j,a_j,s'_j)\}^{n_s}_{j=1}$ with $(s_j,a_j,s'_j)$ collected from a (potentially highly suboptimal) unknown behavior policy~\citep{kim2022demodice,xu2022discriminator}. We denote the union dataset of expert and supplementary data as $\mathcal{D}_o\doteq\mathcal{D}_e\cup\mathcal{D}_s$ of which the state-action distribution is represented as $\tilde\rho^o$.\footnote{If no supplementary data is provided, then $\mathcal{D}_s=\emptyset$.} Clearly, if $\tilde\rho^e(s,a)>0$, $\tilde\rho^o(s,a)>0$.

\section{Challenges}
\label{sec:motivation}

In light of the pros and cons of offline and online IL, our focus is \textit{offline-to-online IL} which pretrains a policy initialization from offline demonstration data, followed by refining this initialization with online interaction. This paradigm enables us to take benefits of both offline and online IL. On one hand, offline pretraining can empower online IL with a warm start, enabling efficient online IL at a limited interactional cost. On the other hand, the subsequent online finetuning serves to rectify the extrapolation error of the offline policy, capable of enhancing the policy's overall robustness and generalizability.

\begin{figure}[t]
    \centering	
    {\includegraphics[width=\columnwidth]{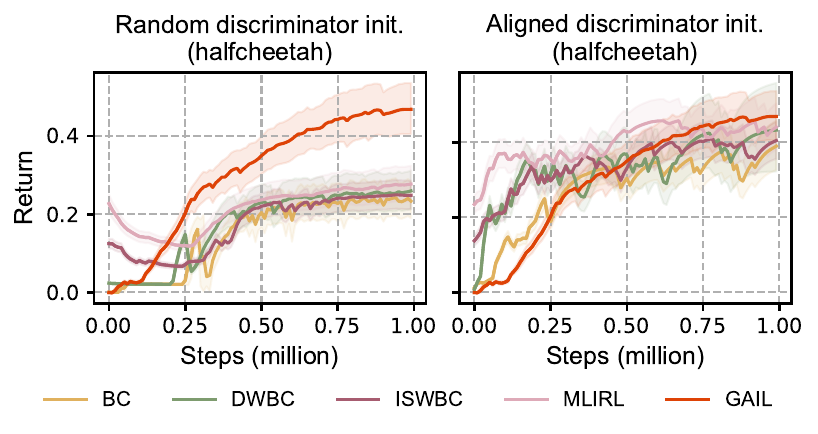}}
    \vskip -0.1in
    \caption{Effect of discriminator alignment. The curves depict the performance of \texttt{GAIL}'s online finetuning using the initial policies pretrained with different offline IL methods, except for the red curve representing running GAIL from scratch. In the left plot, the GAILs are provided with randomly initialized discriminators; in the right plot, they are given the discriminators approximately aligned with the pretrained policies.}
    \vskip -0.2in
    \label{fig:motivation_1}
\end{figure}

Yet, achieving effective offline-to-online IL is non-trivial. It has been recognized that pretraining with \texttt{BC} hardly improves the performance of online IL~\citep{sasaki2019sample,jena2021augmenting,orsini2021matters,watson2023coherent}. In fact, besides \texttt{BC}, other offline IL methods may encounter the same issue. In \cref{fig:motivation_1} (left), we depict the finetuning performance of a variety of recent offline IL methods, including \texttt{DWBC}~\citep{xu2022discriminator},  \texttt{MLIRL}~\citep{zeng2023demonstrations}, and \texttt{ISWBC}~\citep{li2023imitation} (see \cref{sec:motivating_full} for the setup and results on more tasks). Albeit with improved offline IL performance, all these methods suffer performance degradation at the initial stage of finetuning. The underlying rationale of this phenomenon is that the initial discriminator of \texttt{GAIL} operates randomly and mismatches the warm-start policy, thus steering an erroneous policy optimization and inducing the policy to unlearn previous knowledge. To substantiate this claim, we sample more than 100 trajectories by rolling out initial policies in environments, based on which we update initial discriminators sufficiently before finetuning. As shown in \cref{fig:motivation_1} (right), it remedies this issue. 

Unfortunately, this Monte Carlo method requires extensive number of sampled trajectories to estimate the policy's stationary distribution, which is prohibitively expensive. Therefore, we ask: \textit{can we obtain the aligned discriminator more efficiently without repetitive Monte Carlo sampling?} 


\section{Offline-To-Online Imitation Learning}
\label{sec:offline}

In this section, we address the question raised in \cref{sec:motivation} via introducing a principled offline IL method. It can fully utilize both expert and suboptimal data ($\mathcal{D}_e,\mathcal{D}_s$) to learn a near-expert offline policy that enjoys minimal discrepancy with the expert state-action distribution. More importantly, after obtaining the learned policy, our method can cleverly deduce the discriminator \textit{aligned with this policy} with \textit{no} additional computation or environmental interaction. Next, we begin by formally introducing the IL formulation and transforming it to an equivalent form that incorporates suboptimal data and can be optimized entirely offline.


\subsection{A Surrogate Objective for Offline IL}
\label{sec:surrogate}

The objective of IL can be formulated as minimizing the reverse KL-divergence between $\tilde\rho^e$ and the stationary distributions of $\pi$~\citep{fu2018learning,kostrikov2020imitation}:\footnote{Due to $\JS(p\|q)\le \sqrt{2\KL(p\|q)}$, Problem~(\ref{eq:objective_1}) can be seen as minimizing an upper bound of \texttt{GAIL}'s objective.}
\begin{align}
\label{eq:objective_1}
\min_\pi \KL(\rho^\pi\|\tilde\rho^e)=\mathbb{E}_{(s,a)\sim\rho^\pi}\left[\log\frac{\rho^\pi(s,a)}{\tilde\rho^e(s,a)}\right].
\end{align}
Directly dealing with Problem~(\ref{eq:objective_1}) hardly exploits supplementary data like \citet{kostrikov2020imitation}. Hence, we revert the objective to a surrogate form that incorporates $\mathcal{D}_s$:
\begin{align}
\label{eq:objective_2}
\max_\pi\mathbb{E}_{(s,a)\sim\rho^\pi}\big[ \tilde R(s,a)\big] - \KL(\rho^\pi\|\tilde\rho^o)
\end{align}
where $\tilde{R}(s,a)\doteq\log\frac{\tilde\rho^e(s,a)}{\tilde\rho^o(s,a)}$ serves as an auxiliary reward function. The equivalence of \cref{eq:objective_1,eq:objective_2} can be easily seen via adding and subtracting $\mathbb{E}_{\rho^\pi}[\log\tilde\rho^o(s,a)]$ to \cref{eq:objective_1}. Notably, the integration of $\tilde\rho^o$ here is not trivial, and later in \cref{sec:optimizing}, we will show that it enables effective utilization of \textit{dynamics information} within the supplementary data. 
\begin{remark}
    While KL-regularized problems have been studied in the fields of RL~\citep{nachum2019algaedice}, offline RL~\citep{lee2022offline}, and offline IL~\citep{kim2022demodice}, these solutions either require online interactions or suffer from biased gradient estimates, not sufficing guaranteed performance in offline IL (see \cref{sec:supp_related_work} for details). 
\end{remark}




\subsection{Auxiliary Reward Function}

Before delving into Problem~(\ref{eq:objective_2}), we first describe how to calculate the auxiliary reward function in terms of the density ratio. In the low-dimensional tabular setting, we can directly compute $\tilde\rho^e(s,a)$ and $\tilde\rho^o(s,a)$ via the corresponding state-action counts in $\mathcal{D}_e$ and $\mathcal{D}_o$. However, in high-dimensional or continuous domains, estimating the densities separately and then calculating their ratio hardly works well due to error accumulation. An alternative is to estimate the log ratio via learning a discriminator $d^*:\mathcal{S}\times\mathcal{A}\rightarrow(0,1)$:
\begin{align}
	\label{eq:dice}
	\max_d \mathbb{E}_{\tilde\rho^e}\big[\log d(s, a)\big]+\mathbb{E}_{\tilde\rho^o}\big[\log (1-d(s, a))\big]
\end{align}
where the optimal discriminator $d^*$ can recover
\begin{align}
	\label{eq:learned_reward}
	\tilde{R}(s,a) = \log \frac{\tilde\rho^e(s, a)}{\tilde\rho^o(s, a)}=\log \frac{d^*(s, a)}{1-d^*(s, a)}.
\end{align}
In particular, $d^*$ also plays an important role during online finetuning (\cref{sec:online_finetuning}).

\subsection{Optimizing the Surrogate Problem}
\label{sec:optimizing}


We proceed to derive the resolution of Problem~(\ref{eq:objective_2}) that can operate entirely offline without bias. Define the set of stationary distributions satisfying Bellman flow constraints: 
\begin{align}
	\mathcal{Z}\doteq\left\{\rho:\rho(s,a)\ge0,~f_s(\rho)=0,~\forall s\in\mathcal{S},a\in\mathcal{A}\right\}
\end{align}
where $f_s(\rho)\doteq(1-\gamma)\mu(s)+\gamma\sum_{a,s'} T(s|s',a)\rho(s',a)-\sum_a\rho(s,a)$. An elementary result has shown that there exists a one-to-one correspondence between the policy and its stationary state-action distribution: if $\rho\in\mathcal{Z}$, then $\rho$ is the occupancy for policy $\pi_\rho(a|s)\doteq \rho(s,a)/\sum_{a'}\rho(s,a')$; and $\pi_\rho$ is the only stationary policy with $\rho$~\citep[Theorem 2]{syed2008apprenticeship}. Therefore, Problem~(\ref{eq:objective_2}) can be equivalently written as the following form: 
\begin{align}
	\label{eq:objective_3}
	\max_{\rho\ge0}~&\mathbb{E}_{(s,a)\sim\rho}\big[\tilde R(s,a)\big] - \KL(\rho\|\tilde\rho^o)\\
	\label{eq:eq3}
	\mathrm{s.t.}~&f_s(\rho) = 0,~\forall s\in\mathcal{S}.
\end{align}
Since the objective and constraints are concave and affine on $\rho$ respectively, Problem~(\ref{eq:objective_3})--(\ref{eq:eq3}) is a convex optimization problem. Consider the Lagrangian of the above problem:
\begin{align*}
	L(\rho,\nu)\doteq\;&\mathbb{E}_{s,a\sim\rho}[\tilde{R}(s,a)] - \KL(\rho\|\tilde\rho^o)+\sum_s \nu(s)f_s(\rho)
\end{align*}
with $\nu$ the Lagrangian multiplier. Rearranging terms and plugging $\KL(\rho\|\tilde\rho^o) = \sum_{s,a} \rho(s,a)\log(\rho(s,a)/\tilde\rho^o(s,a))$ in the Lagrangian, we obtain
\begin{align}
	\label{eq:lagrangian_function}
	L(\rho,\nu)=\;&\sum_{s,a}\rho(s,a)\bigg(\delta_{\nu}(s,a)-\log\frac{\rho(s,a)}{\tilde\rho^o(s,a)}\bigg) \nonumber\\
	&+ (1-\gamma)\sum_{s}\nu(s)\mu(s)
\end{align}
where $\delta_{\nu}(s,a)\doteq \tilde{R}(s,a) + \gamma\sum_{s'}\nu(s')T(s'|s,a) - \nu(s)$ (informally, it can be considered as an advantage function with $\nu$ treated as the value function). There always exists $\rho$ such that $\rho(s,a)>0$ for every $s\in\mathcal{S},a\in\mathcal{A}$ when each $s\in\mathcal{S}$ is reachable under the MDP $M$. From Slater's condition, the strong duality holds. To find optimal $\rho$, we take the derivative of $L$ w.r.t. $\rho(s,a)$:
\begin{align}
	\label{eq:partial_derivative}
	\frac{\partial L}{\partial\rho(s,a)} = \delta_{\nu}(s,a)-\log\frac{\rho(s,a)}{\tilde\rho^o(s,a)}- 1.
\end{align}
Taking $\frac{\partial L}{\partial\rho(s,a)}=0$ yields
\begin{align}
	\label{eq:optimal_rho}
	\rho(s,a)=\tilde\rho^o(s,a)\exp\left(\delta_{\nu}(s,a)-1\right).
\end{align}
Substituting \cref{eq:optimal_rho} in \cref{eq:lagrangian_function}, we obtain the dual problem (slightly abusing notation $L$) as follows:
\begin{align}
	\label{eq:dp}
	\min_\nu L(\nu) \doteq\;& \mathbb{E}_{(s,a)\sim\tilde\rho^o}\left[\exp\big(\delta_{\nu}(s,a)-1\big)\right]\nonumber\\
	&+ (1-\gamma)\mathbb{E}_{s\sim\mu}\left[\nu(s)\right].
\end{align}
It can be proved that $L(\nu)$ is convex on $\nu$ (see \cref{sec:convexity}). However, it is problematic to directly optimize Problem~(\ref{eq:dp}), because the expectation in $\delta_\nu$ leads to biased stochastic gradients due to double sampling, 
and the exponential term in Problem~(\ref{eq:dp}) easily causes numerical instability in practice.



Next, we overcome this limitation via \textit{convex conjugate}.\footnote{Slightly abusing notations, we use `$*$' to mark both optimal solutions and conjugates to keep consistent with the literature, which can be recognized from the context.} 
\begin{definition}[Convex conjugate]
	For an extended real-value function $f:\mathbb{R}\rightarrow [-\infty,\infty]$, its conjugate is defined by $f^*(y) \doteq \max_{x} yx - f(x)$.
\end{definition}
From the definition, letting $f(x)=\exp(x - 1)$, its conjugate can be expressed as
\begin{align}
	f^*(y) 
	= \max_x yx - \exp(x - 1)
	=y\log y.
\end{align}
From~\citep[Proposition 7.1.1]{bertsekas2003convex}, for any closed, proper and convex function $f$, the conjugate of the conjugate of $f$ is again $f$, i.e., $f^{**} = f$. Replacing $x$ with $\delta_\nu(s,a)$, we have
\begin{align}
	\label{eq:fenchel_duality}
	\exp\left(\delta_{\nu}(s,a)-1\right) =&\max_{y(s,a)} \delta_{\nu}(s,a)y(s,a)\nonumber\\
	&- y(s,a)\log y(s,a)\big  .
\end{align}
Plugging \cref{eq:fenchel_duality} in Problem~(\ref{eq:dp}) and rearranging terms, the dual problem is equivalent to a min-max problem:
\begin{align}
	\label{eq:minimax}
	\min_\nu\max_y\;&F(\nu,y)\doteq \mathbb{E}_{(s,a)\sim\tilde\rho^o}\big[\delta_{\nu}(s,a)y(s,a) - y(s,a)\nonumber\\
	&\cdot \log y(s,a)\big]+(1-\gamma)\mathbb{E}_{s\sim\mu}[\nu(s)].
\end{align}
Since $F(\cdot,y)$ is convex with fixed $y$, and $F(\nu,\cdot)$ is concave with fixed $\nu$, the minimax theorem holds~\citep{du1995minimax}, and Problem~(\ref{eq:minimax}) is in fact a convex-concave Stochastic Saddle Point (SSP) problem. For a transition $(s,a,s')$, denote $\tilde\delta_{\nu}(s,a,s')$ as 
\begin{align}
	\tilde\delta_{\nu}(s,a,s') \doteq \tilde{R}(s,a) + \gamma\nu(s') - \nu(s).
\end{align}
Thus, we obtain the unbiased counterpart of Problem~(\ref{eq:minimax}):
\begin{align}
	\label{eq:empirical_minimax}
	\min_\nu\max_y \tilde F(\nu,y)\doteq\mathbb{E}_{(s,a,s')\sim\mathcal{D}_o}\big[\tilde\delta_{\nu}(s,a,s')y(s,a)\nonumber\\
	- y(s,a)\log y(s,a)\big]+(1-\gamma)\mathbb{E}_{s\sim\mathcal{D}_o(s_0)}[\nu(s)].
\end{align}

\begin{remark}
    \label{remark:computation}
    Problem~(\ref{eq:minimax}) is well-suited for practical computation. First, unbiased estimates of both the objective and its gradients are easy to compute using transition samples. Second, it can be shown that Problem~(\ref{eq:empirical_minimax}) enjoys a convergence rate of $O(1/\epsilon)$ in terms of the duality gap (by Nemirovski's mirror-prox algorithm)~\citep{nemirovski2004prox} and a finite-sample generalization bound of $O(1/\sqrt{n_e+n_s})$ under mild assumptions~\citep{zhang2021generalization}. More importantly, the optimum of $y$ can be directly used for offline policy extraction as well as the follow-up computation of the discriminator initialization (see \cref{sec:policy_extraction,sec:online_finetuning}). 
\end{remark}

\begin{remark}
    \label{remark:dynamics}
    How to extract useful information from imperfect data remains an open problem in offline IL. In contrast to existing works that fit a world model~\citep{yue2023clare,zeng2023demonstrations} or shift the prime objective~\citep{kim2022demodice,xu2022discriminator} (see \cref{sec:supp_related_work}), Problem~(\ref{eq:minimax}) enables effective utilization of dynamics information within $\mathcal{D}_e,\mathcal{D}_s$ in a entirely offline, model-free, and unbiased manner, achieving a correct exploitation of suboptimal data and a mitigation of the covariate shift. 
\end{remark}

\subsection{Offline Policy Extraction}
\label{sec:policy_extraction}

From \cref{eq:optimal_rho}, given optimal $\nu^*$, optimal $\rho^*$ is expressed as
\begin{align}
	\label{eq:optimal_rho_2}
	\rho^*(s,a)=\tilde\rho^o(s,a)\exp\left(\delta_{\nu^*}(s,a)-1\right).
\end{align}
Based on \citet[Theorem 2]{syed2008apprenticeship}, the corresponding policy of $\rho^*$ satisfies
\begin{align}
	\label{eq:offline_policy}
	\pi^*(a|s) = 
	\frac{\rho^*(s,a)}{\sum_{a'}\rho^*(s,a')}
	\propto\tilde\rho^o(s,a)\exp\left(\delta_{\nu^*}(s,a)-1\right).
\end{align}
From \cref{eq:lagrangian_function}, direct calculating $\delta_{\nu^*}(s,a)$ involves estimating an advantage-like function and can result in much additional computation. Fortunately, taking $\frac{\partial F}{\partial y(s,a)}=0$, we immediately obtain
\begin{align}
	\label{eq:eq2}
	y^*(s,a) = \exp(\delta_{\nu^*}(s,a)-1)
\end{align}
where $y^*$ is the optimum (saddle point) of \cref{eq:minimax}. Therefore, we can sidestep fitting $\delta_{\nu^*}(s,a)$ and calculate the optimal policy directly from 
\begin{align}
	\label{eq:imitation_policy}
	\pi^*(a|s) = \frac{\tilde\rho^o(s,a)y^*(s,a)}{z(s)}
\end{align}
where $z$ is the partition function. We provide two options to extract the policy.

\textit{1) Reverse KL-divergence.}
From \cref{eq:learned_reward}, denote $q(s,a)$ as 
\begin{align}
	\label{eq:offline_policy_2}
	q(s,a)\doteq\tilde\rho^e(s,a)y^*(s,a)\left(\frac{1}{d^*(s,a)}-1\right).
\end{align}
Similarly to \texttt{SAC}~\citep{haarnoja18soft}, we can learn the policy via solving the following reverse KL-divergence:
\begin{align}
	\label{eq:offline_policy_objective}
	\min_\pi J(\pi)=\mathbb{E}_{s\sim\mathcal{D}_o}\bigg[\KL\bigg(\pi(\cdot|s)\,\bigg\|\,\frac{q(s,\cdot)}{z(s)}\bigg)\bigg]
\end{align}
which can be optimized via reparametrization in practice. 

\textit{2) Forward KL-divergence.}
Consider the following forward KL-divergence between $\pi^*$ and the learning policy:
\begin{align}
	\label{eq:eq9}
	&\mathbb{E}_{s\sim\rho^*}\left[\KL(\pi^*(\cdot|s)\|\pi(\cdot|s))\right]\nonumber\\
	=\;&\mathbb{E}_{s\sim\rho^*}\left[\mathbb{E}_{a\sim\pi^*(\cdot|s)}\left[\log\pi^*(a|s) - \log\pi(a|s)\right]\right]\nonumber\\
	\Leftrightarrow\;& \mathbb{E}_{(s,a)\sim\rho^*}\left[- \log\pi(a|s)\right]\nonumber\\
	=\;& \mathbb{E}_{(s,a)\sim\tilde\rho^o}\left[- \frac{\rho^*(s,a)}{\tilde\rho^o(s,a)}\log\pi(a|s)\right]
\end{align}
where we omit $ \mathbb{E}_{(s,a)\sim\rho^*}[\log\pi^*(a|s)]$ due to its independency to $\pi$. From \cref{eq:optimal_rho_2,eq:eq2}, the importance weight in \cref{eq:eq9} can be computed as
\begin{align}
	\label{eq:eq10}
	\frac{\rho^*(s,a)}{\tilde\rho^o(s,a)} = \exp\left(\delta_{\nu^*}(s,a)-1\right) = y^*(s,a).
\end{align}
Substituting \cref{eq:eq10} in \cref{eq:eq9}, we can extract the policy via the following weighted behavior cloning:
\begin{align}
	\label{eq:weighted_bc}
	\max_\pi J(\pi)=\mathbb{E}_{(s,a)\sim\mathcal{D}_o}\left[y^*(s,a)\log\pi(a|s)\right]
\end{align}
If we consider $\delta_{\nu^*}$ as an advantage function, the instantiation is consistent with the offline RL method, \texttt{MARWIL}~\cite{wang2018exponentially}. In addition, a keen reader may find the form of \cref{eq:weighted_bc} bears a resemblance to \texttt{DemoDICE}~\citep{kim2022demodice} and \texttt{ISWBC}~\cite{li2023imitation}. We elaborate on their fundamental difference in \cref{sec:supp_related_work}. 


\subsection{Aligned Discriminator}
\label{sec:online_finetuning}




Now, we obtain a near-expert offline policy $\pi^*$. As mentioned earlier, effective fintuning for an offline policy necessitates a well-aligned discriminator which is often data-hungry. However, surprisingly, we find that the discriminator for $\pi^*$ can be directly deduced from the components we have already obtained! Specifically, from \cref{eq:learned_discriminator}, the discriminator for $\pi^*$ in \texttt{GAIL} can be expressed as 
\begin{align}
	D_0(s,a) \doteq \frac{\rho^*(s,a)}{\rho^*(s,a) + \tilde\rho^e(s,a)}= \left(1 + \frac{\tilde\rho^e(s,a)}{\rho^*(s,a)}\right)^{-1}
\end{align}
with $\rho^*$ defined in \cref{eq:optimal_rho}. From \cref{eq:learned_reward,eq:optimal_rho_2,eq:eq2},
\begin{align}
	\label{eq:d_init}
	D_0(s,a)&=\left(1 + \frac{\tilde\rho^e(s,a)}{\tilde\rho^o(s,a)}\cdot\frac{\tilde\rho^o(s,a)}{\rho^*(s,a)}\right)^{-1}\nonumber\\
	&=\left(1 + \frac{d^*(s, a)}{1-d^*(s, a)}\cdot\frac{1}{\exp\left(\delta_{\nu^*}(s,a)-1\right)}\right)^{-1}\nonumber\\
	&=\left(1 + \frac{d^*(s, a)}{1-d^*(s, a)}\cdot\frac{1}{y^*(s,a)}\right)^{-1}.
\end{align}
Therefore, discriminator aligned with $\pi^*$ can be elegantly derived by simply \textit{stitching} $d^*$ and $y^*$, both of which are already accessible from the pretraining phase, with no additional computation or data collection! 


\subsection{Implementation with Function Approximation}

In practical high-dimensional and continuous domain, we use function approximation for $d$, $\nu$, $y$, and $\pi$, parameterized by $\phi_d$, $\phi_\nu$, $\phi_y$, and $\theta$, respectively. We solve the minimax problem (\ref{eq:empirical_minimax}) by \textit{approximating dual descent}, which converges under convexity assumptions~\citep{boyd2004convex} and works very well in the case of nonlinear function approximators  (see \cref{sec:minimax_loss}). We employ the forward policy extraction in experiments (a comparison of forward and reverse updating is included in \cref{sec:comparison_forward_reverse}). During online finetuning, we build the initial discriminator by connecting the parameters of $\phi_d$ and $\phi_y$ learned offline:
\begin{align}
	\label{eq:discriminator_init}
	D_{\phi_y,\phi_d}(s,a)=\left(1 + \frac{\phi_d(s, a)}{1-\phi_d(s, a)}\cdot\frac{1}{\phi_y(s,a)}\right)^{-1}.
\end{align}

\begin{algorithm}[tb]
	\caption{Offline-to-online imitation learning (\texttt{OLLIE})}
	\label{alg:ollie}
	\begin{algorithmic}[1]
		\STATE Initialize parameters $\phi_d$, $\phi_\nu$, $\phi_y$, and $\theta$
		\STATE \texttt{// Offline phase}
		\STATE Estimate reward function $\tilde R$ via \cref{eq:dice,eq:learned_reward}
		\FOR{$i=1$ {\bfseries to} $n$}
		\STATE $\phi_\nu\leftarrow \phi_\nu - \eta_\nu \tilde\nabla_{\phi_\nu }\tilde F(\phi_\nu,\phi_y)$
		\STATE $\phi_y\leftarrow \phi_y + \eta_y \tilde\nabla_{\phi_y}\tilde F(\phi_\nu,\phi_y)$
		\ENDFOR
		\FOR{$i=1$ {\bfseries to} $n'$}
		\STATE $\theta\leftarrow\theta - \eta_\pi \tilde\nabla J(\pi_\theta)$
		\ENDFOR
		\STATE \texttt{// Online phase}
		\STATE Initialize discriminator $D_{\phi_y,\phi_d}(s,a)$ by \cref{eq:discriminator_init}
		\STATE Run \texttt{GAIL} to update $\pi_\theta$ and $D_{\phi_y,\phi_d}(s,a)$
	\end{algorithmic}
\end{algorithm}

Then, we use a standard implementation of \texttt{GAIL} to continue training the policy and discriminator in the environment. We term our proposed method \textit{OffLine-to-onLine Imitation lEarning} (\texttt{OLLIE}). The pseudocode is outlined in \cref{alg:ollie}, where $\tilde\nabla$ denotes the batch gradient. 

\section{Extensions}

In this section, we first extend our method to reward scaling that can be exploited to stabilize training by ensuring the auxiliary rewards with lower variance~\cite{rafailov2023direct,fu2020d4rl}. Subsequently, we generalize our method to undiscounted cases, followed by a discussion on a byproduct for offline RL. For ease of exposition, this section overloads some notations like $\delta_\nu$ when clear from the context.

\textbf{Reward scaling.} For ${\alpha}>0,\beta\ge0$, consider the reward is scaled by $\tilde R_{\alpha}(s,a)\doteq {\alpha}\tilde R(s,a) + \beta$. Using analogous derivation from \cref{eq:objective_3} to \cref{eq:minimax} -- building the Lagrangian and exploiting convex conjugate to transform the dual problem -- we can deduce the following minimax problem:
\begin{align}
	\label{eq:eq6}
	\min_\nu\max_y \;&{\alpha}\mathbb{E}_{(s,a)\sim\tilde\rho^o}\big[\delta_{\nu}(s,a)y(s,a)- {\alpha}\log({\alpha} y(s,a))\nonumber\\
	&\cdot y(s,a)\big]+ (1-\gamma)\mathbb{E}_{s\sim\mu}\left[\nu(s)\right]
\end{align}
where $\delta_{\nu}(s,a)=\tilde{R}_{\alpha}(s,a) + \gamma\sum_{s'}\nu(s')T(s'|s,a) - \nu(s)$. Similarly to Eqs.~(\ref{eq:optimal_rho})--(\ref{eq:imitation_policy}), the offline policy still follows \cref{eq:imitation_policy}. Based on \cref{eq:d_init}, the discriminator of $\pi^*$ satisfies
\begin{align}
	D_0(s,a)=\left(1 + \frac{d^*(s, a)}{1-d^*(s, a)}\cdot\frac{1}{{\alpha} y^*(s,a)}\right)^{-1}.
\end{align}
The detailed derivation can be found in \cref{sec:reward_scaling}.

\textbf{Undiscounted case.} In the undiscounted case where $\gamma=1$, the stationary distribution is expressed as
\begin{align*}
	\rho^\pi(s,a)=\lim_{H\rightarrow\infty}\frac{1}{H}\sum^{H-1}_{h=0}\Pr(s_h=s,a_h=a\mid T,\pi,\mu)
\end{align*}
which renders Problem (\ref{eq:objective_3})--(\ref{eq:eq3}) ill-posed.\footnote{If $\rho^*$ is the optimum, $a\rho^*$ is still the optimizer for any $a>0$.} This can be overcome by introducing an additional normalization constraint $\sum_{s,a}\rho(s,a)=1$ to the original problem. Represent $\lambda$ as the Lagrangian multiplier for the normalization constraint. Following the same line from \cref{eq:objective_3} to \cref{eq:dp}, the dual problem changes to
\begin{align}
	\min_{\nu,\lambda}\max_y\;&\mathbb{E}_{(s,a)\sim\tilde\rho^o}\big[\delta_{\nu}(s,a)y(s,a) + \lambda y(s,a)\nonumber\\
	&- y(s,a)\log y(s,a)\big].
\end{align}
Accordingly, the policy and discriminator match \cref{eq:imitation_policy,eq:d_init}, respectively (see \cref{sec:undiscounted} for details).


\textbf{A byproduct.} Given underlying reward signal $R(s,a)$ instead of $\tilde R(s,a)$, Problem~(\ref{eq:objective_2}) matches a formulation of offline  RL~\citep{lee2021optidice}. Thus, our method can be directly applied to offline RL. We delve into it in \cref{sec:by_product}.

\section{Experiment}
\label{sec:experiment}

In this section, we use experimental studies to evaluate our proposed method by answering the main questions: 
\begin{enumerate}[itemsep=0pt,topsep=0pt]
    \renewcommand{\labelenumi}{Q\theenumi.}
    \item How does it perform in offline IL and online finetuning compared to existing methods across various benchmarks, especially in high-dimensional environments?
    \item How is the performance affected by factors such as the number of expert/imperfect demonstrations and the quality of imperfect data?
    \item What are the effects of components like the discriminator initialization and policy extraction approaches?
\end{enumerate}
Experimental details, full results, and ablations are elaborated in \cref{sec:experiment_setting,sec:complete_experiments,sec:ablation} due to space limitation.

\subsection{Experimental Setup}

\textbf{Environments and datasets.} We run experiments with 5 domains including 21 tasks: 1)~AntMaze (\texttt{umaze}, \texttt{medium}, \texttt{large}), 2)~Adroit (\texttt{pen}, \texttt{hammer}, \texttt{door}, \texttt{relocate}), 3)~MuJoCo (\texttt{ant}, \texttt{hopper},
\texttt{halfcheetah}, \texttt{walker2d}), 4)~FrankaKitchen (\texttt{complete}, \texttt{partial}, \texttt{undirect}), 5)~vision-based Robomimic (\texttt{lift}, \texttt{can}, \texttt{square}), and 6)~vision-based MuJoCo. Specifically, the MuJoCo domain consists of 4 tasks that are popularly used to evaluate OLLIE's basic effectiveness in offline RL/IL. The Adroit benchmarks require controlling a 24-DoF robotic hand and have narrow expert data distributions, which can demonstrate OLLIE's capabilities in dealing with more difficult robot manipulation tasks. The AntMaze tasks require composing parts of suboptimal trajectories to form more optimal policies for reaching goals, thereby capable of evaluating OLLIE's capability in effectively utilizing imperfect demonstrations. Analogously, results on FrankaKitchen can demonstrate the imperfection-leveraging capability of OLLIE, but the tasks are non-navigation and more challenging than AntMaze due to the need of precise long-horizon manipulation. The image-based Robomimic and Mujoco (including 7 tasks) are employed to test OLLIE's scalability to high-dimensional environments.
%

\begin{figure}[htpb]
	\centering  \includegraphics[width=0.995\linewidth]{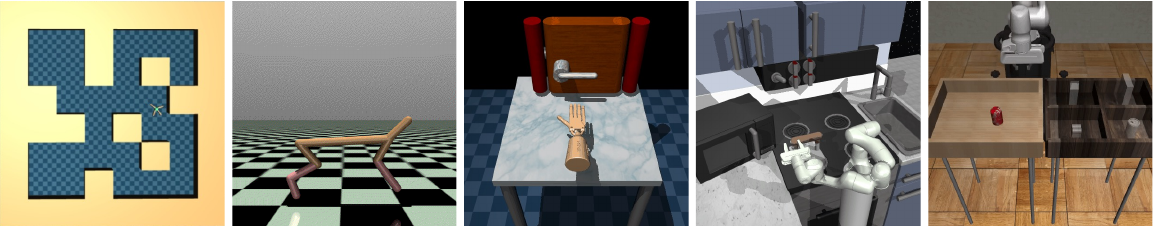}
	\vskip -0.1in
	\caption{Benchmark environments. From left to right: AntMaze, MuJoCo, Adroit, FrankaKitchen, and vision-based Robomimic. We also consider vision-based MuJoCo with image observations.}
	\label{fig:environment}
\end{figure}

\begin{table*}[t]
    \centering
    \renewcommand{\arraystretch}{1.15}
    \caption{Normalized performance in offline IL under limited expert demonstrations and low-quality imperfect data with varying qualities. Uncertainty intervals depict standard deviation over five seeds. Expert trajectories are sampled from \texttt{expert} of \texttt{D4RL} (1 trajectory for MuJoCo and 10 for Adroit), and imperfect trajectories are sampled from the datasets listed in the second column (1000 for each task). In MuJoCo, the trajectory length is less than 1000; in Adroit, it is less than 100. The third column comprises average normalized scores of imperfect data. \texttt{medium-replay} and \texttt{medium-expert} is abbreviated to \texttt{med.-rep.} and \texttt{med.-exp.}, respectively.}
    \label{tab:diverse_data_quality}
    \vskip 0.1in
    \resizebox{\textwidth}{!}{
    \begin{tabular}{l|l|r||r|r|r|r|r|r|r}
        Task                                  & Imperfect data     & Score  & \texttt{BC\ \ \ }    & \texttt{NBCU\ \,}      & \texttt{CSIL\ \,} & \texttt{DWBC\ \,}      & \texttt{MLIRL\;\,} & \texttt{ISWBC\;\,}    & \texttt{OLLIE} (ours)  \\ 
        \hline
        \multirow{4}{*}{\texttt{ant}}         & \texttt{random}    & $9.2$  & $-10.7\pm11.7$ & $31.1\pm7.0$       & $2.8\pm1.3$   & $24.5\pm1.1$       & $39.3\pm9.0$   & $30.4\pm7.7$      & $\bm{57.1\pm7.0}$      \\
        & \texttt{med.-rep.} & $19.0$ & $-10.7\pm11.7$ & $27.0\pm6.0$       & $61.0\pm3.6$  & $29.5\pm1.3$       & $57.9\pm5.1$   & $48.5\pm3.8$      & $\bm{74.2\pm6.1}$      \\
        & \texttt{medium}    & $80.3$ & $-10.7\pm11.7$ & $45.0\pm5.2$       & $78.2\pm8.4$  & $51.6\pm3.7$       & $63.4\pm2.5$   & $59.0\pm4.7$      & $\bm{84.9\pm2.0}$      \\
        & \texttt{med.-exp.} & $90.1$ & $-10.7\pm11.7$ & $70.5\pm2.0$       & $100.5\pm2.2$ & $79.2\pm8.8$       & $76.1\pm7.8$   & $94.1\pm2.7$      & $\bm{123.9\pm3.1}$     \\ 
        \hline
        \multirow{4}{*}{\texttt{halfcheetah}} & \texttt{random}    & $-0.1$ & $0.2\pm1.0$    & $2.2\pm0.1$        & $22.9\pm2.8$  & $0.0\pm0.0$        & $23.8\pm1.0$   & $13.7\pm3.0$      & $\bm{35.5\pm4.0}$      \\
        & \texttt{med.-rep.} & $7.3$  & $0.2\pm1.0$    & $36.1\pm5.1$       & $59.9\pm4.4$  & $36.0\pm7.5$       & $77.8\pm2.4$   & $\bm{87.5\pm3.0}$ & $44.8\pm4.1$           \\
        & \texttt{medium}    & $40.7$ & $0.2\pm1.0$    & $23.8\pm9.0$       & $62.6\pm9.9$  & $35.1\pm9.7$       & $64.3\pm1.9$   & $\bm{83.6\pm3.2}$ & $62.3\pm4.2$           \\
        & \texttt{med.-exp.} & $70.3$ & $0.2\pm1.0$    & $36.1\pm15.2$      & $101.6\pm1.5$ & $5.7\pm3.3$        & $89.7\pm5.5$   & $97.2\pm1.2$      & $\bm{114.3\pm2.0}$     \\ 
        \hline
        \multirow{4}{*}{\texttt{hopper}}      & \texttt{random}    & $1.2$  & $17.8\pm5.4$   & $6.6\pm3.9$        & $17.1\pm6.8$  & $\bm{77.8\pm12.7}$ & $54.7\pm16.7$  & $64.7\pm9.9$      & $71.1\pm3.5$           \\
        & \texttt{med.-rep.} & $6.8$  & $17.8\pm5.4$   & $29.1\pm4.2$       & $28.7\pm8.3$  & $78.9\pm2.0$       & $68.5\pm17.9$  & $90.3\pm3.1$      & $\bm{101.0\pm2.5}$     \\
        & \texttt{medium}    & $44.1$ & $17.8\pm5.4$   & $24.9\pm14.0$      & $49.3\pm4.3$  & $90.6\pm3.0$       & $68.0\pm14.2$  & $73.0\pm7.8$      & $\bm{98.7\pm7.0}$      \\
        & \texttt{med.-exp.} & $72.0$ & $17.8\pm5.4$   & $\bm{111.7\pm1.3}$ & $96.7\pm3.6$  & $100.8\pm2.1$      & $95.0\pm6.0$   & $97.6\pm1.6$      & $108.5\pm1.7$          \\ 
        \hline
        \multirow{4}{*}{\texttt{walker2d}}    & \texttt{random}    & $0.0$  & $4.6\pm3.9$    & $0.0\pm0.0$        & $8.1\pm5.6$   & $56.9\pm11.6$      & $49.9\pm6.7$   & $48.2\pm4.7$      & $\bm{59.8\pm8.5}$      \\
        & \texttt{med.-rep.} & $13.0$ & $4.6\pm3.9$    & $7.4\pm5.0$        & $27.6\pm4.3$  & $53.7\pm4.6$       & $67.6\pm7.2$   & $69.5\pm3.2$      & $\bm{79.0\pm2.3}$      \\
        & \texttt{medium}    & $62.0$ & $4.6\pm3.9$    & $23.9\pm2.7$       & $92.9\pm3.7$  & $67.0\pm7.8$       & $78.5\pm5.5$   & $56.0\pm8.0$      & $\bm{111.7\pm0.9}$     \\
        & \texttt{med.-exp.} & $81.0$ & $4.6\pm3.9$    & $11.6\pm8.4$       & $99.5\pm2.7$  & $92.6\pm5.4$       & $90.0\pm7.8$   & $88.0\pm3.5$      & $\bm{120.2\pm4.4}$     \\ 
        \hline
        \multirow{2}{*}{\texttt{hammer}}      & \texttt{human}     & $2.7$  & $5.7\pm6.8$    & $0.0\pm0.0$        & $16.7\pm8.9$  & $6.6\pm8.3$        & $1.0\pm0.1$    & $6.5\pm5.1$       & $\bm{46.1\pm6.5}$      \\
        & \texttt{clone}     & $0.5$  & $5.7\pm6.8$    & $0.3\pm0.1$        & $14.2\pm5.6$  & $3.3\pm2.8$        & $1.6\pm1.0$    & $2.8\pm2.3$       & $\bm{51.5\pm4.4}$      \\ 
        \hline
        \multirow{2}{*}{\texttt{pen}}         & \texttt{human}     & $2.1$  & $40.3\pm10.3$  & $3.2\pm9.8$        & $39.0\pm5.6$  & $42.8\pm19.3$      & $18.7\pm2.6$   & $49.5\pm2.9$      & $\bm{67.4\pm5.6}$      \\
        & \texttt{clone}     & $59.9$ & $40.3\pm10.3$  & $1.0\pm1.0$        & $45.6\pm4.7$  & $48.0\pm3.9$       & $23.4\pm1.5$   & $51.3\pm6.9$      & $\bm{68.0\pm1.6}$      \\ 
        \hline
        \multirow{2}{*}{\texttt{door}}        & \texttt{human}     & $2.6$  & $2.8\pm3.9$    & $0.0\pm0.0$        & $24.8\pm3.6$  & $0.0\pm0.0$        & $1.0\pm0.8$    & $2.9\pm2.1$       & $\bm{28.9\pm2.9}$      \\
        & \texttt{clone}     & $-0.1$ & $2.8\pm3.9$    & $0.0\pm0.0$        & $21.6\pm2.1$  & $1.0\pm1.0$        & $1.0\pm1.0$    & $1.0\pm1.0$       & $\bm{31.9\pm4.9}$      \\ 
        \hline
        \multirow{2}{*}{\texttt{relocate}}    & \texttt{human}     & $2.3$  & $0.0\pm0.0$    & $0.0\pm0.0$        & $7.7\pm8.8$   & $0.0\pm0.0$        & $1.0\pm0.2$    & $0.0\pm0.0$       & $\bm{31.2\pm5.8}$      \\
        & \texttt{clone}     & $-0.1$ & $0.0\pm0.0$    & $0.0\pm1.0$        & $10.2\pm4.4$  & $1.0\pm1.0$        & $1.0\pm1.0$    & $1.0\pm1.0$       & $\bm{40.8\pm6.9}$      \\
        \hline
        \end{tabular}}
\end{table*}

During offline training, we use the \texttt{D4RL} datasets~\citep{fu2020d4rl} for AntMaze, MuJoCo, Adroit, and FrankaKitchen and use the \texttt{robomimic}~\citep{robomimic2021} datasets for vision-based Robomimic. We construct vision-based MuJoCo datasets using the same method introduced in \citet{fu2020d4rl}. Details on environments and datasets can be found in \cref{sec:environments,sec:datasets}.

\textbf{Baselines.} We evaluate our method against four strong offline IL methods, \texttt{DWBC}~\citep{xu2022discriminator}, \texttt{ISWBC}~\citep{li2023imitation}, \texttt{MLIRL}~\citep{zeng2023demonstrations}, and \texttt{CSIL}~\citep{watson2023coherent}, all of which can utilize imperfect demonstrations (see \cref{sec:supp_related_work,sec:baselines} for more information). We also compare our method with \texttt{BC} and its counterpart with union offline data, termed \texttt{NBCU}~\citep{li2023imitation}. 


\textbf{Reproducibility.} All details of our experiments are provided in the appendices in terms of the tasks, network architectures, hyperparameters, etc. We implement all baselines and environments based on open-source repositories. The code is available at \href{https://github.com/HansenHua/OLLIE-offline-to-online-imitation-learning}{https://github.com/HansenHua/OLLIE-offline-to-online-imitation-learning}. Of note, our method is robust in hyperparameters, identical for all tasks except for the change of neural nets to CNNs in vision-based domains. 

\subsection{Experimental Results}

\subsubsection{Performance in Offline IL}

\begin{figure}[ht]
    \centering
    {\includegraphics[width=\columnwidth]{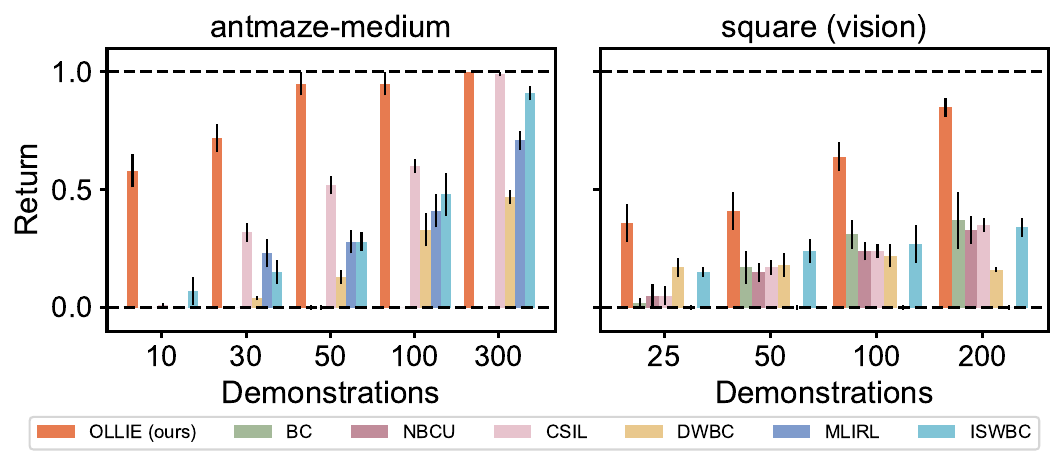}}
    \vskip -0.1in
    \caption{Comparative performance in offline IL with varying expert demonstrations. The scores are normalized and averaged over 5 seeds with standard deviation depicted by uncertainty intervals. }
    \label{fig:offline_selected}
\end{figure}

We evaluate \texttt{OLLIE}'s performance in offline IL across all benchmark environments, with 1000 sampled imperfect trajectories and varying numbers of expert trajectories (ranging from 1 to 30 in AntMaze and MuJuCo, from 10 to 300 in Adroit and FrankaKitchen, and from 25 to 200 in vision-based MuJoCo and Robomimic). The employed datasets can be found in \cref{table:dataset}. We present two selected results in \cref{fig:offline_selected} and provide full results in \cref{sec:demonstration_effeciency}. \texttt{OLLIE} consistently and significantly outperforms existing methods in terms of performance, convergence speed, and demonstration efficiency, especially in challenging robotic manipulation and vision-based domains. For example, with limited expert data, \texttt{OLLIE} outperforms baseline methods often by \textbf{2-4x} and converges often within \textbf{0.2m} steps (\cref{fig:performance_mujoco,fig:curve_mujoco,fig:performance_antmaze,fig:curve_antmaze,fig:performance_adroit,fig:curve_adroit,fig:performance_kitchen,fig:curve_kitchen,fig:performance_mujoco_image,fig:curve_mujoco_image,fig:performance_robomimic,fig:curve_robomimic}).  

We also run experiments with different qualities of imperfect data. As showcased in \cref{tab:diverse_data_quality}, \texttt{OLLIE} surpasses the baselines in \textbf{20/24} settings by wide margins, corroborating its robustness and superiority in offline IL. Learning curves are provided in \cref{sec:data_quality}.

\subsubsection{Performance in Online Fintuning}

\begin{figure}[hb]
    \centering
    {\includegraphics[width=\columnwidth]{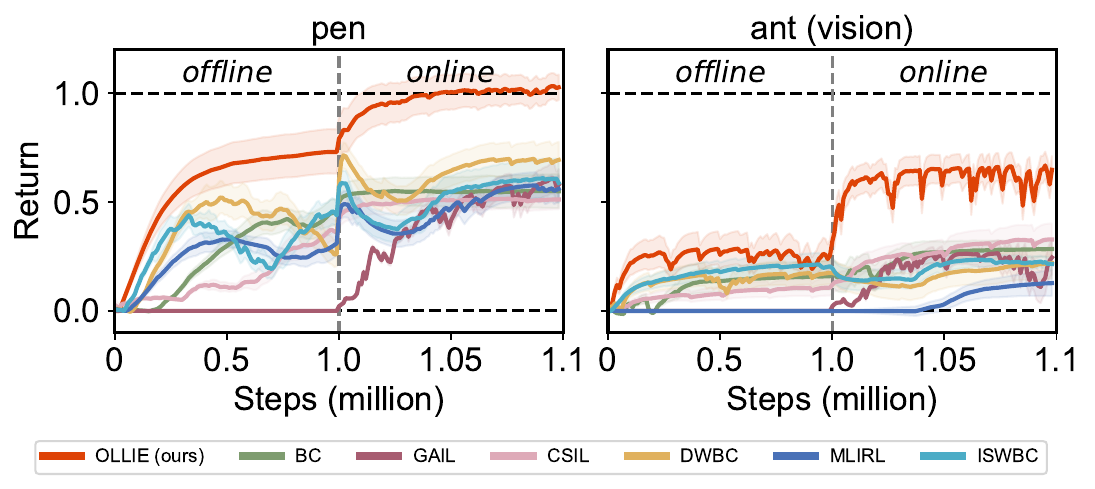}}
    \vskip -0.1in
    \caption{End-to-end performance from offline pretraining to online finetuning. The results are normalized and averaged over 5 seeds with standard deviation depicted by uncertainty intervals.}
    \label{fig:online_selected}
\end{figure}

After obtaining pretrained policies, we examine the finetuning performance under different quantities of expert demonstrations. \cref{fig:online_selected} showcases two selected results, and \cref{sec:performance_online} includes comprehensive results for all tasks. We find \texttt{OLLIE} not only avoids the unlearning phenomenon but also hastens online training. For example, as demonstrated in \cref{fig:online_selected} (right), \texttt{OLLIE} improves the offline performance by \textbf{2x} with fewer than \textbf{10} environmental episodes. Importantly, even in challenging tasks like vision-based \texttt{can} and \texttt{square} where \texttt{GAIL} from scratch fails (\cref{fig:online_curve_robomimic}), \texttt{OLLIE} performs well. This underscores the significance of effective pretraining in the context of IL.

\section{Limitation and Discussion}

In this paper, we study offline-to-online IL that pretrains a good policy initialization, followed by online finetuning with minimal interaction. First, we derive a surrogate objective for standard IL, of which the dual problem is convex. Then, we employ the convex conjugate to transform the dual problem into a convex-concave SSP that can be optimized with unbiased stochastic gradients in an entirely offline fashion. Importantly, the transformation enables us to directly extract the optimal policy using weighted behavior cloning and deduce its corresponding discriminator which can be seamlessly used in \texttt{GAIL} to achieve continual learning. 
We believe our method can benefit many real-world domains including robotics, autonomous driving, and foundation model training where designing a reward function is challenging, while environmental interaction is necessary yet costly. 


A limitation of \texttt{OLLIE} is the \texttt{GAIL}-based finetuning. The on-policy and adversarial nature of \texttt{GAIL} may lead to sample inefficiency and training instability in some scenarios. An avenue for future work is to render \texttt{OLLIE} compatible with non-adversarial or off-policy IL methods. In addition, \texttt{OLLIE}'s connection with offline RL suggests there is scope to bridge offline IL and offline RL. Another future direction is to explore the best utilization of unlabeled data in offline RL with the aid of IL.

\section*{Acknowledgments}
This research was supported in part by the National Natural Science Foundation of China under Grant No. 62302260, 62341201, 62122095, 62072472 and 62172445, by the National Key R\&D Program of China under Grant No. 2022YFF0604502, by China Postdoctoral Science Foundation under Grant No. 2023M731956, and by a grant from the Guoqiang Institute, Tsinghua University.

\section*{Impact Statement}

This paper presents work whose goal is to advance the field of Reinforcement Learning. There are many potential societal consequences of our work, none which we feel must be specifically highlighted here.

\bibliography{references}

\begin{thebibliography}{101}
\providecommand{\natexlab}[1]{#1}
\providecommand{\url}[1]{\texttt{#1}}
\expandafter\ifx\csname urlstyle\endcsname\relax
  \providecommand{\doi}[1]{doi: #1}\else
  \providecommand{\doi}{doi: \begingroup \urlstyle{rm}\Url}\fi

\bibitem[Abbeel \& Ng(2004)Abbeel and Ng]{abbeel2004apprenticeship}
Abbeel, P. and Ng, A.~Y.
\newblock Apprenticeship learning via inverse reinforcement learning.
\newblock In \emph{Proceedings of the 21st International Conference on Machine Learning}, pp.\ ~1. ACM, 2004.

\bibitem[Al-Hafez et~al.(2023)Al-Hafez, Tateo, Arenz, Zhao, and Peters]{al2023ls}
Al-Hafez, F., Tateo, D., Arenz, O., Zhao, G., and Peters, J.
\newblock {LS-IQ}: Implicit reward regularization for inverse reinforcement learning.
\newblock In \emph{International Conference on Learning Representations}, 2023.

\bibitem[Arora \& Doshi(2021)Arora and Doshi]{arora2021survey}
Arora, S. and Doshi, P.
\newblock A survey of inverse reinforcement learning: Challenges, methods and progress.
\newblock \emph{Artificial Intelligence}, 297:\penalty0 103500, 2021.

\bibitem[Ball et~al.(2023)Ball, Smith, Kostrikov, and Levine]{ball2023efficient}
Ball, P.~J., Smith, L., Kostrikov, I., and Levine, S.
\newblock Efficient online reinforcement learning with offline data.
\newblock In \emph{Proceedings of the 40th International Conference on Machine Learning}, volume 202, pp.\  1577--1594. PMLR, 2023.

\bibitem[Barde et~al.(2020)Barde, Roy, Jeon, Pineau, Pal, and Nowrouzezahrai]{barde2020adversarial}
Barde, P., Roy, J., Jeon, W., Pineau, J., Pal, C., and Nowrouzezahrai, D.
\newblock Adversarial soft advantage fitting: Imitation learning without policy optimization.
\newblock In \emph{Advances in Neural Information Processing Systems}, volume~33, pp.\  12334--12344. Curran Associates, 2020.

\bibitem[Bertsekas et~al.(2003)Bertsekas, Nedic, and Ozdaglar]{bertsekas2003convex}
Bertsekas, D., Nedic, A., and Ozdaglar, A.
\newblock \emph{Convex analysis and optimization}, volume~1.
\newblock Athena Scientific, 2003.

\bibitem[Blond\'{e} \& Kalousis(2019)Blond\'{e} and Kalousis]{blonde2019sample}
Blond\'{e}, L. and Kalousis, A.
\newblock Sample-efficient imitation learning via generative adversarial nets.
\newblock In \emph{Proceedings of the 22nd International Conference on Artificial Intelligence and Statistics}, volume~89, pp.\  3138--3148. PMLR, 2019.

\bibitem[Boyd \& Vandenberghe(2004)Boyd and Vandenberghe]{boyd2004convex}
Boyd, S.~P. and Vandenberghe, L.
\newblock \emph{Convex optimization}.
\newblock Cambridge university press, 2004.

\bibitem[Brantley et~al.(2020)Brantley, Sun, and Henaff]{brantley2020disagreement}
Brantley, K., Sun, W., and Henaff, M.
\newblock Disagreement-regularized imitation learning.
\newblock In \emph{International Conference on Learning Representations}, 2020.

\bibitem[Brown et~al.(2020)Brown, Mann, Ryder, Subbiah, Kaplan, Dhariwal, Neelakantan, Shyam, Sastry, Askell, et~al.]{brown2020language}
Brown, T., Mann, B., Ryder, N., Subbiah, M., Kaplan, J.~D., Dhariwal, P., Neelakantan, A., Shyam, P., Sastry, G., Askell, A., et~al.
\newblock Language models are few-shot learners.
\newblock In \emph{Proceedings of Advances in Neural Information Processing Systems}, pp.\  1877--1901, 2020.

\bibitem[Chan \& van~der Schaar(2021)Chan and van~der Schaar]{chan2021scalable}
Chan, A.~J. and van~der Schaar, M.
\newblock Scalable bayesian inverse reinforcement learning.
\newblock In \emph{International Conference on Learning Representations}, 2021.

\bibitem[Chang et~al.(2021)Chang, Uehara, Sreenivas, Kidambi, and Sun]{chang2022mitigating}
Chang, J., Uehara, M., Sreenivas, D., Kidambi, R., and Sun, W.
\newblock Mitigating covariate shift in imitation learning via offline data with partial coverage.
\newblock In \emph{Advances in Neural Information Processing Systems}, volume~34, pp.\  965--979. Curran Associates, 2021.

\bibitem[Chen et~al.(2019)Chen, Wang, Liu, Yang, Li, Wang, and Zhao]{chen2019computation}
Chen, M., Wang, Y., Liu, T., Yang, Z., Li, X., Wang, Z., and Zhao, T.
\newblock On computation and generalization of generative adversarial imitation learning.
\newblock In \emph{International Conference on Learning Representations}, 2019.

\bibitem[Choi \& Kim(2013)Choi and Kim]{choi2013bayesian}
Choi, J. and Kim, K.-E.
\newblock Bayesian nonparametric feature construction for inverse reinforcement learning.
\newblock In \emph{Proceedings of the 23rd International Joint Conference on Artificial Intelligence}, pp.\  1287–1293. AAAI Press, 2013.

\bibitem[Cideron et~al.(2023)Cideron, Tabanpour, Curi, Girgin, Hussenot, Dulac-Arnold, Geist, Pietquin, and Dadashi]{cideron2023get}
Cideron, G., Tabanpour, B., Curi, S., Girgin, S., Hussenot, L., Dulac-Arnold, G., Geist, M., Pietquin, O., and Dadashi, R.
\newblock Get back here: Robust imitation by return-to-distribution planning.
\newblock \emph{arXiv preprint arXiv:2305.01400}, 2023.

\bibitem[Dadashi et~al.(2021)Dadashi, Hussenot, Geist, and Pietquin]{dadashi2021primal}
Dadashi, R., Hussenot, L., Geist, M., and Pietquin, O.
\newblock Primal wasserstein imitation learning.
\newblock In \emph{International Conference on Learning Representations}, 2021.

\bibitem[Ding et~al.(2019)Ding, Florensa, Abbeel, and Phielipp]{ding2019goal}
Ding, Y., Florensa, C., Abbeel, P., and Phielipp, M.
\newblock Goal-conditioned imitation learning.
\newblock In \emph{Advances in Neural Information Processing Systems}, volume~32, pp.\  15324--15335. Curran Associates, 2019.

\bibitem[Du \& Pardalos(1995)Du and Pardalos]{du1995minimax}
Du, D.-Z. and Pardalos, P.~M.
\newblock \emph{Minimax and applications}, volume~4.
\newblock Springer Science \& Business Media, 1995.

\bibitem[Finn et~al.(2016)Finn, Levine, and Abbeel]{finn2016guided}
Finn, C., Levine, S., and Abbeel, P.
\newblock Guided cost learning: Deep inverse optimal control via policy optimization.
\newblock In \emph{Proceedings of the 33rd International Conference on Machine Learning}, volume~48, pp.\  49--58. PMLR, 2016.

\bibitem[Florence et~al.(2022)Florence, Lynch, Zeng, Ramirez, Wahid, Downs, Wong, Lee, Mordatch, and Tompson]{florence2022implicit}
Florence, P., Lynch, C., Zeng, A., Ramirez, O.~A., Wahid, A., Downs, L., Wong, A., Lee, J., Mordatch, I., and Tompson, J.
\newblock Implicit behavioral cloning.
\newblock In \emph{Proceedings of the 5th Conference on Robot Learning}, volume 164, pp.\  158--168. PMLR, 2022.

\bibitem[Fu et~al.(2018)Fu, Luo, and Levine]{fu2018learning}
Fu, J., Luo, K., and Levine, S.
\newblock Learning robust rewards with adverserial inverse reinforcement learning.
\newblock In \emph{International Conference on Learning Representations}, 2018.

\bibitem[Fu et~al.(2020)Fu, Kumar, Nachum, Tucker, and Levine]{fu2020d4rl}
Fu, J., Kumar, A., Nachum, O., Tucker, G., and Levine, S.
\newblock {D4RL}: Datasets for deep data-driven reinforcement learning.
\newblock \emph{arXiv preprint arXiv:2004.07219}, 2020.

\bibitem[Fujimoto \& Gu(2021)Fujimoto and Gu]{fujimoto2021minimalist}
Fujimoto, S. and Gu, S.~S.
\newblock A minimalist approach to offline reinforcement learning.
\newblock In \emph{Advances in Neural Information Processing Systems}, volume~34, pp.\  20132--20145. Curran Associates, 2021.

\bibitem[Garg et~al.(2021)Garg, Chakraborty, Cundy, Song, and Ermon]{garg2021iq}
Garg, D., Chakraborty, S., Cundy, C., Song, J., and Ermon, S.
\newblock {IQ-Learn}: Inverse soft-q learning for imitation.
\newblock In \emph{Advances in Neural Information Processing Systems}, volume~34, pp.\  4028--4039. Curran Associates, 2021.

\bibitem[Ghasemipour et~al.(2020)Ghasemipour, Zemel, and Gu]{ghasemipour2020divergence}
Ghasemipour, S. K.~S., Zemel, R., and Gu, S.
\newblock A divergence minimization perspective on imitation learning methods.
\newblock In \emph{Proceedings of the 3rd Conference on Robot Learning}, volume 100, pp.\  1259--1277. PMLR, 2020.

\bibitem[Goodfellow et~al.(2014)Goodfellow, Pouget-Abadie, Mirza, Xu, Warde-Farley, Ozair, Courville, and Bengio]{goodfello2016generative}
Goodfellow, I., Pouget-Abadie, J., Mirza, M., Xu, B., Warde-Farley, D., Ozair, S., Courville, A., and Bengio, Y.
\newblock Generative adversarial nets.
\newblock In \emph{Advances in Neural Information Processing Systems}, volume~27, pp.\  2672--2680. Curran Associates, 2014.

\bibitem[Guan et~al.(2021)Guan, Xu, and Liang]{guan2021generative}
Guan, Z., Xu, T., and Liang, Y.
\newblock When will generative adversarial imitation learning algorithms attain global convergence.
\newblock In \emph{Proceedings of The 24th International Conference on Artificial Intelligence and Statistics}, volume 130, pp.\  1117--1125. PMLR, 2021.

\bibitem[Gupta et~al.(2019)Gupta, Kumar, Lynch, Levine, and Hausman]{gupta2019relay}
Gupta, A., Kumar, V., Lynch, C., Levine, S., and Hausman, K.
\newblock Relay policy learning: Solving long-horizon tasks via imitation and reinforcement learning.
\newblock \emph{arXiv preprint arXiv:1910.11956}, 2019.

\bibitem[Haarnoja et~al.(2018)Haarnoja, Zhou, Abbeel, and Levine]{haarnoja18soft}
Haarnoja, T., Zhou, A., Abbeel, P., and Levine, S.
\newblock Soft actor-critic: Off-policy maximum entropy deep reinforcement learning with a stochastic actor.
\newblock In \emph{Proceedings of the 35th International Conference on Machine Learning}, volume~80, pp.\  1861--1870. PMLR, 2018.

\bibitem[He et~al.(2019)He, Girshick, and Doll{\'a}r]{he2019rethinking}
He, K., Girshick, R., and Doll{\'a}r, P.
\newblock Rethinking {ImageNet} pre-training.
\newblock In \emph{Proceedings of the IEEE/CVF International Conference on Computer Vision}, pp.\  4918--4927, 2019.

\bibitem[He et~al.(2022)He, Chen, Xie, Li, Doll{\'a}r, and Girshick]{he2022masked}
He, K., Chen, X., Xie, S., Li, Y., Doll{\'a}r, P., and Girshick, R.
\newblock Masked autoencoders are scalable vision learners.
\newblock In \emph{Proceedings of the IEEE/CVF conference on computer vision and pattern recognition}, pp.\  16000--16009, 2022.

\bibitem[Ho \& Ermon(2016)Ho and Ermon]{ho2016generative}
Ho, J. and Ermon, S.
\newblock Generative adversarial imitation learning.
\newblock In \emph{Advances in Neural Information Processing Systems}, volume~29, pp.\  4572--4580. Curran Associates, 2016.

\bibitem[Hussein et~al.(2017)Hussein, Gaber, Elyan, and Jayne]{hussein2017imitation}
Hussein, A., Gaber, M.~M., Elyan, E., and Jayne, C.
\newblock Imitation learning: A survey of learning methods.
\newblock \emph{ACM Computing Survey}, 50\penalty0 (2):\penalty0 1--35, 2017.

\bibitem[Jarrett et~al.(2020)Jarrett, Bica, and van~der Schaar]{jarrett2020strictly}
Jarrett, D., Bica, I., and van~der Schaar, M.
\newblock Strictly batch imitation learning by energy-based distribution matching.
\newblock In \emph{Advances in Neural Information Processing Systems}, volume~33, pp.\  7354--7365. Curran Associates, 2020.

\bibitem[Jena et~al.(2021)Jena, Liu, and Sycara]{jena2021augmenting}
Jena, R., Liu, C., and Sycara, K.
\newblock Augmenting {GAIL} with {BC} for sample efficient imitation learning.
\newblock In \emph{Proceedings of the 4th Conference on Robot Learning}, pp.\  80--90. PMLR, 2021.

\bibitem[Ke et~al.(2021)Ke, Choudhury, Barnes, Sun, Lee, and Srinivasa]{ke2021imitation}
Ke, L., Choudhury, S., Barnes, M., Sun, W., Lee, G., and Srinivasa, S.
\newblock Imitation learning as $f$-divergence minimization.
\newblock In \emph{Algorithmic Foundations of Robotics XIV}, pp.\  313--329. Springer, 2021.

\bibitem[Kim et~al.(2022{\natexlab{a}})Kim, Lee, Jang, Yang, and Kim]{kim2022lobsdice}
Kim, G.-H., Lee, J., Jang, Y., Yang, H., and Kim, K.-E.
\newblock {LobsDICE}: Offline learning from observation via stationary distribution correction estimation.
\newblock In \emph{Advances in Neural Information Processing Systems}, volume~35, pp.\  8252--8264. Curran Associates, 2022{\natexlab{a}}.

\bibitem[Kim et~al.(2022{\natexlab{b}})Kim, Seo, Lee, Jeon, Hwang, Yang, and Kim]{kim2022demodice}
Kim, G.-H., Seo, S., Lee, J., Jeon, W., Hwang, H., Yang, H., and Kim, K.-E.
\newblock {DemoDICE}: Offline imitation learning with supplementary imperfect demonstrations.
\newblock In \emph{International Conference on Learning Representations}, 2022{\natexlab{b}}.

\bibitem[Kostrikov et~al.(2018)Kostrikov, Agrawal, Dwibedi, Levine, and Tompson]{kostrikov2018discriminator}
Kostrikov, I., Agrawal, K.~K., Dwibedi, D., Levine, S., and Tompson, J.
\newblock Discriminator-actor-critic: Addressing sample inefficiency and reward bias in adversarial imitation learning.
\newblock In \emph{International Conference on Learning Representations}, 2018.

\bibitem[Kostrikov et~al.(2020)Kostrikov, Nachum, and Tompson]{kostrikov2020imitation}
Kostrikov, I., Nachum, O., and Tompson, J.
\newblock Imitation learning via off-policy distribution matching.
\newblock In \emph{International Conference on Learning Representations}, 2020.

\bibitem[Kuefler et~al.(2017)Kuefler, Morton, Wheeler, and Kochenderfer]{kuefler2017imitating}
Kuefler, A., Morton, J., Wheeler, T., and Kochenderfer, M.
\newblock Imitating driver behavior with generative adversarial networks.
\newblock In \emph{Proceedings of the 28th IEEE Intelligent Vehicles Symposium}, pp.\  204--211. IEEE, 2017.

\bibitem[Kumar et~al.(2020)Kumar, Zhou, Tucker, and Levine]{kumar2020conservative}
Kumar, A., Zhou, A., Tucker, G., and Levine, S.
\newblock Conservative q-learning for offline reinforcement learning.
\newblock In \emph{Advances in Neural Information Processing Systems}, volume~33, pp.\  1179--1191. Curran Associates, 2020.

\bibitem[Lee et~al.(2019)Lee, Srinivasan, and Doshi-Velez]{lee2019truly}
Lee, D., Srinivasan, S., and Doshi-Velez, F.
\newblock Truly batch apprenticeship learning with deep successor features.
\newblock In \emph{Proceedings of the 28th International Joint Conference on Artificial Intelligence}, pp.\  5909--5915, 2019.

\bibitem[Lee et~al.(2021)Lee, Jeon, Lee, Pineau, and Kim]{lee2021optidice}
Lee, J., Jeon, W., Lee, B., Pineau, J., and Kim, K.-E.
\newblock {OptiDICE}: Offline policy optimization via stationary distribution correction estimation.
\newblock In \emph{Proceedings of the 38th International Conference on Machine Learning}, volume 139, pp.\  6120--6130. PMLR, 2021.

\bibitem[Lee et~al.(2022)Lee, Seo, Lee, Abbeel, and Shin]{lee2022offline}
Lee, S., Seo, Y., Lee, K., Abbeel, P., and Shin, J.
\newblock Offline-to-online reinforcement learning via balanced replay and pessimistic {Q}-ensemble.
\newblock In \emph{Proceedings of the 5th Conference on Robot Learning}, volume 164, pp.\  1702--1712. PMLR, 2022.

\bibitem[Levine et~al.(2010)Levine, Popovic, and Koltun]{levine2010feature}
Levine, S., Popovic, Z., and Koltun, V.
\newblock Feature construction for inverse reinforcement learning.
\newblock In \emph{Advances in Neural Information Processing Systems}, volume~23, pp.\  1342--1350. Curran Associates, 2010.

\bibitem[Levine et~al.(2011)Levine, Popovic, and Koltun]{levine2011nonlinear}
Levine, S., Popovic, Z., and Koltun, V.
\newblock Nonlinear inverse reinforcement learning with {Gaussian} processes.
\newblock In \emph{Advances in Neural Information Processing Systems}, volume~24, pp.\  19--27. Curran Associates, 2011.

\bibitem[Li et~al.(2023{\natexlab{a}})Li, Zhang, Ghosh, Zhang, and Levine]{li2023accelerating}
Li, Q., Zhang, J., Ghosh, D., Zhang, A., and Levine, S.
\newblock Accelerating exploration with unlabeled prior data.
\newblock In \emph{The 37th Conference on Neural Information Processing Systems}, 2023{\natexlab{a}}.

\bibitem[Li et~al.(2017)Li, Song, and Ermon]{li2017infogail}
Li, Y., Song, J., and Ermon, S.
\newblock {InfoGAIL}: Interpretable imitation learning from visual demonstrations.
\newblock In \emph{Advances in Neural Information Processing Systems}, volume~30, pp.\  3815--3825. Curran Associates, 2017.

\bibitem[Li et~al.(2023{\natexlab{b}})Li, Xu, Qin, Yu, and Luo]{li2023imitation}
Li, Z., Xu, T., Qin, Z., Yu, Y., and Luo, Z.-Q.
\newblock Imitation learning from imperfection: Theoretical justifications and algorithms.
\newblock In \emph{The 37th Conference on Neural Information Processing Systems}, 2023{\natexlab{b}}.

\bibitem[Ma et~al.(2022)Ma, Shen, Jayaraman, and Bastani]{ma2022versatile}
Ma, Y., Shen, A., Jayaraman, D., and Bastani, O.
\newblock Versatile offline imitation from observations and examples via regularized state-occupancy matching.
\newblock In \emph{Proceedings of the 39th International Conference on Machine Learning}, volume 162, pp.\  14639--14663. PMLR, 2022.

\bibitem[Mandlekar et~al.(2022)Mandlekar, Xu, Wong, Nasiriany, Wang, Kulkarni, Fei-Fei, Savarese, Zhu, and Mart\'in-Mart\'in]{robomimic2021}
Mandlekar, A., Xu, D., Wong, J., Nasiriany, S., Wang, C., Kulkarni, R., Fei-Fei, L., Savarese, S., Zhu, Y., and Mart\'in-Mart\'in, R.
\newblock What matters in learning from offline human demonstrations for robot manipulation.
\newblock In \emph{Proceedings of the 5th Conference on Robot Learning}, volume 164, pp.\  1678--1690. PMLR, 2022.

\bibitem[Mark et~al.(2022)Mark, Ghadirzadeh, Chen, and Finn]{mark2022finetuning}
Mark, M.~S., Ghadirzadeh, A., Chen, X., and Finn, C.
\newblock Fine-tuning offline policies with optimistic action selection.
\newblock In \emph{NeurIPS Workshop on Deep Reinforcement Learning}, 2022.

\bibitem[Nachum et~al.(2019{\natexlab{a}})Nachum, Chow, Dai, and Li]{nachum2019dualdice}
Nachum, O., Chow, Y., Dai, B., and Li, L.
\newblock {DualDICE}: Behavior-agnostic estimation of discounted stationary distribution corrections.
\newblock In \emph{Advances in Neural Information Processing Systems}, volume~32, pp.\  2318--2328. Curran Associates, 2019{\natexlab{a}}.

\bibitem[Nachum et~al.(2019{\natexlab{b}})Nachum, Dai, Kostrikov, Chow, Li, and Schuurmans]{nachum2019algaedice}
Nachum, O., Dai, B., Kostrikov, I., Chow, Y., Li, L., and Schuurmans, D.
\newblock {AlgaeDICE}: Policy gradient from arbitrary experience.
\newblock \emph{arXiv preprint arXiv:1912.02074}, 2019{\natexlab{b}}.

\bibitem[Nakamoto et~al.(2023)Nakamoto, Zhai, Singh, Ma, Finn, Kumar, and Levine]{nakamoto2023cal}
Nakamoto, M., Zhai, Y., Singh, A., Ma, Y., Finn, C., Kumar, A., and Levine, S.
\newblock {Cal-QL}: Calibrated offline rl pre-training for efficient online fine-tuning.
\newblock In \emph{The 37th Conference on Neural Information Processing Systems}, 2023.

\bibitem[Nemirovski(2004)]{nemirovski2004prox}
Nemirovski, A.
\newblock Prox-method with rate of convergence {$O(1/t)$} for variational inequalities with {Lipschitz} continuous monotone operators and smooth convex-concave saddle point problems.
\newblock \emph{SIAM Journal on Optimization}, 15\penalty0 (1):\penalty0 229--251, 2004.

\bibitem[Ng \& Russell(2000)Ng and Russell]{ng2000algorithms}
Ng, A.~Y. and Russell, S.~J.
\newblock Algorithms for inverse reinforcement learning.
\newblock In \emph{Proceedings of the 17th International Conference on Machine Learning}, pp.\  663–670. Morgan Kaufmann, 2000.

\bibitem[Ni et~al.(2021)Ni, Sikchi, Wang, Gupta, Lee, and Eysenbach]{ni2021f}
Ni, T., Sikchi, H., Wang, Y., Gupta, T., Lee, L., and Eysenbach, B.
\newblock {$f$-IRL}: Inverse reinforcement learning via state marginal matching.
\newblock In \emph{Proceedings of the 4th Conference on Robot Learning}, pp.\  529--551. PMLR, 2021.

\bibitem[Orsini et~al.(2021)Orsini, Raichuk, Hussenot, Vincent, Dadashi, Girgin, Geist, Bachem, Pietquin, and Andrychowicz]{orsini2021matters}
Orsini, M., Raichuk, A., Hussenot, L., Vincent, D., Dadashi, R., Girgin, S., Geist, M., Bachem, O., Pietquin, O., and Andrychowicz, M.
\newblock What matters for adversarial imitation learning?
\newblock In \emph{Advances in Neural Information Processing Systems}, volume~34, pp.\  14656--14668. Curran Associates, 2021.

\bibitem[Osa et~al.(2018)Osa, Pajarinen, Neumann, Bagnell, Abbeel, Peters, et~al.]{osa2018algorithmic}
Osa, T., Pajarinen, J., Neumann, G., Bagnell, J.~A., Abbeel, P., Peters, J., et~al.
\newblock An algorithmic perspective on imitation learning.
\newblock \emph{Foundations and Trends in Robotics}, 7\penalty0 (1-2):\penalty0 1--179, 2018.

\bibitem[Pomerleau(1988)]{pomerleau1988alvinn}
Pomerleau, D.~A.
\newblock {ALVINN}: An autonomous land vehicle in a neural network.
\newblock In \emph{Advances in Neural Information Processing Systems}, volume~1, pp.\  305--313. Morgan Kaufmann, 1988.

\bibitem[Puterman(2014)]{puterman2014markov}
Puterman, M.~L.
\newblock \emph{Markov decision processes: discrete stochastic dynamic programming}.
\newblock John Wiley \& Sons, 2014.

\bibitem[Rafailov et~al.(2023)Rafailov, Sharma, Mitchell, Ermon, Manning, and Finn]{rafailov2023direct}
Rafailov, R., Sharma, A., Mitchell, E., Ermon, S., Manning, C.~D., and Finn, C.
\newblock Direct preference optimization: Your language model is secretly a reward model.
\newblock In \emph{The 37th Conference on Neural Information Processing Systems}, 2023.

\bibitem[Rajaraman et~al.(2020)Rajaraman, Yang, Jiao, and Ramchandran]{rajaraman2020toward}
Rajaraman, N., Yang, L., Jiao, J., and Ramchandran, K.
\newblock Toward the fundamental limits of imitation learning.
\newblock In \emph{Advances in Neural Information Processing Systems}, volume~33, pp.\  2914--2924. Curran Associates, 2020.

\bibitem[Rajeswaran et~al.(2017)Rajeswaran, Kumar, Gupta, Vezzani, Schulman, Todorov, and Levine]{rajeswaran2017learning}
Rajeswaran, A., Kumar, V., Gupta, A., Vezzani, G., Schulman, J., Todorov, E., and Levine, S.
\newblock Learning complex dexterous manipulation with deep reinforcement learning and demonstrations.
\newblock \emph{arXiv preprint arXiv:1709.10087}, 2017.

\bibitem[Ratliff et~al.(2006)Ratliff, Bagnell, and Zinkevich]{ratliff2006maximum}
Ratliff, N.~D., Bagnell, J.~A., and Zinkevich, M.~A.
\newblock Maximum margin planning.
\newblock In \emph{Proceedings of the 23rd International Conference on Machine Learning}, pp.\  729–736. ACM, 2006.

\bibitem[Reddy et~al.(2019)Reddy, Dragan, and Levine]{reddy2019sqil}
Reddy, S., Dragan, A.~D., and Levine, S.
\newblock {SQIL}: Imitation learning via reinforcement learning with sparse rewards.
\newblock In \emph{International Conference on Learning Representations}, 2019.

\bibitem[Ross \& Bagnell(2010)Ross and Bagnell]{ross2010efficient}
Ross, S. and Bagnell, D.
\newblock Efficient reductions for imitation learning.
\newblock In \emph{Proceedings of the 13th International Conference on Artificial Intelligence and Statistics}, volume~9, pp.\  661--668. PMLR, 2010.

\bibitem[Russell(1998)]{russell1998learning}
Russell, S.
\newblock Learning agents for uncertain environments (extended abstract).
\newblock In \emph{Proceedings of the 7th Annual Conference on Computational Learning Theory}, pp.\  101–103. ACM, 1998.

\bibitem[Sammut et~al.(1992)Sammut, Hurst, Kedzier, and Michie]{sammut1992learning}
Sammut, C., Hurst, S., Kedzier, D., and Michie, D.
\newblock Learning to fly.
\newblock In \emph{Machine Learning Proceedings 1992}, pp.\  385--393. Morgan Kaufmann, 1992.

\bibitem[Sasaki \& Yamashina(2021)Sasaki and Yamashina]{sasaki2021behavioral}
Sasaki, F. and Yamashina, R.
\newblock Behavioral cloning from noisy demonstrations.
\newblock In \emph{International Conference on Learning Representations}, 2021.

\bibitem[Sasaki et~al.(2019)Sasaki, Yohira, and Kawaguchi]{sasaki2019sample}
Sasaki, F., Yohira, T., and Kawaguchi, A.
\newblock Sample efficient imitation learning for continuous control.
\newblock In \emph{International conference on learning representations}, 2019.

\bibitem[Song et~al.(2022)Song, Zhou, Sekhari, Bagnell, Krishnamurthy, and Sun]{song2022hybrid}
Song, Y., Zhou, Y., Sekhari, A., Bagnell, D., Krishnamurthy, A., and Sun, W.
\newblock {Hybrid RL}: Using both offline and online data can make {RL} efficient.
\newblock In \emph{International Conference on Learning Representations}, 2022.

\bibitem[Sun et~al.(2021)Sun, Mahajan, Hofmann, and Whiteson]{sun2021softdice}
Sun, M., Mahajan, A., Hofmann, K., and Whiteson, S.
\newblock {SoftDICE} for imitation learning: Rethinking off-policy distribution matching.
\newblock \emph{arXiv preprint arXiv:2106.03155}, 2021.

\bibitem[Swamy et~al.(2021{\natexlab{a}})Swamy, Choudhury, Bagnell, and Wu]{swamy2021moments}
Swamy, G., Choudhury, S., Bagnell, J.~A., and Wu, S.
\newblock Of moments and matching: A game-theoretic framework for closing the imitation gap.
\newblock In \emph{Proceedings of the 38th International Conference on Machine Learning}, volume 139, pp.\  10022--10032. PMLR, 2021{\natexlab{a}}.

\bibitem[Swamy et~al.(2021{\natexlab{b}})Swamy, Choudhury, Bagnell, and Wu]{swamy2021critique}
Swamy, G., Choudhury, S., Bagnell, J.~A., and Wu, Z.~S.
\newblock A critique of strictly batch imitation learning.
\newblock \emph{arXiv preprint arXiv:2110.02063}, 2021{\natexlab{b}}.

\bibitem[Syed \& Schapire(2007)Syed and Schapire]{syed2007game}
Syed, U. and Schapire, R.~E.
\newblock A game-theoretic approach to apprenticeship learning.
\newblock In \emph{Advances in Neural Information Processing Systems}, volume~20, pp.\  1449--1456. Curran Associates, 2007.

\bibitem[Syed et~al.(2008)Syed, Bowling, and Schapire]{syed2008apprenticeship}
Syed, U., Bowling, M., and Schapire, R.~E.
\newblock Apprenticeship learning using linear programming.
\newblock In \emph{Proceedings of the 25th International Conference on Machine Learning}, pp.\  1032–1039. ACM, 2008.

\bibitem[Viano et~al.(2022)Viano, Kamoutsi, Neu, Krawczuk, and Cevher]{viano2022proximal}
Viano, L., Kamoutsi, A., Neu, G., Krawczuk, I., and Cevher, V.
\newblock Proximal point imitation learning.
\newblock In \emph{Advances in Neural Information Processing Systems}, volume~35, pp.\  24309--24326. Curran Associates, 2022.

\bibitem[Wagenmaker \& Pacchiano(2023)Wagenmaker and Pacchiano]{wagenmaker2023leveraging}
Wagenmaker, A. and Pacchiano, A.
\newblock Leveraging offline data in online reinforcement learning.
\newblock In \emph{Proceedings of the 40th International Conference on Machine Learning}, volume 202, pp.\  35300--35338. PMLR, 2023.

\bibitem[Wang et~al.(2023{\natexlab{a}})Wang, Lin, and Zhang]{wang2023warm}
Wang, H., Lin, S., and Zhang, J.
\newblock Warm-start actor-critic: From approximation error to sub-optimality gap.
\newblock In \emph{Proceedings of the 40th International Conference on Machine Learning}, volume 202, pp.\  35989--36019. PMLR, 2023{\natexlab{a}}.

\bibitem[Wang et~al.(2018)Wang, Xiong, Han, sun, Liu, and Zhang]{wang2018exponentially}
Wang, Q., Xiong, J., Han, L., sun, p., Liu, H., and Zhang, T.
\newblock Exponentially weighted imitation learning for batched historical data.
\newblock In \emph{Advances in Neural Information Processing Systems}, volume~31, pp.\  6291--6300. Curran Associates, 2018.

\bibitem[Wang et~al.(2019)Wang, Ciliberto, Amadori, and Demiris]{wang2019random}
Wang, R., Ciliberto, C., Amadori, P.~V., and Demiris, Y.
\newblock Random expert distillation: Imitation learning via expert policy support estimation.
\newblock In \emph{Proceedings of the 36th International Conference on Machine Learning}, volume~97, pp.\  6536--6544. PMLR, 2019.

\bibitem[Wang et~al.(2023{\natexlab{b}})Wang, Yang, Gao, Lin, CHEN, Wu, Jia, Song, and Huang]{wang2023train}
Wang, S., Yang, Q., Gao, J., Lin, M.~G., CHEN, H., Wu, L., Jia, N., Song, S., and Huang, G.
\newblock Train once, get a family: State-adaptive balances for offline-to-online reinforcement learning.
\newblock In \emph{The 37th Conference on Neural Information Processing Systems}, 2023{\natexlab{b}}.

\bibitem[Watson et~al.(2024)Watson, Huang, and Heess]{watson2023coherent}
Watson, J., Huang, S., and Heess, N.
\newblock Coherent soft imitation learning.
\newblock In \emph{Advances in Neural Information Processing Systems}, volume~36, pp.\  14540--14583. Curran Associates, 2024.

\bibitem[Xu et~al.(2022{\natexlab{a}})Xu, Zhan, Yin, and Qin]{xu2022discriminator}
Xu, H., Zhan, X., Yin, H., and Qin, H.
\newblock Discriminator-weighted offline imitation learning from suboptimal demonstrations.
\newblock In \emph{Proceedings of the 39th International Conference on Machine Learning}, volume 162, pp.\  24725--24742. PMLR, 2022{\natexlab{a}}.

\bibitem[Xu et~al.(2022{\natexlab{b}})Xu, Li, Yu, and Luo]{xu2022understanding}
Xu, T., Li, Z., Yu, Y., and Luo, Z.-Q.
\newblock Understanding adversarial imitation learning in small sample regime: A stage-coupled analysis.
\newblock \emph{arXiv preprint arXiv:2208.01899}, 2022{\natexlab{b}}.

\bibitem[Yang et~al.(2023)Yang, Yu, Chen, et~al.]{yang2023hybrid}
Yang, H., Yu, C., Chen, S., et~al.
\newblock Hybrid policy optimization from imperfect demonstrations.
\newblock In \emph{The 37th Conference on Neural Information Processing Systems}, 2023.

\bibitem[Yu et~al.(2023)Yu, Yu, Song, Neiswanger, and Ermon]{yu2023offline}
Yu, L., Yu, T., Song, J., Neiswanger, W., and Ermon, S.
\newblock Offline imitation learning with suboptimal demonstrations via relaxed distribution matching.
\newblock In \emph{Proceedings of the AAAI Conference on Artificial Intelligence}, volume~37, pp.\  11016--11024. AAAI Press, 2023.

\bibitem[Yu et~al.(2022)Yu, Kumar, Chebotar, Hausman, Finn, and Levine]{yu2022how}
Yu, T., Kumar, A., Chebotar, Y., Hausman, K., Finn, C., and Levine, S.
\newblock How to leverage unlabeled data in offline reinforcement learning.
\newblock In \emph{Proceedings of the 39th International Conference on Machine Learning}, volume 162, pp.\  25611--25635. PMLR, 2022.

\bibitem[Yue et~al.(2023)Yue, Wang, Shao, Zhang, Lin, Ren, and Zhang]{yue2023clare}
Yue, S., Wang, G., Shao, W., Zhang, Z., Lin, S., Ren, J., and Zhang, J.
\newblock {CLARE}: Conservative model-based reward learning for offline inverse reinforcement learning.
\newblock In \emph{International Conference on Learning Representations}, 2023.

\bibitem[Yue et~al.(2024)Yue, Deng, Wang, Ren, and Zhang]{yue2024federated}
Yue, S., Deng, Y., Wang, G., Ren, J., and Zhang, Y.
\newblock Federated offline reinforcement learning with proximal policy evaluation.
\newblock \emph{Chinese Journal of Electronics}, 33\penalty0 (6):\penalty0 1--13, 2024.

\bibitem[Zeng et~al.(2022)Zeng, Li, Garcia, and Hong]{zeng2022maximum}
Zeng, S., Li, C., Garcia, A., and Hong, M.
\newblock Maximum-likelihood inverse reinforcement learning with finite-time guarantees.
\newblock In \emph{Advances in Neural Information Processing Systems}, volume~35, pp.\  10122--10135. Curran Associates, 2022.

\bibitem[Zeng et~al.(2023)Zeng, Li, Garcia, and Hong]{zeng2023demonstrations}
Zeng, S., Li, C., Garcia, A., and Hong, M.
\newblock When demonstrations meet generative world models: A maximum likelihood framework for offline inverse reinforcement learning.
\newblock In \emph{The 37th Conference on Neural Information Processing Systems}, 2023.

\bibitem[Zhang et~al.(2021)Zhang, Hong, Wang, and Zhang]{zhang2021generalization}
Zhang, J., Hong, M., Wang, M., and Zhang, S.
\newblock Generalization bounds for stochastic saddle point problems.
\newblock In \emph{Proceedings of the 24th International Conference on Artificial Intelligence and Statistics}, pp.\  568--576. PMLR, 2021.

\bibitem[Zhang \& Zanette(2023)Zhang and Zanette]{zhang2023policy}
Zhang, R. and Zanette, A.
\newblock Policy finetuning in reinforcement learning via design of experiments using offline data.
\newblock In \emph{The 37th Conference on Neural Information Processing Systems}, 2023.

\bibitem[Zhang et~al.(2020)Zhang, Cai, Yang, and Wang]{zhang2020generative}
Zhang, Y., Cai, Q., Yang, Z., and Wang, Z.
\newblock Generative adversarial imitation learning with neural network parameterization: Global optimality and convergence rate.
\newblock In \emph{Proceedings of the 37th International Conference on Machine Learning}, volume 119, pp.\  11044--11054. PMLR, 2020.

\bibitem[Ziebart et~al.(2008)Ziebart, Maas, Bagnell, and Dey]{ziebart2008maximum}
Ziebart, B.~D., Maas, A., Bagnell, J.~A., and Dey, A.~K.
\newblock Maximum entropy inverse reinforcement learning.
\newblock In \emph{Proceedings of the 23rd National Conference on Artificial Intelligence}, pp.\  1433–1438. AAAI Press, 2008.

\bibitem[Zolna et~al.(2020)Zolna, Novikov, Konyushkova, Gulcehre, Wang, Aytar, Denil, de~Freitas, and Reed]{zolna2020offline}
Zolna, K., Novikov, A., Konyushkova, K., Gulcehre, C., Wang, Z., Aytar, Y., Denil, M., de~Freitas, N., and Reed, S.
\newblock Offline learning from demonstrations and unlabeled experience.
\newblock In \emph{NeurIPS Workshop on Offline Reinforcement Learning}, 2020.

\bibitem[Zou et~al.(2018)Zou, Su, Song, and Zhu]{zou2018understanding}
Zou, H., Su, H., Song, S., and Zhu, J.
\newblock Understanding human behaviors in crowds by imitating the decision-making process.
\newblock In \emph{Proceedings of the AAAI Conference on Artificial Intelligence}, volume~32, pp.\  7648--7655. AAAI Press, 2018.

\end{thebibliography}
\bibliographystyle{icml2024}

\newpage
\appendix
\onecolumn



\section{Related Work (Extended)}
\label{sec:supp_related_work}

In this section, we discuss recent relevant literature in detail. We shed more light on the differences and strengths of the proposed methods in comparison with the existing offline IL counterparts.

\subsection{Online Imitation Learning}

Imitation learning has a long history (see  \citet{osa2018algorithmic,arora2021survey} for comprehensive overviews). Classical IL methods often cast IL as IRL to improve the efficiency in utilization of expert demonstrations~\citep{russell1998learning,ng2000algorithms,abbeel2004apprenticeship,ratliff2006maximum,syed2007game,ziebart2008maximum,levine2010feature,levine2011nonlinear,choi2013bayesian}. Yet, these methods are computationally prohibitive because they require repetitively running RL as intermediate steps. In the seminal work~\citep{ho2016generative}, the authors introduce \texttt{GAIL} that circumvents the inner-loop RL via building the connection between IL and GAN~\citep{goodfello2016generative}. It trains a discriminator network to distinguish between the state-actions `generated' by the expert and the learning policy; the discriminator in turn acts as a local reward function guiding the policy to take expert behaviors. \texttt{GAIL} and its follow-up AIL works~\citep{li2017infogail,fu2018learning,kostrikov2018discriminator,blonde2019sample,sasaki2019sample,wang2019random,barde2020adversarial,ghasemipour2020divergence,ke2021imitation,ni2021f,swamy2021moments,viano2022proximal,al2023ls} have been proven particularly successful from low-dimensional continuous control to complex and high-dimensional domains like autonomous driving from raw pixels as input~\citep{kuefler2017imitating,zou2018understanding,ding2019goal,arora2021survey,jena2021augmenting}. In theory, \citet{chen2019computation,zhang2020generative,guan2021generative} establish guarantees for \texttt{GAIL} in terms of global convergence and generalization. However, the AIL methods are typically resource-intensive as they require sampling a large number of trajectories during training to approximate the stationary distribution of the learning policy. This nature is inherently risky and costly, particularly in the initial stage where the policy performs randomly, thereby precluding the use of these methods in settings where interactions with the environment are expensive and limited such as in autonomous driving and industrial processes. There is a clear need for IL methods with minimal environmental interactions.

\subsection{Offline Imitation Learning}

The simplest approach to offline IL is \texttt{BC}~\citep{pomerleau1988alvinn} that directly mimics the behavior using regression, whereas it is fundamentally limited by disregarding dynamics information in the demonstration data. It is prone to covariate shift and inevitably suffers from error compounding, i.e., there is no way for the policy to learn how to recover if it deviates from the expert behavior to a state not seen in the expert demonstrations~\citep{rajaraman2020toward}. Considerable research has been devoted to developing new offline IL methods to remedy this problem, which can be generally divided into two categories, \textit{offline IRL} and \textit{direct policy extraction}. We discuss them in what follows.

\textbf{Offline inverse reinforcement learning.} Offline IRL aims at learning a reward function from offline datasets to comprehend and generalize the underlying intentions behind expert behaviors~\citep{lee2019truly}. \citet{zolna2020offline} propose \texttt{ORIL} that constructs a reward function to discriminate expert and exploratory trajectories, followed by an offline RL progress. \citet{chan2021scalable} use a variational method to jointly learn an approximate posterior distribution over the reward and policy. \citet{garg2021iq} simplify the AIL game-theoretic objective over policy and reward functions to an optimization over the soft $Q$-function which implicitly represents both reward and policy. \citet{watson2023coherent} study a problem related to this work on how to improve the \texttt{BC} policy using additional experience. They develop \texttt{CSIL} that exploits a \texttt{BC} policy to define an estimate of a shaped reward function that can then be used to finetune the policy using online interactions. Unfortunately, the heteroscedastic parametric reward functions have undefined values beyond the offline data manifold and easily collapse to the reward limits due to the Tanh transformation and network extrapolation. The reward extrapolation errors may induce the learned reward functions to incorrectly explain the task and misguide the agent in unseen environments~\citep{yue2023clare}. 

To deal with this problem, recent works on offline IRL focus more on model-based methods. \citet{chang2022mitigating} introduce \texttt{MILO} that uses a model uncertainty estimate to penalize the learning reward function on out-of-distribution state-actions. \citet{yue2023clare} incorporate an additional conservatism term to the MaxEnt IRL framework~\cite{ziebart2008maximum}, implicitly penalizing out-of-distribution behaviors from model rollouts. Analogously, \citet{zeng2022maximum} propose \texttt{MLIRL} that can recover the reward function, whose corresponding optimal policy maximizes the likelihood of observed expert demonstrations under a learned conservative world model. However, these model-based approaches introduce additional difficulty in fitting the world model and struggle to scale in high-dimensional environments.


\textbf{Direct policy extraction.} \citet{jarrett2020strictly} propose \texttt{EDM} to enable sampling from an energy-based psuedo-state visitation distribution of the learning policy rather than actual online rollouts; however, it has been questioned that the psuedo-state distribution might be disconnected from the learning policy's true state distribution (see \citet{swamy2021critique}). \citet{sasaki2021behavioral} analyze why the imitation policy trained by \texttt{BC} deteriorates its performance when using noisy demonstrations. They reuse an ensemble of policies learned from the previous iteration as the weight of the original \texttt{BC} objective to extract the expert behaviors. However, this requires that expert data occupies the majority proportion of the offline dataset, otherwise the policy will be misguided to imitate the suboptimal data. \citet{florence2022implicit} propose to reformulate \texttt{BC} using implicit models. They empirically show that it is beneficial to use a conditional energy-based model to represent the policy instead of feed-forward neural networks in the domain of robotic control. \citet{xu2022discriminator} introduce \texttt{DWBC} that exploits a crafted discriminator to distinguish the expert and imperfect data, the outputs of which weights the policy likelihood in \texttt{BC}, with the aim of imitating the noisy demonstrations selectively. Yet, \texttt{DWBC} incorporates the density of the learning policy into the input of the discriminator, losing a rigorous connection with the IL objective. Analogously, \texttt{ISWBC}~\cite{li2023imitation} weight \texttt{BC} by the density ratio of empirical expert data and union offline data, whereas its performance can be guaranteed only when the offline data cover the \textit{stationary state-action distribution} of the expert~\citep[Theorem 3]{li2023imitation}, which is often impractical. In particular, although our reverse KL-regularized policy extraction in \cref{eq:weighted_bc} is similar to the form of weighted \texttt{BC}, there exists a fundamental difference in the form of importance weights:
\begin{align}
	\label{eq:eq12}
	\texttt{ISWBC:}&\quad\max_\pi\mathbb{E}_{(s,a)\sim\tilde\rho^o}\bigg[\frac{\color{blue}\tilde\rho^e(s,a)}{\tilde\rho^o(s,a)}\log\pi(a|s)\bigg]~\Leftrightarrow~\min_\pi \mathbb{E}_{s\sim{\color{blue}\tilde\rho^e}}\left[\KL({\color{blue}\tilde\pi^e(\cdot|s)}\|\pi(\cdot|s))\right]\\
	\label{eq:eq13}
	\texttt{OLLIE:}&\quad\max_\pi\mathbb{E}_{(s,a)\sim\tilde\rho^o}\bigg[\frac{\color{blue}\rho^*(s,a)}{\tilde\rho^o(s,a)}\log\pi(a|s)\bigg]~\Leftrightarrow~\min_\pi\mathbb{E}_{s\sim{\color{blue}\rho^*}}\left[\KL({\color{blue}\pi^*(\cdot|s)}\|\pi(\cdot|s))\right]
\end{align}
where $\tilde\rho^e$ and $\tilde\rho^*$ are the \textit{empirical} expert distribution and the optimum of $\min_\pi\KL(\rho^\pi\|\tilde\rho^*)$, respectively. It is worth noting that $\tilde\rho^e$ typically does \textit{not} satisfy the Bellman flow constraint; in contrast, $\tilde\rho^*$ complies with the constraint and is thereby a real stationary state-action distribution that stays closest to the empirical expert distribution. According to \citet[Theorem 2]{syed2008apprenticeship}, the optimal policy of \cref{eq:eq13} recovers $\tilde\rho^*$, but it is not be ensured by optimizing \cref{eq:eq12}. As a consequence, \texttt{ISWBC} typically necessitates a relatively larger number of expert demonstrations to attain a satisfactory performance (see \cref{sec:complete_experiments}).


From a technical point of view, this work bears a resemblance to the \textit{Stationary Distribution Correction} (DICE) family~\citep{nachum2019dualdice,nachum2019algaedice,kostrikov2020imitation,lee2021optidice,kim2022demodice,kim2022lobsdice,ma2022versatile,yu2023offline}. The terminology DICE is first introduced in \citet{nachum2019dualdice} for off-policy estimation, referring to the ratio between the state-action marginals of the learning policy and a reference policy. Their follow-up work~\citep{nachum2019algaedice} operates the technique in (online) policy optimization. Building on \citet{nachum2019dualdice}, \citet{kostrikov2020imitation} propose \texttt{ValueDICE} that exploits the Donsker-Varadhan representation of KL-divergence and change of variables to transform the AIL problem (distribution matching) to an entirely off-policy objective. Since \texttt{ValueDICE} imitates all given demonstrations, it often requires a large amount of clean expert data, which can be expensive for real-world tasks. To deal with this issue, \citet{kim2022demodice} introduce \texttt{DemoDICE} that incorporates an additional KL-regularization on imperfect demonstrations to enhance offline data support. However, the regularization term requires carefully weighting to avoid overfitting in imperfect data~\citep{yu2023offline} and results in a biased optimization to the IL objective~\citep{li2023imitation}; in contrast, the surrogate objective considered in this work is equivalent to the original problem, $\min_\pi\KL(\rho^\pi\|\tilde\rho^e)$:
\begin{align}
	(\texttt{DemoDICE})~\min_\pi \KL(\rho^\pi\|\tilde\rho^e)+ \alpha \KL(\rho^\pi\|\tilde\rho^o),\quad(\texttt{OLLIE})~ \max_\pi\mathbb{E}_{(s,a)\sim\rho^\pi}\left[ \log\frac{\tilde\rho^e(s,a)}{\tilde\rho^o(s,a)}\right] - \KL(\rho^\pi\|\tilde\rho^o).
\end{align}
Recently, \citet{yu2023offline} propose \texttt{RelaxDICE} employ an asymmetrically-relaxed $f$-divergence instead of KL-divergence to ameliorate the potentially over conservatism of \texttt{DemoDICE}, whereas it still suffers from a biased objective. In addition, \citet{kim2022lobsdice,ma2022versatile} study a relevant but different problem of learning from observation with no access to expert actions. Of note, while these DICE-based methods also explore the Lagrangian duality to deal with the prime problem, their induced dual problems often follow the form of `logarithm-expectation-exponential-expectation'~\citep{kostrikov2020imitation,kim2022demodice,kim2022lobsdice,lee2021optidice,ma2022versatile,yu2023offline}, whereby the two expectations cannot be estimated without bias from the batch of samples~\citep{swamy2021moments,sun2021softdice}. Instead, our derived dual objective follows the convex-concave SSP, enabling an unbiased estimate and inheriting the properties of the convex-concave SSP in both methodology and theory. More importantly, the optimal auxiliary variable $y^*$ of the dual problem is critical in policy extraction and subsequent online finetuning.


\subsection{Offline-To-Online Reinforcement Learning}

The recipe of \textit{pretraining and finetuning} has led to great success in many modern machine learning domains~\citep{brown2020language,he2022masked}. Very recently, numerous efforts seek to translate such a recipe to decision-making problems, which utilizes offline RL for initializing value functions and policies from fixed datasets and subsequently uses RL to improve the initialization~\cite{lee2022offline,mark2022finetuning,song2022hybrid,ball2023efficient,li2023accelerating,nakamoto2023cal,wagenmaker2023leveraging,wang2023warm,wang2023train,yang2023hybrid,zhang2023policy}. In light of these advances, a potential solution to offline-to-online IL might be abstracting a reward function from offline, followed by adapting forward RL to further finetune the policy. However, it is inherently indirect, and the reward extrapolation would largely exacerbate the difficulty in both offline and online IL.

\section{Convexity of $L(\nu)$}
\label{sec:convexity}

Recall the expression of $L(\nu)$:
\begin{align}
	L(\nu) = (1-\gamma)\mathbb{E}_{s\sim\mu}[\nu(s)] + \mathbb{E}_{(s,a)\sim\tilde\rho^o}\Big[\exp\big(\tilde{R}(s,a) + \gamma\sum_{s'}\nu(s')T(s'|s,a) - \nu(s)-1\big)\Big].
\end{align}
For any $\nu_1,\nu_2$ ($\nu_1\neq\nu_2$) and $\lambda\in(0,1)$, we have
\begin{align}
	L(\lambda\nu_1 + (1-\lambda)\nu_2)=\;&   \mathbb{E}_{(s,a)\sim\tilde\rho^o}\Big[\exp\Big(\tilde{R}(s,a)+ \gamma\sum\nolimits_{s'}\big(\lambda\nu_1(s')+ (1-\lambda)\nu_2(s')\big)T(s'|s,a)  \nonumber\\
	& - \big(\lambda\nu_1(s) + (1-\lambda)\nu_2(s)\big)-1\Big)\Big] + (1-\gamma)\mathbb{E}_{s\sim\mu}\big[\lambda\nu_1(s) + (1-\lambda)\nu_2(s)\big] \nonumber\\
	=\;& \mathbb{E}_{(s,a)\sim\tilde\rho^o}\Big[\exp\Big(\lambda\big(\tilde{R}(s,a) + \gamma\sum\nolimits_{s'}\nu_1(s')T(s'|s,a) - \nu_1(s)-1\big)\nonumber\\
	&+(1-\lambda)\big(\tilde{R}(s,a) + \gamma\sum\nolimits_{s'}\nu_2(s')T(s'|s,a) - \nu_2(s)-1\big)\Big)\Big] \nonumber\\
	&+ \lambda(1-\gamma)\mathbb{E}_{s\sim\mu}[\nu_1(s)]+ (1-\lambda)(1-\gamma)\mathbb{E}_{s\sim\mu}[\nu_2(s)]\nonumber\\
	\le\;& \mathbb{E}_{(s,a)\sim\tilde\rho^o}\Big[\lambda\exp\big(\tilde{R}(s,a) + \gamma\sum\nolimits_{s'}\nu_1(s')T(s'|s,a) - \nu_1(s)-1\big)\nonumber\\
	&+(1-\lambda)\exp\big(\tilde{R}(s,a) + \gamma\sum\nolimits_{s'}\nu_2(s')T(s'|s,a) - \nu_2(s)-1\big)\Big] \nonumber\\
	&+ \lambda(1-\gamma)\mathbb{E}_{s\sim\mu}[\nu_1(s)]+ (1-\lambda)(1-\gamma)\mathbb{E}_{s\sim\mu}[\nu_2(s)]\nonumber\\
	=\;& \lambda L(\lambda\nu_1) + (1-\lambda)L(\nu_2)
\end{align}
where the last inequality follows the convexity of $\exp(\cdot)$. From the definition of the convex function, we obtain the result.

\section{Auxiliary Reward Scaling}
\label{sec:reward_scaling}

For ${\alpha}>0,\beta\ge0$, consider the reward function is scaled by $\tilde R_{\alpha}(s,a)\doteq {\alpha}\tilde R(s,a) + \beta$.  Problem~(\ref{eq:objective_3})--(\ref{eq:eq3}) can be rewritten as
\begin{align}
	\max_{\rho\ge0}&~\mathbb{E}_{(s,a)\sim\rho}\big[\tilde{R}_{\alpha}(s,a)\big] - {\alpha}\KL(\rho\|\tilde\rho^o)\\
	\mathrm{s.t.}&~f_s(\rho)=0,~\forall s\in\mathcal{S}.
\end{align}
The objective and constraints are concave and affine on $\rho$,  and hence it is a convex optimization problem. The corresponding Lagrangian can be expressed as
\begin{align}
	\label{eq:eq11}
	L(\rho,\nu)
	&=\sum_{s,a}\rho(s,a)\bigg(\tilde{R}_{\alpha}(s,a) + \gamma\sum_{s'}\nu(s')T(s'|s,a) - \nu(s)-{\alpha}\log\frac{\rho(s,a)}{\tilde\rho^o(s,a)}\bigg) + (1-\gamma)\sum_{s}\nu(s)\mu(s)\nonumber\\
	&=\sum_{s,a}\rho(s,a)\bigg(\delta_\nu(s,a)-{\alpha}\log\frac{\rho(s,a)}{\tilde\rho^o(s,a)}\bigg) + (1-\gamma)\sum_{s}\nu(s)\mu(s)
\end{align}
where $\nu$ is the Lagrangian multiplier, and $\delta_{\nu}(s,a)=\tilde{R}_{\alpha}(s,a) + \gamma\sum_{s'}\nu(s')T(s'|s,a) - \nu(s)$. From the Slater's condition, the strong duality holds. Taking derivative of $L$ w.r.t. $\rho(s,a)$, we have
\begin{align}
	\frac{\partial L}{\partial\rho(s,a)} = \delta_{\nu}(s,a) -{\alpha}\log\frac{\rho(s,a)}{\tilde\rho^o(s,a)} - {\alpha}.
\end{align}
Letting the derivative to 0, we can write
\begin{align}
	\label{eq:eq14}
	\rho(s,a)=\tilde\rho^o(s,a)\exp\left(\frac{1}{{\alpha}}\delta_{\nu}(s,a)-1\right).
\end{align}
Substituting \cref{eq:eq14} in \cref{eq:eq11}, we obtain the dual problem:
\begin{align}
	\min_\nu L(\nu)= {\alpha}\mathbb{E}_{(s,a)\sim\tilde\rho^o}\left[\exp\left(\frac{1}{{\alpha}}\delta_{\nu}(s,a)-1\right)\right]+ (1-\gamma)\mathbb{E}_{s\sim\mu}\left[\nu(s)\right].
\end{align}
Denote $f(x)\doteq\exp(ax - b)$ ($a>0,b\ge0$). From the definition of convex conjugate, the following fact holds:
\begin{align}
	f^*(y) 
	= \max_x yx - \exp(ax - b)
	= \left(\frac{b}{a} - \frac{1}{a} + \frac{1}{a}\log\left(\frac{1}{a}y\right)\right)y.
\end{align}
Due to the strict convexity of $f(x)$, $f^{**} = f$, thereby
\begin{align}
	\exp\left(\frac{1}{{\alpha}}\delta_{\nu}(s,a)-1\right) =\max_{y(s,a)} \delta_{\nu}(s,a)y(s,a)- {\alpha}\log\left({\alpha} y(s,a)\right)y(s,a).
\end{align}
Thus, the dual problem is equivalent to the following minimax problem:
\begin{align}
	\label{eq:eq15}
	\min_\nu\max_y F(\nu,y)= {\alpha}\mathbb{E}_{(s,a)\sim\tilde\rho^o}\left[\delta_{\nu}(s,a)y(s,a)- {\alpha}\log\left({\alpha} y(s,a)\right)y(s,a)\right]+ (1-\gamma)\mathbb{E}_{s\sim\mu}\left[\nu(s)\right].
\end{align}
Denote $\tilde\delta_{\nu}(s,a,s')=\tilde{R}_{\alpha}(s,a) + \gamma\nu(s') - \nu(s)$. The empirical counterpart of \cref{eq:eq15} is 
\begin{align}
	\min_\nu\max_y \tilde F(\nu,y)= {\alpha}\mathbb{E}_{(s,a,s')\sim\mathcal{D}_o}\left[\tilde\delta_{\nu}(s,a,s')y(s,a)- {\alpha}\log\left({\alpha} y(s,a)\right)y(s,a)\right]+ (1-\gamma)\mathbb{E}_{s\sim\mu}\left[\nu(s)\right].
\end{align}
Taking $\frac{\partial F}{\partial y(s,a)}=0$ yields
\begin{align}
	\exp\left(\frac{1}{{\alpha}}\delta_{\nu^*}(s,a)-1\right) = {\alpha} y^*(s,a).
\end{align}
From \cref{eq:eq14}, given the optimal $\nu^*$, the optimal $\rho^*$ equals
\begin{align}
	\rho^*(s,a)=\tilde\rho^o(s,a)\exp\left(\frac{1}{{\alpha}}\delta_{\nu^*}(s,a)-1\right) = \alpha y^*(s,a)\tilde\rho^o(s,a).
\end{align}
Similarly to Eqs.~(\ref{eq:optimal_rho})--(\ref{eq:imitation_policy}), the offline policy still follows \cref{eq:imitation_policy}. Based on \cref{eq:d_init}, the discriminator initialization changes to
\begin{align}
	D_0(s,a)=\frac{\rho^*(s,a)}{\rho^*(s,a) + \tilde\rho^e(s,a)}=\left(1 + \frac{\tilde\rho^e(s,a)}{\tilde\rho^o(s,a)}\cdot\frac{\tilde\rho^o(s,a)}{\rho^*(s,a)}\right)^{-1}=\left(1 + \frac{d^*(s, a)}{1-d^*(s, a)}\cdot\frac{1}{{\alpha} y^*(s,a)}\right)^{-1}.
\end{align}



\section{Undiscounted Case}
\label{sec:undiscounted}

In undiscounted case ($\gamma=1$), the stationary state-action distribution is expressed as
\begin{align}
	\rho^\pi(s,a)=\lim_{H\rightarrow\infty}\frac{1}{H}\sum^{H-1}_{h=0}\Pr(s_h=s,a_h=a\mid T,\pi,\mu).
\end{align}
It renders Problem (\ref{eq:objective_3})--(\ref{eq:eq3}) ill-posed: if $\rho^*$ is the optimizer to the problem, $a\rho^*$ is still the optimizer for any $a>0$. Building on \citet{lee2021optidice}, it can be overcome by introducing normalization constraint $\sum_{s,a}\rho(s,a)=1$ to Problem (\ref{eq:objective_3})--(\ref{eq:eq3}). We reformulate Problem~(\ref{eq:objective_2}) as follows:
\begin{align}
	\max_{\rho\ge0}&~\mathbb{E}_{(s,a)\sim\rho}\big[\tilde{R}(s,a)\big] - \KL(\rho\|\tilde\rho^o)\\
	\mathrm{s.t.}&~\sum_{s,a}\rho(s,a)=1,~f_s(\rho)=0,~\forall s.
\end{align}
Here, $f_s(\rho)=\sum_{a,s'} T(s|s',a)\rho(s',a)-\sum_a\rho(s,a)$. The objective and constraints are concave and affine on $\rho$,  and hence it remains a convex optimization problem. The corresponding Lagrangian can be expressed as
\begin{align}
	L(\rho,\nu,\lambda)
	\doteq\sum_{s,a}\rho(s,a)(\delta_{\nu}(s,a)-\log\frac{\rho(s,a)}{\tilde\rho^o(s,a)}+\lambda)  - \lambda
\end{align}
where $\delta_{\nu}(s,a)\doteq \tilde{R}(s,a) + \gamma\sum_{s'}\nu(s')T(s'|s,a) - \nu(s)$. The derivative w.r.t. $\rho(s,a)$ is
\begin{align}
	\frac{\partial L}{\partial\rho(s,a)} = \tilde{R}(s,a) +\gamma\sum_{s'}\nu(s')T(s'|s,a)-\nu(s) -\log\frac{\rho(s,a)}{\tilde\rho^o(s,a)} + \lambda - 1.
\end{align}
Taking the derivative to 0 yields
\begin{align}
	\rho(s,a)=\tilde\rho^o(s,a)\exp\left(\delta_{\nu}(s,a)+\lambda-1\right).
\end{align}
Clearly, the dual problem is
\begin{align}
	\min_{\nu,\lambda} L(\nu,\lambda) \doteq\mathbb{E}_{(s,a)\sim\tilde\rho^o}\big[\exp(\delta_{\nu}(s,a) +\lambda-1)\big].
\end{align}
Denote $f(x)\doteq\exp(x + \lambda - 1)$. From the definition of convex conjugate, we can obtain
\begin{align}
	f^*(y) 
	= \max_x yx - \exp(x + \lambda - 1)
	= y\log y - \lambda y.
\end{align}
Due to the strict convexity of $f(x)$, the following fact holds true:
\begin{align}
	\exp\left(\delta_{\nu}(s,a)+\lambda-1\right) =\max_{y(s,a)} \delta_{\nu}(s,a)y(s,a) + \lambda y(s,a)- y(s,a)\log y(s,a).
\end{align}
The dual problem equals the following minimax problem:
\begin{align}
	\label{eq:eq7}
	\min_{\nu,\lambda}\max_y F(\nu,\lambda,y)= \mathbb{E}_{(s,a)\sim\tilde\rho^o}\left[\delta_{\nu}(s,a)y(s,a) + \lambda y(s,a)- y(s,a)\log y(s,a)\right].
\end{align}
Denote the optimum of \cref{eq:eq7} as $\nu^*,\lambda^*,y^*$, which satisfies
\begin{align}
	y^*(s,a) = \exp \left(\delta_{\nu^*}(s,a) + \lambda^*-1\right).
\end{align}
Hence, we have
\begin{align}
	\pi^*(a|s) = \frac{\rho^*(s,a)}{\sum_{a'}\rho^*(s,a')}=\frac{\tilde\rho^o(s,a)\exp\left(\delta_{\nu^*}(s,a)+ \lambda^*-1\right)}{z(s)} = \frac{\tilde\rho^o(s,a)y^*(s,a)}{z(s)}.
\end{align}
Regarding the discriminator, the following holds:
\begin{align}
	D_0(s,a) = \frac{\rho^*(s,a)}{\rho^*(s,a) + \tilde\rho^e(s,a)}=\left(1 + \frac{\tilde\rho^e(s,a)}{\tilde\rho^o(s,a)}\cdot\frac{\tilde\rho^o(s,a)}{\rho^*(s,a)}\right)^{-1}=\left(1 + \frac{d^*(s, a)}{1-d^*(s, a)}\cdot\frac{1}{ y^*(s,a)}\right)^{-1}.
\end{align}
Therefore, the policy updating and discriminator initialization keep consistent with \cref{eq:imitation_policy,eq:d_init}.

\section{A Byproduct for Offline Reinforcement Learning}
\label{sec:by_product}

\begin{algorithm}[t]
	\centering
	\caption{Adapting \texttt{OLLIE} to offline reinforcement learning}
	\label{alg:offline_rl_variant}
	\begin{algorithmic}[1]
		\STATE Initialize parameters $\phi_\nu$, $\phi_y$, and $\theta$
		\FOR{$i=1$ {\bfseries to} $n$}
		\STATE $\phi_\nu\leftarrow \phi_\nu - \eta_\nu \tilde\nabla_{\phi_\nu } J(\phi_\nu,\phi_y)$
		\STATE $\phi_y\leftarrow \phi_y + \eta_y \tilde\nabla_{\phi_y}J(\phi_\nu,\phi_y)$
		\ENDFOR
		\FOR{$i=1$ {\bfseries to} $n'$}
		\STATE $\theta\leftarrow\theta - \eta_\pi \tilde\nabla J(\pi_\theta)$
		\ENDFOR
	\end{algorithmic}
\end{algorithm}

Given true reward function $R$, the offline policy optimization problem can be cast as~\citep{lee2021optidice}:
\begin{align}
	\label{eq:offline_rl}
	\max_\pi\mathbb{E}_{(s,a)\sim\rho^\pi}\big[ R(s,a)\big] - \alpha\KL(\rho^\pi\|\tilde\rho^o)
\end{align}
where $\rho^o$ is the empirical state-action distribution of an offline dataset, $\mathcal{D}_o\doteq\{(s_i,a_i,r_i,s'_i)\}^{n_o}_{i=1}$, with MDP transition $(s_i,a_i,r_i,s'_i)$ collected from an unknown behavior policy. Here, hyperparameter $\alpha>0$ balances between the reward maximization and the penalization of the distributional shift.\footnote{The solutions for offline RL revolve around the idea that the learned policy should be confined close to the data-generating process to remedy the performance degradation due to extrapolation error~\citep{fujimoto2021minimalist}.} From the derivation from Eqs.~(\ref{eq:objective_3})--(\ref{eq:weighted_bc}), we immediately obtain an offline algorithm for Problem~(\ref{eq:offline_rl}) as follows.
\begin{enumerate}[itemsep=0pt,topsep=0pt]
	\renewcommand{\labelenumi}{\textit{\theenumi)}}
	\item \textit{Estimating the exponential advantage.} First, solve the following convex-concave SSP iteratively to converge:
	\begin{align*}
		\min_\nu\max_y J(\nu,y) =  \alpha\mathbb{E}_{(s,a,r,s')\sim\mathcal{D}_o}\left[\tilde\delta_{\nu}(s,a,r,s')y(s,a)-\alpha\log\left(\alpha y(s,a)\right) y(s,a)\right]+ (1-\gamma)\mathbb{E}_{s\sim\mathcal{D}_o(s_0)}\left[\nu(s)\right]
	\end{align*}
	with $\tilde\delta_{\nu}(s,a,r,s')= r + \gamma\nu(s') - \nu(s)$. 
	\item \textit{Extracting the offline policy.} After obtaining $y^*$, extract the offline policy by the weighted BC:\footnote{We exclude the policy extraction based on reverse KL-divergence since offline data is typically diverse, and estimating $\rho^o$ is thereby challenging.}
	\begin{align*}
		\max_\pi J(\pi)=\mathbb{E}_{(s,a)\sim\mathcal{D}_o}\left[y^*(s,a)\log\pi(a|s)\right].
	\end{align*}
\end{enumerate}
We encapsulate the pseudocode in \cref{alg:offline_rl_variant}. In contrast with \citet{lee2019truly} using a biased gradient, \cref{alg:offline_rl_variant} is unbiased and inherits the properties of convex-concave SSPs in terms of convergence and generalization. For a deeper understanding of the rationale behind \cref{eq:offline_rl}, we take a close look at Problem~(\ref{eq:offline_rl}):
\begin{align}
	&\mathbb{E}_{(s,a)\sim\rho^\pi}\left[R(s,a)\right]- \alpha\KL(\rho^\pi\|\rho^o)\nonumber\\
	=\;&\mathbb{E}_{(s,a)\sim\rho^\pi}\left[R(s,a)-\alpha\log\frac{\rho^\pi(s,a)}{\rho^o(s,a)}\right]\nonumber\\
	\Leftrightarrow\;&\mathbb{E}_{(s,a)\sim\rho^\pi}\left[\frac{1}{\alpha}R(s,a)-\log\frac{\rho^\pi(s,a)}{\rho^o(s,a)}\right]\tag{omitting coefficient $\alpha$}\\
	=\;&\mathbb{E}_{(s,a)\sim\rho^\pi}\left[-\left(\log\frac{\rho^\pi(s,a)}{\rho^o(s,a)}-\log\exp\left(\frac{1}{\alpha}R(s,a)\right)\right)\right]\nonumber\\
	=\;&\mathbb{E}_{(s,a)\sim\rho^\pi}\left[-\log\frac{\rho^\pi(s,a)}{\rho^o(s,a)\exp(\frac{1}{\alpha}R(s,a))}\right]\nonumber\\
	=\;&\mathbb{E}_{(s,a)\sim\rho^\pi}\left[\log Z-\log\frac{\rho^\pi(s,a)}{\frac{1}{Z}\rho^o(s,a)\exp(\frac{1}{\alpha}R(s,a))}\right]\nonumber\\
	\Leftrightarrow\;&-\KL(\rho^\pi\|\rho^*)
\end{align}
where $Z\doteq\sum_{s,a}\rho^o(s,a)\exp\left(\frac{1}{\alpha}r(s,a)\right)$ is the normalization term, and $\rho^*$ here represent
\begin{align}
	\label{eq:eq8}
	\rho^*(s,a) = \frac{\rho^o(s,a)\exp(\frac{1}{\alpha}R(s,a))}{Z}.
\end{align}
\cref{eq:eq8} reveals the objective of Problem~(\ref{eq:offline_rl}) encourages the learning policy to pursue higher rewards while staying more on the data support to combat the distributional shift. 

\begin{figure*}[t]
	\centering
	{\includegraphics[width=\textwidth]{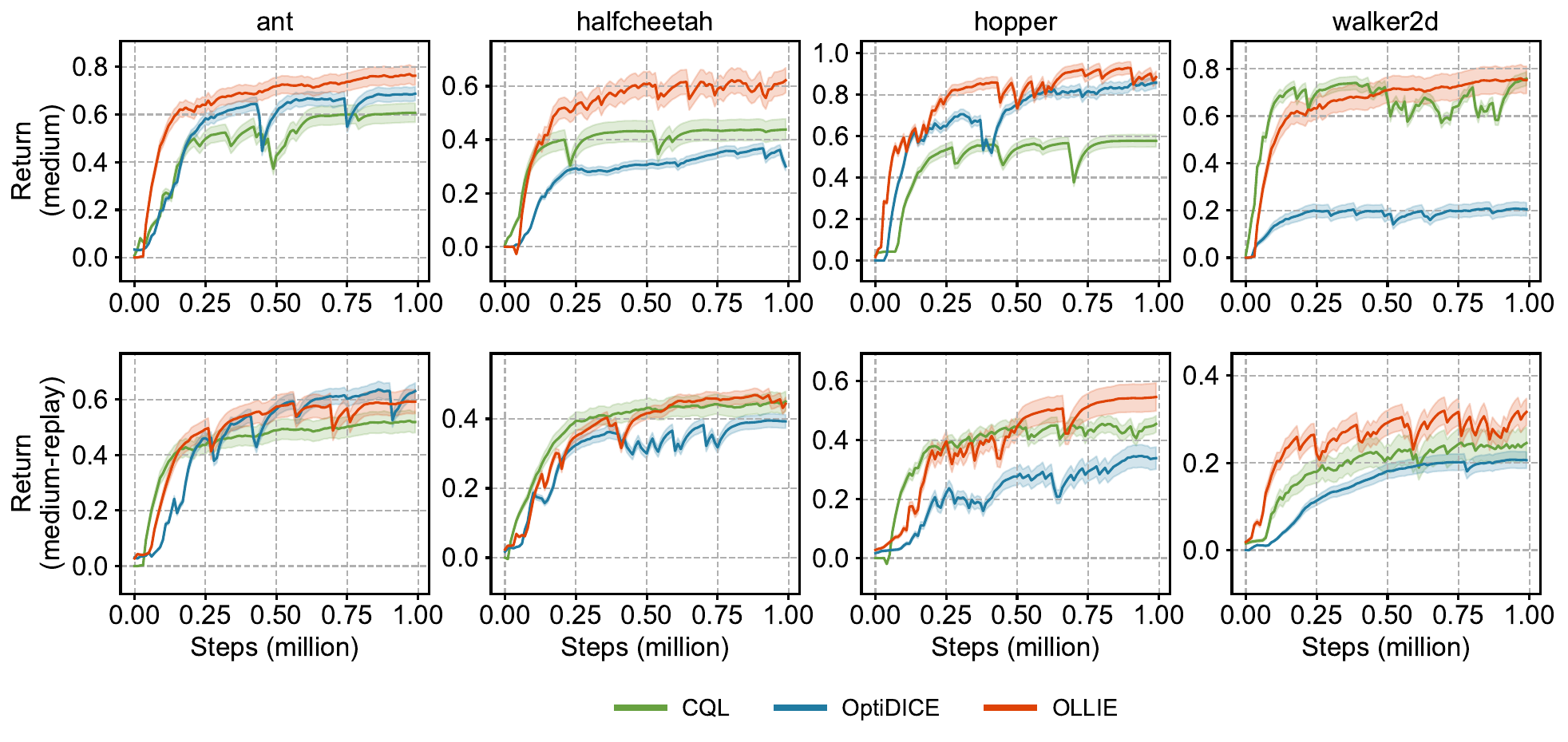}}
	\caption{Comparable performance in offline RL. The results are normalized and averaged over five random seeds with standard deviation depicted by uncertainty intervals. We employ the \texttt{medium} and \texttt{random} datasets from \texttt{D4RL}.}
	\label{fig:offline_rl}
\end{figure*}

Empirically, we test the performance of  \cref{alg:offline_rl_variant} against \texttt{OptiDICE}~\citep{lee2021optidice} and \texttt{CQL}~\citep{kumar2020conservative} across four widely-used MuJoCo continuous-control tasks. We implement \texttt{OptiDICE} using its official implementation (\url{https://github.com/secury/optidice}) and \texttt{CQL} using an off-the-shelf implementation (\url{https://github.com/corl-team/CORL/blob/main/algorithms/offline/cql.py}). As illustrated in \cref{fig:offline_rl}, \texttt{OLLIE} exhibits competitive performance against \texttt{OptiDICE} and outperforms \texttt{CQL} by a wide margin, revealing potential in the offline RL problems. We leave the comparison between \texttt{OLLIE} and more recent methods for future work.

\clearpage
\section{Experimental Setup}
\label{sec:experiment_setting}

\subsection{Environments and Tasks}
\label{sec:environments}

We evaluate our method on a number of environments (Robomimic, MuJoCo, Adroit, FrankaKitchen, and AntMaze) which are widely used in prior studies~\citep{nakamoto2023cal,watson2023coherent}. We elaborate in what follows.
\begin{itemize}[leftmargin=*,itemsep=0pt,topsep=0pt]
	\item \textbf{Vision-based Robomimic.} The Robomimic tasks (\texttt{lift}, \texttt{can}, \texttt{square}) involve controlling a 7-DoF simulated hand robot~\citep{robomimic2021}, with pixelized observations as shown in \cref{fig:robomimic}. The robot is tasked with lifting objects, picking and placing cans, and picking up a square nut to place it on a rod from random initializations. 
	\begin{figure}[ht]
		\centering  \includegraphics[width=0.55\linewidth]{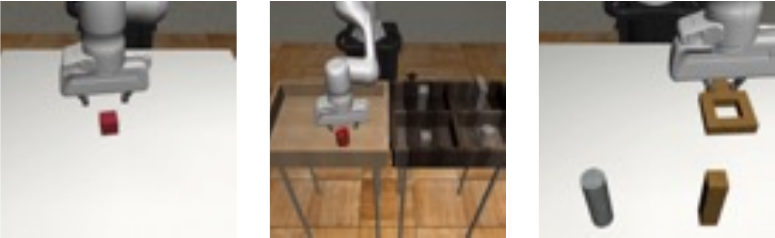}
		\vskip -0.1in
		\caption{Observations of vision-based Robomimic tasks. From left to right: \texttt{lift}, \texttt{can}, \texttt{square}.}
		\vskip -0.1in
		\label{fig:robomimic}
	\end{figure}
	
	\item \textbf{Vision-based MuJoCo.} The MuJoCo locomotion tasks (\texttt{ant}, \texttt{hopper}, \texttt{halfcheetah}, \texttt{walker2d}) are popular benchmarks used in existing work. In addition to the standard setting, we also consider vision-based MuJoCo tasks which uses the image observation as input (see \cref{fig:mujoco_vision}).
	\begin{figure}[ht]
		\centering  \includegraphics[width=0.7\linewidth]{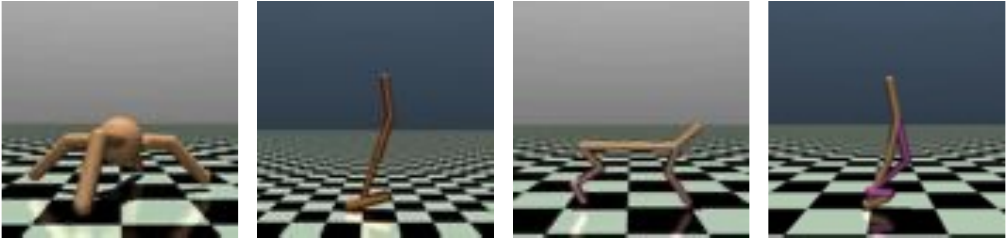}
		\vskip -0.1in
		\caption{Observations of vision-based MuJoCo tasks. From left to right: \texttt{ant}, \texttt{hopper}, \texttt{halfcheetah}, \texttt{walker2d}.}
		\vskip -0.1in
		\label{fig:mujoco_vision}
	\end{figure}
	
	\item \textbf{Adroit.} The Adroit tasks (\texttt{hammer}, \texttt{door}, \texttt{pen}, and \texttt{relocate}) \cite{rajeswaran2017learning} involve controlling a  28-DoF hand with five fingers tasked with hammering a nail, opening a door, twirling a pen, or picking up and moving a ball. 
	\begin{figure}[ht]
		\centering  \includegraphics[width=0.7\linewidth]{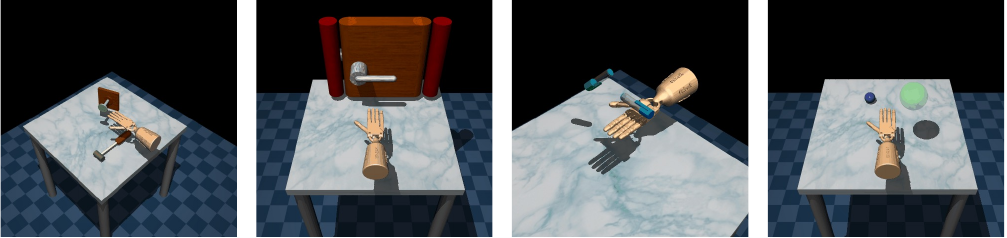}
		\vskip -0.1in
		\caption{Adroit tasks: \texttt{hammer}, \texttt{door}, \texttt{pen}, and \texttt{relocate} (from left to right).}
		\vskip -0.1in
		\label{fig:adroit}
	\end{figure}
	
	\item \textbf{FrankaKitchen.} The FrankaKitchen tasks (\texttt{complete}, \texttt{partial}, \texttt{undirect}), proposed by \citet{gupta2019relay}, involve controlling a 9-DoF Franka robot in a kitchen environment containing several common household items: a microwave, a kettle, an overhead light, cabinets, and an oven. The goal of each task is to interact with the items to reach a desired goal configuration.
	In the \texttt{undirect} task, the robot requires opening the microwave. In the \texttt{partial} task, the robot must first opening the microwave and subsequently moving the kettle. In the \texttt{complete} task, the robot need to open the microwave, move the kettle, flip the light switch, and slide open the cabinet door sequentially (see \cref{fig:kitchen}). These tasks are especially challenging, since they require composing parts of trajectories, preciselong-horizon manipulation, and handling human-provided teleoperation data.
	\begin{figure}[t]
		\centering  \includegraphics[width=0.95\linewidth]{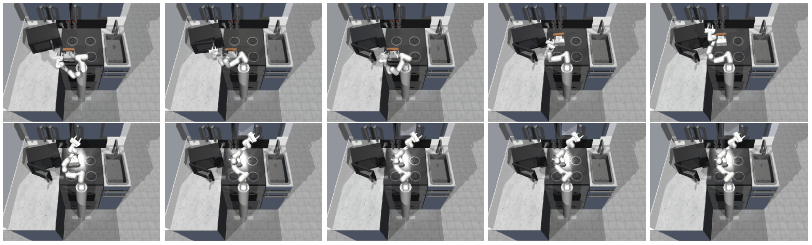}
		\vskip -0.1in
		\caption{Visualized success for opening the microwave, moving the kettle, turning on the light switch, sliding the slider.}
		\label{fig:kitchen}
	\end{figure}
	
	\item \textbf{AntMaze.} The AntMaze tasks require controlling an 8-Degree of Freedom (DoF) quadruped robot to move from a startingpoint to a fixed goal location~\citep{fu2020d4rl}. 
	Three maze layouts (\texttt{umaze}, \texttt{medium}, and \texttt{large}) are provided from small to large.
	\begin{figure}[ht]
		\centering  \includegraphics[width=0.5\linewidth]{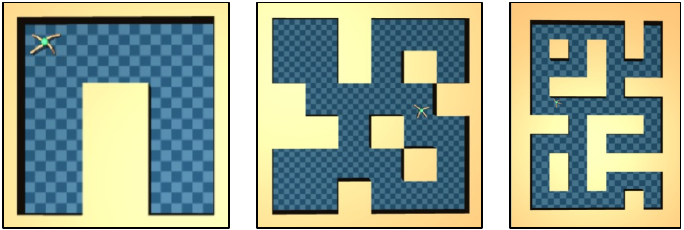}
		\vskip -0.1in
		\caption{AntMaze with three maze layouts, \texttt{umaze}, \texttt{medium}, and \texttt{large} (from left to right).}
		\label{fig:antmaze_layout}
		\vskip -0.2in
	\end{figure}
\end{itemize}


Detailed information about the environments including observation space, action space, and expert performance is provided in \cref{tab:env_performance,tab:env_image_performance}, where expert and random scores are averaged over 1000 episodes.

\begin{table}[ht]
	\vskip -0.25in
	\centering
	\caption{Details of continuous-control tasks.}
	\label{tab:env_performance}
	\vskip 0.1in
	{\small \begin{threeparttable}
	\begin{tabular}{l|c|c|r|r}
		Task & State dim. & Action dim. & \texttt{random}\tnote{*} & \texttt{expert}\tnote{*} \\ 
		\hline
		\texttt{ant} & $27$ & $8$ & $-325.60$ & $3879.70$ \\
		\texttt{halfcheetah} & $17$ & $6$ & $-280.18$ & $12135.00$ \\
		\texttt{hopper} & $11$ & $3$ & $-20.27$ & $3234.30$ \\
		\texttt{walker2d} & $17$ & $6$ & $1.63$ & $4592.30$ \\
		\texttt{antmaze} & $27$ & $8$ & $0.00$ & $1.00$ \\
		\texttt{door} & $39$ & $28$ & $-56.51$ & $2880.57$ \\
		\texttt{hammer} & $46$ & $26$ & $-274.86$ & $12794.13$ \\
		\texttt{pen} & $45$ & $24$ & $96.26$ & $3076.83$ \\
		\texttt{relocate} & $39$ & $30$ & $-6.43$ & $4233.88$ \\
		\texttt{FrankaKitchen} & $59$ & $9$ & $0.00$ & $1.00 $ \\
		\hline
	\end{tabular}	{\scriptsize \begin{tablenotes}
		\item[*] Average scores over 1000 trajectories of \texttt{expert} and \texttt{random}.
	\end{tablenotes}}
	\end{threeparttable}}
	\vskip -0.2in
\end{table}

\begin{table}[ht]
	\vskip -0.15in
	\centering
	\caption{Details of vision-based tasks.}
	\label{tab:env_image_performance}
	\vskip 0.1in
	{\small \begin{threeparttable}
	\begin{tabular}{l|c|c|r|r}
		Task & \multicolumn{1}{l|}{State dim.} & Action dim. & \texttt{random}\tnote{*} & \texttt{expert}\tnote{*} \\ 
		\hline
		\texttt{ant} & ($84\times84$) & $8$ & $-325.60$ & $3879.70$ \\
		\texttt{halfcheetah} & ($84\times84$) & $6$ & $-280.18$ & $12135.00$ \\
		\texttt{hopper} & ($84\times84$) & $3$ & $-20.27$ & $3234.30$ \\
		\texttt{walker2d} & ($84\times84$) & $6$ & $1.63$ & $4592.30$ \\
		\texttt{lift} & ($84\times84$) & $7$ & $0.00$ & $1.00$ \\
		\texttt{can} & ($84\times84$) & $7$ & $0.00$ & $1.00$ \\
		\texttt{square} & ($84\times84$) & $7$ & $0.00$ & $1.00$ \\
		\hline
	\end{tabular}
	\begin{tablenotes}
		{\scriptsize \item[*] Average scores over 1000 trajectories of \texttt{expert} and \texttt{random}.}
	\end{tablenotes}
	\end{threeparttable}}
	\vskip -0.2in
\end{table}

\subsection{Datasets}
\label{sec:datasets}

During offline training, we use the \texttt{D4RL} datasets~\citep{fu2020d4rl} for AntMaze, MuJoCo, Adroit, and FrankaKitchen, and we use the \texttt{robomimic}~\citep{robomimic2021} datasets for Robomimic. cref{table:dataset} details the specific datasets used for the expert and imperfect data in each task (the exploited numbers of expert and imperfect trajectories may vary across experiments). In addition, we construct vision-based MuJoCo datasets using the same method as \citet{fu2020d4rl}: the expert and imperfect data use video samples from a policy trained to completion with \texttt{SAC}~\citep{haarnoja18soft} and a randomly initialized policy, respectively.

\begin{sidewaystable}
	\centering
	\caption{Datasets used in different tasks.}
	\label{table:dataset}
	\vskip 0.1in
	{\small \begin{threeparttable}	
	\begin{tabular}{l|l|l|l|ll}
		\multirow{2}{*}{Domain} & \multirow{2}{*}{Dataset} & \multirow{2}{*}{Task} & \multirow{2}{*}{Expert data} & \multirow{2}{*}{Imperfect data} &  \\
		&  &  &  &  &  \\ 
		\cline{1-5}
		\multirow{16}{*}{MuJoCo} & \multirow{16}{*}{\texttt{D4RL}} & \multirow{4}{*}{\texttt{ant}} & \multirow{4}{*}{\texttt{ant-expert-v2}} & \texttt{ant-random-v2} &  \\
		&  &  &  & \texttt{ant-medium-replay-v2} &  \\
		&  &  &  & \texttt{ant-medium-v2} &  \\
		&  &  &  & \texttt{ant-medium-expert-v2} &  \\ 
		\cline{3-5}
		&  & \multirow{4}{*}{\texttt{hopper}} & \multirow{4}{*}{\texttt{hopper-expert-v2}} & \texttt{hopper-random-v2} &  \\
		&  &  &  & \texttt{hopper-medium-replay-v2} &  \\
		&  &  &  & \texttt{hopper-medium-v2} &  \\
		&  &  &  & \texttt{hopper-medium-expert-v2} &  \\ 
		\cline{3-5}
		&  & \multirow{4}{*}{\texttt{halfcheetah}} & \multirow{4}{*}{\texttt{halfcheetah-expert-v2}} & \texttt{halfcheetah-random-v2} &  \\
		&  &  &  & \texttt{halfcheetah-medium-replay-v2} &  \\
		&  &  &  & \texttt{halfcheetah-medium-v2} &  \\
		&  &  &  & \texttt{halfcheetah-medium-expert-v2} &  \\ 
		\cline{3-5}
		&  & \multirow{4}{*}{\texttt{walker2d}} & \multirow{4}{*}{\texttt{walker2d-expert-v2}} & \texttt{walker2d-random-v2} &  \\
		&  &  &  & \texttt{walker2d-medium-replay-v2} &  \\
		&  &  &  & \texttt{walker2d-medium-v2} &  \\
		&  &  &  & \texttt{walker2d-medium-expert-v2} &  \\ 
		\cline{1-5}
		\multirow{8}{*}{Adroit} & \multirow{8}{*}{\texttt{D4RL}} & \multirow{2}{*}{\texttt{pen}} & \multirow{2}{*}{\texttt{pen-expert-v1}} & \texttt{pen-cloned-v1} &  \\
		&  &  &  & \texttt{pen-human-v1} &  \\ 
		\cline{3-5}
		&  & \multirow{2}{*}{\texttt{hammer}} & \multirow{2}{*}{\texttt{hammer-expert-v1}} & \texttt{hammer-cloned-v1} &  \\
		&  &  &  & \texttt{hammer-human-v1} &  \\ 
		\cline{3-5}
		&  & \multirow{2}{*}{\texttt{door}} & \multirow{2}{*}{\texttt{door-expert-v1}} & \texttt{door-cloned-v1} &  \\
		&  &  &  & \texttt{door-human-v1} &  \\ 
		\cline{3-5}
		&  & \multirow{2}{*}{\texttt{relocate}} & \multirow{2}{*}{\texttt{relocate-expert-v1}} & \texttt{relocate-cloned-v1} &  \\
		&  &  &  & \texttt{relocate-human-v1} &  \\ 
		\cline{1-5}
		\multirow{3}{*}{AntMaze} & \multirow{3}{*}{\texttt{D4RL}} & \texttt{umaze} & \texttt{antmaze-umaze-v0} & \texttt{antmaze-umaze-diverse-v0} &  \\
		&  & \texttt{medium} & \texttt{antmaze-medium-v0} & \texttt{antmaze-medium-diverse-v0} &  \\
		&  & \texttt{large} & \texttt{antmaze-large-v0} & \texttt{antmaze-large-diverse-v0} &  \\ 
		\cline{1-5}
		\multirow{3}{*}{FrankaKitchen} & \multirow{3}{*}{\texttt{D4RL}} & \texttt{complete} & \texttt{kitchen-complete-v0} & \texttt{kitchen-mixed-v0} &  \\
		&  & \texttt{partial} & \texttt{kitchen-partial-v0} & \texttt{kitchen-mixed-v0} &  \\
		&  & \texttt{indirect} & \texttt{kitchen-partial-v0} & \texttt{kitchen-mixed-v0} &  \\ 
		\cline{1-5}
		\multirow{3}{*}{Robomimic} & \multirow{3}{*}{\texttt{robomimic}} & \texttt{lift} & \texttt{lift-proficient-human} & \texttt{lift-paired-bad} &  \\
		&  & \texttt{can-paired-bad} & \texttt{can-proficient-human} & \texttt{can-paired-bad} &  \\
		&  & \texttt{square-paired-bad} & \texttt{square-proficient-human} & \texttt{square-paired-bad} &  \\ 
		\cline{1-5}
		MuJoCo (vision) & - & - & collected by expert policies & collected by random policies &  \\
		\cline{1-5}
	\end{tabular}
	{\scriptsize \begin{tablenotes}
		\item[1] In the experiments on demonstration efficiency, we employ  \texttt{random} for MuJoCo tasks and \texttt{cloned} for Adroit tasks.
		\item[2] For vision-based MuJoCo tasks, the expert and imperfect data use video samples from a policy trained to completion with \texttt{SAC}~\citep{haarnoja18soft} and a randomly initialized policy, respectively.
	\end{tablenotes}}
	\end{threeparttable}}
\end{sidewaystable}

\subsection{Baselines}
\label{sec:baselines}


We test our method against four strong offline IL methods, all of which support the utilization of supplementary imperfect demonstrations (see \cref{sec:supp_related_work} for details). We implement them based on their publicly available implementations with the same policy network structures as ours. The tuned codes are included in the supplementary material.
\begin{enumerate}[itemsep=0pt,topsep=0pt,label=\arabic*)]
	\item \textit{{D}iscriminator-{W}eighted {B}ehavioral {C}loning} (\texttt{DWBC})~\citep{xu2022discriminator} that jointly trains a discriminator to carefully re-weight the \texttt{BC} objective (\url{https://github.com/ryanxhr/DWBC});
	\item \textit{{I}mportance-{S}ampling-{W}eighted {B}ehavioral {C}loning}  (\texttt{ISWBC})~\citep{li2023imitation} that adopts importance sampling to enhance \texttt{BC} (\url{https://github.com/liziniu/ISWBC}).
	\item \textit{{M}aximum {L}ikelihood-{I}nverse {R}einforcement {L}earning}  (\texttt{MLIRL})~\citep{zeng2023demonstrations}, a recent model-based offline IRL algorithm (\url{https://github.com/Cloud0723/Offline-MLIRL}).
	\item \textit{{C}oherent {S}oft {I}mitation {L}earning }(\texttt{CSIL})~\citep{watson2023coherent}, a model-free offline IRL method that learns a shaped reward function by entropy-regularized \texttt{BC}. The learned reward function can be exploited to anotate additional offline data and subsequently engage in offline RL, or used in finetuning the policy with further environmental interactions (\url{https://joemwatson.github.io/csil}).
\end{enumerate}
We also compare our results to that of standard \texttt{BC} and its counterpart with union offline data (running \texttt{BC} on $\mathcal{D}_o$), termed \texttt{NBCU}~\citep{li2023imitation}. During online fintuning, we continue training the policies pretrained from the offline IL methods \texttt{BC}, \texttt{NBCU}, \texttt{DWBC}, \texttt{ISWBC}, and \texttt{MLIRL} by \texttt{GAIL} (\url{https://github.com/Khrylx/PyTorch-RL}) with the hyperparameters recommended by \citet{orsini2021matters}. In addition, we also compare with another instantiation of tuning \texttt{MLIRL}, which runs \texttt{SAC} with its offline learned reward function (\cref{sec:comparision_mlirl}). However, we find the results not competitive and skip it out of our main results for brevity.

\subsection{Implementation Details}

Our method is straightforward to implement and forgiving to hyperparameters. Of note, except for the network structures in vision-based tasks (requiring the employment of CNNs), all hyperparameters are identical across tasks and settings.

For all neural nets, we adopt \texttt{ReLU} as activations, \texttt{Adam} as the optimizer, and 256 as the batchsize. For vision-based tasks, a simple CNN architecture is employed for the discriminator ($\phi_d$) and dual `variables' ($\phi_\nu$ and $\phi_y$), comprising two convolutional layers, each with a $3\times3$ convolutional kernel, $2\times2$ max pooling. For the other tasks, we represent $\phi_d$, $\phi_\nu$, $\phi_y$ as 2-layer feedforward neural networks with $256$ hidden units. The policy networks share the same structures but generate Tanh Gaussian outputs. The output of $\phi_d$ is clipped to $[0.1,0.9]$ for training stability. The learning rates for the policy, $\phi_d$, $\phi_\nu$, $\phi_y$, $D_{\phi_y,\phi_v}$ are 1e-4, 1e-5, 3e-4, 3e-4, 1e-4, respectively. The hyperparameters are summarized in \cref{table:hyperparameters}.

\begin{table}[H]
	\vskip -0.2in
	\centering
	\caption{Hyperparameters (identical across tasks).} 
	\label{table:hyperparameters}
	\vskip 0.1in
	{\small \begin{tabular}{l|l} 
		Hyperparameter & Value \\ 
		\hline
		Optimizer & \texttt{Adam} \\
		Activation function & \texttt{ReLU} \\
		Batchsize & 256 \\
		Policy learning rate (offline/online) & 1e-4 \\
		Learning rate of $\phi_d,D_{\phi_y,\phi_d}$ & 1e-5 \\
		Learning rate of $\phi_v$ and $\phi_y$ & 3e-4 \\
		Discount factor $\gamma$ & 0.99 \\
		\hline
	\end{tabular}}
	\vskip -0.15in
\end{table}

We employ the \textit{forward-KL-divergence}-based policy extraction in our experiments (the comparison between \textit{forward} and \textit{reverse} methods is included in \cref{sec:comparison_forward_reverse}). We implement our code using Pytorch 1.8.1, built upon the open-source framework of offline RL algorithms, provided at \url{https://github.com/tinkoff-ai/CORL} (under the Apache-2.0 License) and the implementation of \texttt{DWBC}, provided at \url{https://github.com/ryanxhr/DWBC} (under the MIT License).
All the experiments are run on Ubuntu 20.04.2 LTS with 8 NVIDIA GeForce RTX 4090 GPUs.

\subsection{Performance Measure}

We train a policy using 5 random seeds and evaluate it by running it in the environment for 10 episodes and computing the average undiscounted return of the environment reward. Akin to \citet{fu2020d4rl}, we use the normalized scores in figures and tables, which are measured by $\texttt{score}=\frac{\texttt{score} - \texttt{random\_score}}{\texttt{expert\_score} - \texttt{random\_score}}$ or $\texttt{score}=100\times\frac{\texttt{score} - \texttt{random\_score}}{\texttt{expert\_score} - \texttt{random\_score}}$.

\clearpage
\section{Experimental Results}
\label{sec:complete_experiments}

This section provides full experimental results to comprehensively answer the questions raised in \cref{sec:experiment}. We also provide some complementary experiments for a better understanding of the proposed method in \cref{sec:ablation}.

\subsection{Performance in Offline Imitation Learning}
\label{sec:performance_offline}


\subsubsection{Demonstration Effeciency}
\label{sec:demonstration_effeciency}

First, we evaluate \texttt{OLLIE}'s performance in offline IL with varying quantities of expert trajectories, ranging from 1 to 30 in AntMaze and MuJuCo, from 10 to 300 in Adroit and FrankaKitchen, and from 25 to 200 in vision-based MuJoCo and Robomimic. The number of imperfect trajectories is set as 1000. The sampling datasets can be found in \cref{table:dataset}. The comparative results and learning curves are shown below. 

\textbf{Summary of key findings.} \texttt{OLLIE} consistently and significantly outperforms existing methods in terms of the \textit{performance}, \textit{convergence speed}, and \textit{demonstration efficiency}, especially in challenging robotic manipulation and vision-based tasks. 

\begin{table*}[ht]
	\centering
	\caption{Normalized performance in offline IL.}
	{(\# expert trajectories: \textbf{1} in AntMaze/MuJoCo and \textbf{10} in Adroit/FrankaKitchen)}
	\vskip 0.1in
		{\small \begin{tabular}{@{}lrrrrrrr@{}}
			\toprule
			Task&\texttt{OLLIE}  &\texttt{BC}  &\texttt{NBCU}  &\texttt{CSIL}  &\texttt{DWBC}  &\texttt{MLIRL}  &\texttt{ISWBC}\\ \midrule
			\texttt{ant} &  $\bm{57.1\pm7.0}$ & $-10.7\pm11.7$ & $31.1\pm7.0$ & $2.8\pm1.3$ & $24.5\pm1.1$ & $39.3\pm9.0$ & $30.4\pm7.7$ \\
			\texttt{halfcheetah} &  $\bm{35.5\pm4.0}$ & $0.2\pm1.0$ & $2.2\pm0.1$ & $22.9\pm2.8$ & $0.0\pm0.0$ & $23.8\pm1.0$ & $13.7\pm3.0$ \\
			\texttt{hopper} &  $71.1\pm3.5$ & $17.8\pm5.4$ & $6.6\pm3.9$ & $17.1\pm6.8$ & $\bm{77.8\pm12.7}$ & $54.7\pm16.7$ & $64.7\pm9.9$ \\
			\texttt{walker2d} &  $\bm{59.8\pm8.5}$ & $4.6\pm3.9$ & $0.0\pm0.0$ & $8.1\pm5.6$ & $56.9\pm11.6$ & $49.9\pm6.7$ & $48.2\pm4.7$ \\
			\texttt{umaze} &  $\bm{75.4\pm3.8}$ & $3.5\pm3.4$ & $4.5\pm4.3$ & $11.1\pm4.9$ & $23.1\pm3.9$ & $7.8\pm3.9$ & $10.1\pm1.3$ \\
			\texttt{medium} &  $\bm{58.4\pm7.3}$ & $0.0\pm0.0$ & $0.0\pm0.0$ & $1.0\pm1.0$ & $0.0\pm0.0$ & $0.0\pm0.0$ & $7.2\pm6.7$ \\
			\texttt{large} &  $\bm{37.5\pm4.3}$ & $0.0\pm0.0$ & $0.0\pm0.0$ & $1.0\pm1.0$ & $0.0\pm0.0$ & $0.0\pm0.0$ & $5.3\pm1.9$ \\
			\texttt{hammer} &  $\bm{46.1\pm6.5}$ & $5.7\pm6.8$ & $0.0\pm0.0$ & $16.7\pm8.9$ & $6.6\pm8.3$ & $1.0\pm0.1$ & $6.5\pm5.1$ \\
			\texttt{pen} &  $\bm{67.4\pm5.6}$ & $40.3\pm10.3$ & $3.2\pm9.8$ & $39.0\pm5.6$ & $42.8\pm19.3$ & $18.7\pm2.6$ & $49.5\pm2.9$ \\
			\texttt{door} &  $\bm{28.9\pm2.9}$ & $2.8\pm3.9$ & $0.0\pm0.0$ & $24.8\pm3.6$ & $0.0\pm0.0$ & $1.0\pm0.8$ & $2.9\pm2.1$ \\
			\texttt{relocate} &  $\bm{31.2\pm5.8}$ & $0.0\pm0.0$ & $0.0\pm0.0$ & $7.7\pm8.8$ & $0.0\pm0.0$ & $1.0\pm0.2$ & $0.0\pm0.0$ \\
			\texttt{paritial} &  $\bm{34.6\pm6.6}$ & $1.0\pm0.5$ & $0.0\pm0.0$ & $26.1\pm1.4$ & $1.0\pm1.4$ & $1.0\pm0.7$ & $3.4\pm1.1$ \\
			\texttt{complete} &  $\bm{34.9\pm1.8}$ & $1.0\pm0.3$ & $0.0\pm0.0$ & $19.1\pm7.9$ & $1.0\pm0.6$ & $1.0\pm0.8$ & $3.1\pm1.9$ \\
			\texttt{undirect} &  $\bm{48.1\pm3.9}$ & $1.0\pm0.2$ & $0.0\pm0.0$ & $37.0\pm9.0$ & $-1.0\pm1.6$ & $1.0\pm0.8$ & $3.1\pm1.0$ \\ \bottomrule
		\end{tabular}}
	\label{tab:offline_d_1}
\end{table*}

\begin{table*}[ht]
	\centering
	\caption{Normalized performance in offline IL.} 
	{(\# expert trajectories: \textbf{3} in AntMaze/MuJoCo and \textbf{30} in Adroit/FrankaKitchen)}
	\vskip 0.1in
		{\small \begin{tabular}{@{}lrrrrrrr@{}}
			\toprule
			Task&\texttt{OLLIE}  &\texttt{BC}  &\texttt{NBCU}  &\texttt{CSIL}  &\texttt{DWBC}  &\texttt{MLIRL}  &\texttt{ISWBC}\\ \midrule
			\texttt{ant} &  $\bm{78.0\pm6.2}$ & $50.5\pm27.8$ & $28.9\pm3.4$ & $10.4\pm1.2$ & $32.3\pm10.1$ & $51.2\pm7.2$ & $30.2\pm2.3$ \\
			\texttt{halfcheetah} &  $\bm{79.5\pm6.1}$ & $5.7\pm1.1$ & $3.7\pm1.3$ & $20.0\pm3.8$ & $50.9\pm11.1$ & $66.6\pm3.4$ & $37.4\pm7.1$ \\
			\texttt{hopper} &  $\bm{88.7\pm6.9}$ & $47.3\pm6.7$ & $10.6\pm3.2$ & $73.9\pm3.9$ & $79.8\pm14.0$ & $74.8\pm13.6$ & $53.8\pm3.9$ \\
			\texttt{walker2d} &  $75.6\pm2.8$ & $14.8\pm2.1$ & $5.6\pm1.9$ & $17.3\pm5.4$ & $62.6\pm3.2$ & $\bm{87.2\pm11.7}$ & $60.9\pm2.2$ \\
			\texttt{umaze} &  $\bm{90.6\pm3.1}$ & $22.4\pm9.6$ & $30.4\pm12.9$ & $55.1\pm5.5$ & $51.8\pm3.0$ & $37.2\pm6.6$ & $28.6\pm2.5$ \\
			\texttt{medium} &  $\bm{72.8\pm6.2}$ & $0.0\pm0.0$ & $0.0\pm0.0$ & $32.5\pm4.5$ & $4.6\pm1.7$ & $23.5\pm6.9$ & $15.3\pm5.7$ \\
			\texttt{large} &  $\bm{48.5\pm1.4}$ & $0.0\pm0.0$ & $0.0\pm0.0$ & $19.3\pm1.5$ & $0.0\pm0.0$ & $12.3\pm3.2$ & $10.2\pm7.3$ \\
			\texttt{hammer} &  $\bm{69.5\pm8.5}$ & $1.0\pm1.0$ & $1.0\pm1.0$ & $51.1\pm9.4$ & $1.0\pm0.0$ & $1.0\pm0.0$ & $9.1\pm4.3$ \\
			\texttt{pen} &  $\bm{76.3\pm3.6}$ & $54.2\pm15.4$ & $27.1\pm6.8$ & $47.9\pm12.0$ & $55.6\pm13.7$ & $57.1\pm9.1$ & $64.3\pm8.1$ \\
			\texttt{door} &  $\bm{66.3\pm3.7}$ & $3.7\pm1.9$ & $2.9\pm1.3$ & $48.4\pm3.9$ & $2.4\pm0.3$ & $1.0\pm0.0$ & $12.8\pm2.5$ \\
			\texttt{relocate} &  $\bm{46.6\pm6.5}$ & $0.0\pm1.0$ & $0.0\pm1.0$ & $26.1\pm6.2$ & $0.0\pm0.0$ & $1.0\pm0.0$ & $8.6\pm4.0$ \\
			\texttt{paritial} &  $\bm{69.3\pm1.8}$ & $2.7\pm1.6$ & $1.0\pm0.0$ & $61.2\pm1.9$ & $3.0\pm1.7$ & $1.0\pm0.0$ & $3.8\pm1.9$ \\
			\texttt{complete} &  $\bm{60.9\pm1.8}$ & $2.0\pm1.3$ & $1.0\pm0.0$ & $45.9\pm2.5$ & $3.5\pm1.7$ & $10.0\pm0.0$ & $3.5\pm1.4$ \\
			\texttt{undirect} &  $\bm{75.1\pm1.7}$ & $2.2\pm1.8$ & $1.0\pm0.0$ & $62.4\pm8.5$ & $3.1\pm1.9$ & $1.0\pm0.0$ & $5.1\pm1.4$ \\ \bottomrule
		\end{tabular}}
	\label{tab:offline_d_3}
\end{table*}

\begin{table*}[ht]
	\centering
	\caption{Normalized performance in offline IL.} 
	{(\# expert trajectories: \textbf{5} in AntMaze/MuJoCo and \textbf{50} in Adroit/FrankaKitchen)}
	\vskip 0.1in
		{\small \begin{tabular}{@{}lrrrrrrr@{}}
			\toprule
			Task&\texttt{OLLIE}  &\texttt{BC}  &\texttt{NBCU}  &\texttt{CSIL}  &\texttt{DWBC}  &\texttt{MLIRL}  &\texttt{ISWBC}\\ \midrule
			\texttt{ant} &  $\bm{105.2\pm4.3}$ & $61.9\pm3.2$ & $58.8\pm2.5$ & $61.5\pm4.0$ & $80.3\pm2.8$ & $57.3\pm4.1$ & $5.2\pm6.8$ \\
			\texttt{halfcheetah} &  $\bm{82.0\pm9.2}$ & $31.6\pm8.3$ & $34.6\pm6.5$ & $6.2\pm2.0$ & $48.2\pm5.6$ & $83.5\pm1.1$ & $53.5\pm2.7$ \\
			\texttt{hopper} &  $\bm{97.4\pm4.9}$ & $39.5\pm5.7$ & $25.7\pm7.9$ & $88.3\pm5.4$ & $75.8\pm7.6$ & $79.7\pm8.5$ & $64.1\pm3.3$ \\
			\texttt{walker2d} &  $\bm{94.6\pm2.5}$ & $16.9\pm3.6$ & $17.9\pm7.5$ & $69.5\pm6.1$ & $67.1\pm7.7$ & $88.7\pm4.9$ & $61.6\pm3.4$ \\
			\texttt{umaze} &  $\bm{100.0\pm0.0}$ & $29.6\pm5.5$ & $33.8\pm7.0$ & $82.3\pm5.4$ & $56.2\pm5.1$ & $42.4\pm3.1$ & $38.9\pm5.8$ \\
			\texttt{medium} &  $\bm{95.0\pm5.0}$ & $1.0\pm1.0$ & $1.0\pm1.0$ & $52.1\pm4.2$ & $13.3\pm3.7$ & $28.8\pm5.0$ & $28.0\pm4.2$ \\
			\texttt{large} &  $\bm{80.7\pm7.8}$ & $1.0\pm1.0$ & $1.0\pm1.0$ & $38.4\pm4.9$ & $7.7\pm6.9$ & $15.3\pm5.7$ & $16.9\pm3.6$ \\
			\texttt{hammer} &  $\bm{79.3\pm9.5}$ & $1.0\pm1.0$ & $1.0\pm1.0$ & $61.4\pm3.8$ & $1.0\pm1.0$ & $1.0\pm1.0$ & $23.2\pm5.9$ \\
			\texttt{pen} &  $\bm{86.6\pm2.9}$ & $50.5\pm7.5$ & $26.0\pm2.5$ & $63.0\pm5.8$ & $80.8\pm5.0$ & $57.4\pm4.8$ & $63.8\pm3.5$ \\
			\texttt{door} &  $\bm{79.4\pm4.9}$ & $1.0\pm1.0$ & $1.0\pm1.0$ & $56.3\pm7.2$ & $1.0\pm1.0$ & $1.0\pm1.0$ & $23.2\pm2.5$ \\
			\texttt{relocate} &  $\bm{69.4\pm6.7}$ & $10.0\pm1.0$ & $1.0\pm1.0$ & $6.7\pm7.7$ & $1.0\pm1.0$ & $1.0\pm1.0$ & $29.7\pm4.3$ \\
			\texttt{paritial} &  $\bm{76.3\pm3.7}$ & $4.1\pm6.4$ & $1.0\pm1.0$ & $74.3\pm7.9$ & $1.3\pm79.0$ & $1.0\pm1.0$ & $1.0\pm3.3$ \\
			\texttt{complete} &  $\bm{64.2\pm4.1}$ & $3.3\pm5.2$ & $1.0\pm1.0$ & $51.4\pm5.9$ & $9.4\pm4.6$ & $1.0\pm1.0$ & $4.1\pm6.1$ \\
			\texttt{undirect} &  $\bm{74.3\pm3.4}$ & $8.6\pm3.3$ & $1.0\pm1.0$ & $84.2\pm7.3$ & $12.5\pm4.8$ & $1.0\pm1.0$ & $16.6\pm6.1$ \\ \bottomrule
		\end{tabular}}
	\label{tab:offline_d_5}
\end{table*}

\begin{table*}[ht]
	\centering
	\caption{Normalized performance in offline IL.} 
	{(\# expert trajectories: \textbf{10} in AntMaze/MuJoCo and \textbf{100} in Adroit/FrankaKitchen)}
	\vskip 0.1in
		{\small \begin{tabular}{@{}lrrrrrrr@{}}
			\toprule
			Task&\texttt{OLLIE}  &\texttt{BC}  &\texttt{NBCU}  &\texttt{CSIL}  &\texttt{DWBC}  &\texttt{MLIRL}  &\texttt{ISWBC}\\ \midrule
			\texttt{ant} &  $\bm{115.7\pm4.3}$ & $64.6\pm6.0$ & $58.9\pm16.7$ & $83.1\pm6.1$ & $88.7\pm5.5$ & $65.7\pm4.1$ & $79.4\pm7.5$ \\
			\texttt{halfcheetah} &  $\bm{90.2\pm10.3}$ & $61.5\pm4.0$ & $3.5\pm1.9$ & $77.8\pm7.4$ & $58.9\pm3.6$ & $82.9\pm3.9$ & $77.1\pm3.0$ \\
			\texttt{hopper} &  $\bm{99.2\pm3.9}$ & $61.8\pm19.6$ & $25.7\pm2.6$ & $95.4\pm3.4$ & $69.1\pm3.1$ & $89.2\pm6.2$ & $81.6\pm8.4$ \\
			\texttt{walker2d} &  $\bm{95.9\pm3.8}$ & $17.6\pm2.4$ & $17.6\pm3.4$ & $91.4\pm9.0$ & $71.3\pm9.8$ & $92.0\pm7.0$ & $62.0\pm9.0$ \\
			\texttt{umaze} &  $\bm{100.0\pm0.0}$ & $46.1\pm3.9$ & $40.4\pm9.9$ & $94.1\pm6.2$ & $69.9\pm3.2$ & $55.1\pm13.7$ & $54.3\pm6.3$ \\
			\texttt{medium} &  $\bm{95.0\pm5.0}$ & $0.0\pm0.0$ & $0.0\pm0.0$ & $60.9\pm3.3$ & $33.4\pm7.6$ & $41.2\pm7.3$ & $48.3\pm9.3$ \\
			\texttt{large} &  $\bm{85.0\pm10.0}$ & $0.0\pm0.0$ & $0.0\pm0.0$ & $46.2\pm6.8$ & $25.5\pm5.6$ & $21.4\pm2.8$ & $25.2\pm4.1$ \\
			\texttt{hammer} &  $\bm{82.6\pm9.8}$ & $41.9\pm1.5$ & $1.0\pm1.0$ & $66.8\pm5.3$ & $70.6\pm3.9$ & $1.0\pm0.0$ & $43.2\pm7.6$ \\
			\texttt{pen} &  $\bm{94.6\pm6.6}$ & $56.5\pm7.6$ & $24.0\pm1.0$ & $70.0\pm11.8$ & $69.7\pm2.3$ & $58.6\pm5.1$ & $62.6\pm5.0$ \\
			\texttt{door} &  $\bm{92.0\pm1.8}$ & $26.6\pm6.2$ & $0.0\pm1.0$ & $59.7\pm7.7$ & $20.5\pm6.0$ & $1.0\pm0.0$ & $39.6\pm3.3$ \\
			\texttt{relocate} &  $\bm{92.1\pm8.5}$ & $35.0\pm4.2$ & $0.0\pm1.0$ & $74.6\pm4.3$ & $38.6\pm9.8$ & $1.0\pm0.0$ & $41.2\pm7.2$ \\
			\texttt{paritial} &  $\bm{99.1\pm1.8}$ & $10.4\pm0.1$ & $3.4\pm1.0$ & $79.0\pm5.7$ & $27.3\pm5.7$ & $1.0\pm0.0$ & $20.0\pm2.2$ \\
			\texttt{complete} &  $\bm{98.6\pm2.8}$ & $6.1\pm7.5$ & $3.1\pm1.0$ & $54.9\pm5.8$ & $23.6\pm1.0$ & $1.0\pm0.0$ & $5.9\pm5.7$ \\
			\texttt{undirect} &  $\bm{100.0\pm0.0}$ & $22.7\pm5.3$ & $3.5\pm1.4$ & $93.6\pm7.4$ & $34.9\pm7.5$ & $1.0\pm0.0$ & $32.6\pm8.7$ \\ \bottomrule
		\end{tabular}}
	\label{tab:offline_d_10}
\end{table*}

\begin{table*}[ht]
	\centering
	\caption{Normalized performance in offline IL.} 
	{(\# expert trajectories: \textbf{30} in AntMaze/MuJoCo and \textbf{300} in Adroit/FrankaKitchen)}
	\vskip 0.1in
		{\small \begin{tabular}{@{}lrrrrrrr@{}}
			\toprule
			Task&\texttt{OLLIE}  &\texttt{BC}  &\texttt{NBCU}  &\texttt{CSIL}  &\texttt{DWBC}  &\texttt{MLIRL}  &\texttt{ISWBC}\\ \midrule
			\texttt{ant} &  $\bm{117.5\pm3.6}$ & $100.6\pm1.1$ & $100.7\pm2.8$ & $106.8\pm0.9$ & $91.3\pm5.4$ & $107.1\pm5.6$ & $105.6\pm8.6$ \\
			\texttt{halfcheetah} &  $\bm{100.6\pm0.8}$ & $86.4\pm3.3$ & $3.2\pm1.9$ & $102.1\pm2.5$ & $89.4\pm4.2$ & $93.6\pm5.8$ & $101.3\pm3.3$ \\
			\texttt{hopper} &  $\bm{108.7\pm1.2}$ & $56.8\pm17.9$ & $31.9\pm10.4$ & $94.0\pm3.7$ & $107.7\pm2.2$ & $103.9\pm7.2$ & $99.1\pm2.6$ \\
			\texttt{walker2d} &  $\bm{105.7\pm2.9}$ & $41.1\pm21.3$ & $26.5\pm11.6$ & $101.5\pm1.4$ & $97.8\pm3.6$ & $92.9\pm13.4$ & $92.0\pm3.4$ \\
			\texttt{umaze} &  $\bm{100.0\pm0.0}$ & $65.0\pm30.2$ & $55.5\pm25.2$ & ${100.0\pm0.0}$ & $90.0\pm5.0$ & ${100.0\pm0.0}$ & ${100.0\pm0.0}$ \\
			\texttt{medium} &  $\bm{100.0\pm0.0}$ & $0.0\pm0.0$ & $0.0\pm0.0$ & $99.0\pm1.0$ & $47.0\pm3.0$ & $71.5\pm4.2$ & $91.0\pm3.0$ \\
			\texttt{large} &  ${100.0\pm0.0}$ & $0.0\pm0.0$ & $0.0\pm0.0$ & $79.9\pm6.6$ & $40.3\pm3.2$ & $65.1\pm3.2$ & $73.0\pm4.0$ \\
			\texttt{hammer} &  ${100.7\pm9.3}$ & $\bm{105.8\pm5.1}$ & $0.0\pm1.0$ & $100.0\pm3.1$ & $93.8\pm4.6$ & $53.1\pm5.9$ & ${102.4\pm2.8}$ \\
			\texttt{pen} &  $\bm{99.3\pm3.7}$ & $85.9\pm9.3$ & $57.0\pm4.8$ & $94.0\pm6.5$ & $98.9\pm2.0$ & $77.5\pm9.6$ & $92.2\pm2.7$ \\
			\texttt{door} &  $\bm{108.5\pm1.3}$ & $34.0\pm5.6$ & $0.0\pm1.0$ & $105.7\pm4.3$ & $71.3\pm9.4$ & $41.5\pm7.2$ & $65.2\pm4.5$ \\
			\texttt{relocate} &  ${102.7\pm5.3}$ & $101.7\pm2.2$ & $0.0\pm1.0$ & $98.6\pm8.5$ & $100.9\pm0.1$ & $82.6\pm3.8$ & $\bm{104.8\pm5.9}$ \\
			\texttt{paritial} &  $\bm{100.0\pm0.0}$ & $34.7\pm2.3$ & $13.4\pm2.3$ & $\bm{100.0\pm0.0}$ & $58.1\pm5.3$ & $1.0\pm0.0$ & $39.9\pm4.7$ \\
			\texttt{complete} &  $\bm{100.0\pm0.0}$ & $21.6\pm3.8$ & $10.7\pm1.3$ & $85.0\pm7.9$ & $32.3\pm3.9$ & $1.0\pm0.0$ & $28.8\pm6.0$ \\
			\texttt{undirect} &  $\bm{100.0\pm0.0}$ & $48.7\pm4.0$ & $16.0\pm3.7$ & ${100.0\pm0.0}$ & $63.7\pm1.0$ & $1.0\pm0.0$ & $58.1\pm3.3$ \\ \bottomrule
		\end{tabular}}
	\label{tab:offline_d_30}
\end{table*}

\begin{table*}[ht]
	\centering
	\caption{Normalized performance in offline IL across vision-based tasks.} 
	{(\# expert trajectories: \textbf{25})}
	\vskip 0.1in
		{\small \begin{tabular}{@{}lrrrrrrr@{}}
			\toprule
			Task&\texttt{OLLIE}  &\texttt{BC}  &\texttt{NBCU}  &\texttt{CSIL}  &\texttt{DWBC}  &\texttt{MLIRL}  &\texttt{ISWBC}\\ \midrule
			\texttt{ant} &  $\bm{30.8\pm5.1}$ & $16.1\pm5.5$ & $16.3\pm3.7$ & $11.6\pm3.6$ & $18.2\pm4.3$ & $0.1\pm1.0$ & $22.4\pm3.1$ \\
			\texttt{halfcheetah} &  $\bm{41.9\pm3.1}$ & $27.3\pm4.7$ & $29.5\pm6.8$ & $26.3\pm5.8$ & $19.7\pm8.2$ & $0.1\pm1.0$ & $26.0\pm2.9$ \\
			\texttt{hopper} &  $\bm{57.2\pm2.3}$ & $13.3\pm5.9$ & $12.8\pm7.2$ & $12.7\pm5.3$ & $17.6\pm3.9$ & $0.1\pm1.0$ & $16.1\pm8.4$ \\
			\texttt{walker2d} &  $\bm{53.3\pm2.6}$ & $10.3\pm3.2$ & $8.0\pm8.0$ & $9.0\pm8.2$ & $26.5\pm7.7$ & $0.1\pm1.0$ & $28.4\pm4.7$ \\
			\texttt{lift} &  $\bm{84.6\pm6.9}$ & $49.0\pm6.6$ & $29.9\pm4.2$ & $53.7\pm3.4$ & $47.7\pm7.8$ & $0.1\pm1.0$ & $57.1\pm3.1$ \\
			\texttt{can} &  $\bm{39.8\pm7.8}$ & $14.1\pm12.2$ & $22.3\pm3.9$ & $24.4\pm4.8$ & $25.0\pm2.1$ & $0.1\pm1.0$ & $10.7\pm15.8$ \\
			\texttt{square} &  $\bm{36.1\pm8.0}$ & $2.1\pm2.1$ & $5.5\pm5.0$ & $5.2\pm4.3$ & $17.1\pm4.5$ & $0.1\pm1.0$ & $15.3\pm2.3$ \\ \bottomrule
		\end{tabular}}
	\label{tab:offline_d_25}
\end{table*}

\begin{table*}[ht]
	\centering
	\caption{Normalized performance in offline IL across vision-based tasks.} 
	{(\# expert trajectories: \textbf{50})}
	\vskip 0.1in
		{\small \begin{tabular}{@{}lrrrrrrr@{}}
			\toprule
			Task&\texttt{OLLIE}  &\texttt{BC}  &\texttt{NBCU}  &\texttt{CSIL}  &\texttt{DWBC}  &\texttt{MLIRL}  &\texttt{ISWBC}\\ \midrule
			\texttt{ant} &  $\bm{58.6\pm7.2}$ & $26.8\pm9.7$ & $25.4\pm3.9$ & $20.5\pm6.8$ & $28.6\pm3.5$ & $0.0\pm1.0$ & $32.7\pm4.4$ \\
			\texttt{halfcheetah} &  $\bm{61.7\pm6.7}$ & $42.7\pm7.2$ & $23.8\pm4.7$ & $19.7\pm5.2$ & $26.6\pm6.6$ & $0.0\pm1.0$ & $35.3\pm5.8$ \\
			\texttt{hopper} &  $\bm{85.2\pm6.4}$ & $21.3\pm8.4$ & $13.7\pm9.3$ & $18.8\pm5.7$ & $16.9\pm3.2$ & $0.0\pm1.0$ & $23.5\pm4.7$ \\
			\texttt{walker2d} &  $\bm{64.5\pm8.6}$ & $22.0\pm6.5$ & $15.5\pm5.5$ & $24.6\pm9.4$ & $27.6\pm5.0$ & $0.0\pm1.0$ & $38.3\pm5.5$ \\
			\texttt{lift} &  $\bm{95.0\pm5.0}$ & $75.6\pm8.2$ & $63.3\pm7.0$ & $62.4\pm2.8$ & $70.1\pm8.8$ & $0.0\pm0.0$ & $77.2\pm5.3$ \\
			\texttt{can} &  $\bm{61.3\pm5.3}$ & $25.0\pm11.2$ & $23.9\pm4.5$ & $38.9\pm2.5$ & $28.6\pm9.5$ & $0.0\pm0.0$ & $32.8\pm3.8$ \\
			\texttt{square} &  $\bm{41.3\pm8.3}$ & $17.0\pm7.4$ & $15.9\pm4.9$ & $17.4\pm3.2$ & $18.3\pm5.3$ & $0.0\pm0.0$ & $24.5\pm5.1$ \\ \bottomrule
		\end{tabular}}
	\label{tab:offline_d_50}
\end{table*}

\begin{table*}[ht]
	\centering
	\caption{Normalized performance in offline IL across vision-based tasks.} 
	{(\# expert trajectories: \textbf{100})}
	\vskip 0.1in
		{\small \begin{tabular}{@{}lrrrrrrr@{}}
			\toprule
			Task&\texttt{OLLIE}  &\texttt{BC}  &\texttt{NBCU}  &\texttt{CSIL}  &\texttt{DWBC}  &\texttt{MLIRL}  &\texttt{ISWBC}\\ \midrule
			\texttt{ant} &  $\bm{71.1\pm3.9}$ & $43.7\pm2.7$ & $38.0\pm6.0$ & $43.9\pm4.6$ & $26.0\pm3.8$ & $0.0\pm1.0$ & $45.4\pm9.6$ \\
			\texttt{halfcheetah} &  $\bm{75.8\pm4.3}$ & $50.9\pm5.4$ & $38.0\pm2.7$ & $44.0\pm5.5$ & $18.2\pm2.7$ & $0.0\pm1.0$ & $55.3\pm4.6$ \\
			\texttt{hopper} &  $\bm{91.7\pm2.7}$ & $31.2\pm2.6$ & $30.8\pm6.8$ & $29.4\pm7.9$ & $23.6\pm9.0$ & $0.0\pm1.0$ & $33.6\pm2.1$ \\
			\texttt{walker2d} &  $\bm{82.1\pm8.1}$ & $39.4\pm6.9$ & $40.2\pm5.0$ & $49.1\pm7.2$ & $35.9\pm2.7$ & $0.0\pm1.0$ & $48.2\pm5.4$ \\
			\texttt{lift} &  $\bm{99.0\pm1.0}$ & $89.0\pm11.0$ & $91.0\pm9.0$ & $62.9\pm2.5$ & $94.0\pm6.0$ & $0.0\pm0.0$ & $95.0\pm5.0$ \\
			\texttt{can} &  $\bm{78.0\pm7.5}$ & $49.7\pm5.2$ & $58.8\pm8.7$ & $51.6\pm1.4$ & $56.2\pm3.9$ & $0.0\pm0.0$ & $55.8\pm2.9$ \\
			\texttt{square} &  $\bm{64.1\pm6.4}$ & $31.9\pm6.9$ & $24.3\pm4.7$ & $24.9\pm3.6$ & $22.6\pm5.3$ & $0.0\pm0.0$ & $27.8\pm8.6$ \\ \bottomrule
		\end{tabular}}
	\label{tab:offline_d_100}
\end{table*}

\begin{table*}[ht]
	\centering
	\caption{Normalized performance in offline IL across vision-based tasks.} 
	{(\# expert trajectories: \textbf{200})}
	\vskip 0.1in
	{\small 	\begin{tabular}{@{}lrrrrrrr@{}}
			\toprule
			Task&\texttt{OLLIE}  &\texttt{BC}  &\texttt{NBCU}  &\texttt{CSIL}  &\texttt{DWBC}  &\texttt{MLIRL}  &\texttt{ISWBC}\\ \midrule
			\texttt{ant} &  $\bm{82.0\pm4.5}$ & $48.5\pm5.3$ & $57.0\pm5.1$ & $57.8\pm3.3$ & $15.0\pm4.2$ & $0.0\pm1.0$ & $59.6\pm4.2$ \\
			\texttt{halfcheetah} &  $\bm{93.9\pm7.8}$ & $54.6\pm6.1$ & $58.7\pm6.0$ & $59.1\pm5.5$ & $34.6\pm7.1$ & $0.0\pm1.0$ & $60.7\pm4.8$ \\
			\texttt{hopper} &  $\bm{98.0\pm3.3}$ & $56.6\pm2.4$ & $49.1\pm9.1$ & $45.0\pm2.9$ & $12.5\pm7.2$ & $0.0\pm1.0$ & $47.7\pm2.8$ \\
			\texttt{walker2d} &  $\bm{90.3\pm5.9}$ & $43.5\pm5.4$ & $73.1\pm6.1$ & $71.0\pm4.4$ & $10.1\pm8.4$ & $0.0\pm1.0$ & $71.4\pm6.3$ \\
			\texttt{lift} &  $\bm{100.0\pm0.0}$ & $91.0\pm9.0$ & $97.0\pm3.0$ & $63.1\pm2.0$ & $95.0\pm5.0$ & $0.0\pm0.0$ & $94.0\pm6.0$ \\
			\texttt{can} &  $\bm{94.0\pm6.0}$ & $66.8\pm11.1$ & $64.8\pm4.8$ & $57.6\pm3.4$ & $73.6\pm4.2$ & $0.0\pm0.0$ & $83.9\pm5.0$ \\
			\texttt{square} &  $\bm{85.2\pm4.4}$ & $37.5\pm12.7$ & $33.7\pm6.9$ & $35.5\pm3.1$ & $16.5\pm1.6$ & $0.0\pm0.0$ & $34.9\pm4.4$ \\ \bottomrule
		\end{tabular}}
	\label{tab:offline_d_200}
\end{table*}

\clearpage

\begin{figure*}[t]
	\centering
	{\includegraphics[width=\textwidth]{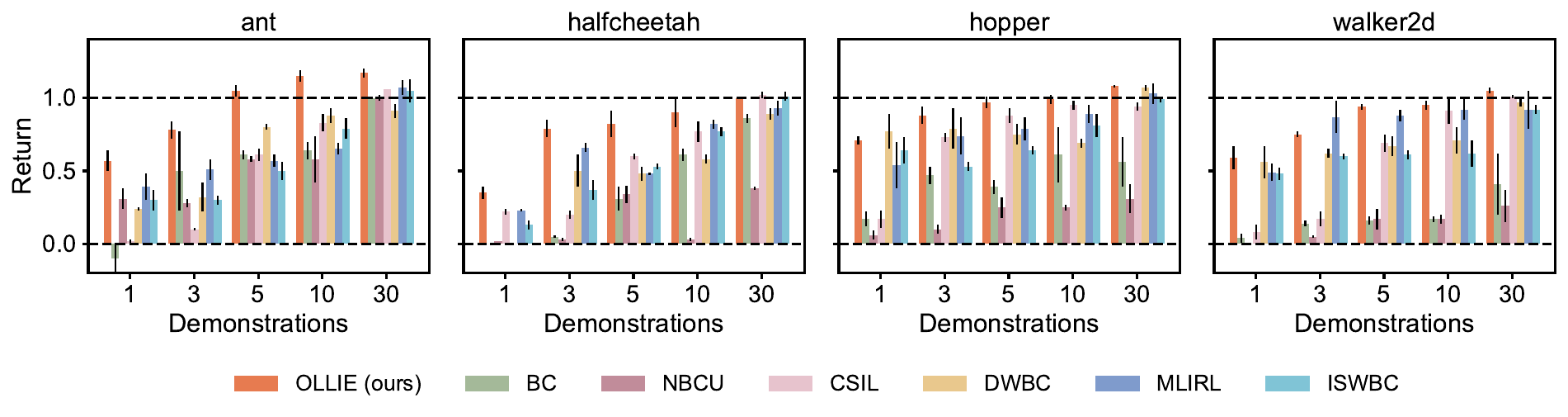}} 
	\vskip -0.1in
	\caption{Performance in offline IL with varying quantities of expert trajectories in \textbf{\textit{MuJoCo}}. Uncertainty intervals depict standard deviation over five seeds. \texttt{OLLIE} uses fewer expert demonstrations to attain expert performance, demonstrating its great demonstration efficiency compared to existing methods.}
	\label{fig:performance_mujoco}
\end{figure*}

\begin{figure*}[t]
	\centering
	{\includegraphics[width=\textwidth]{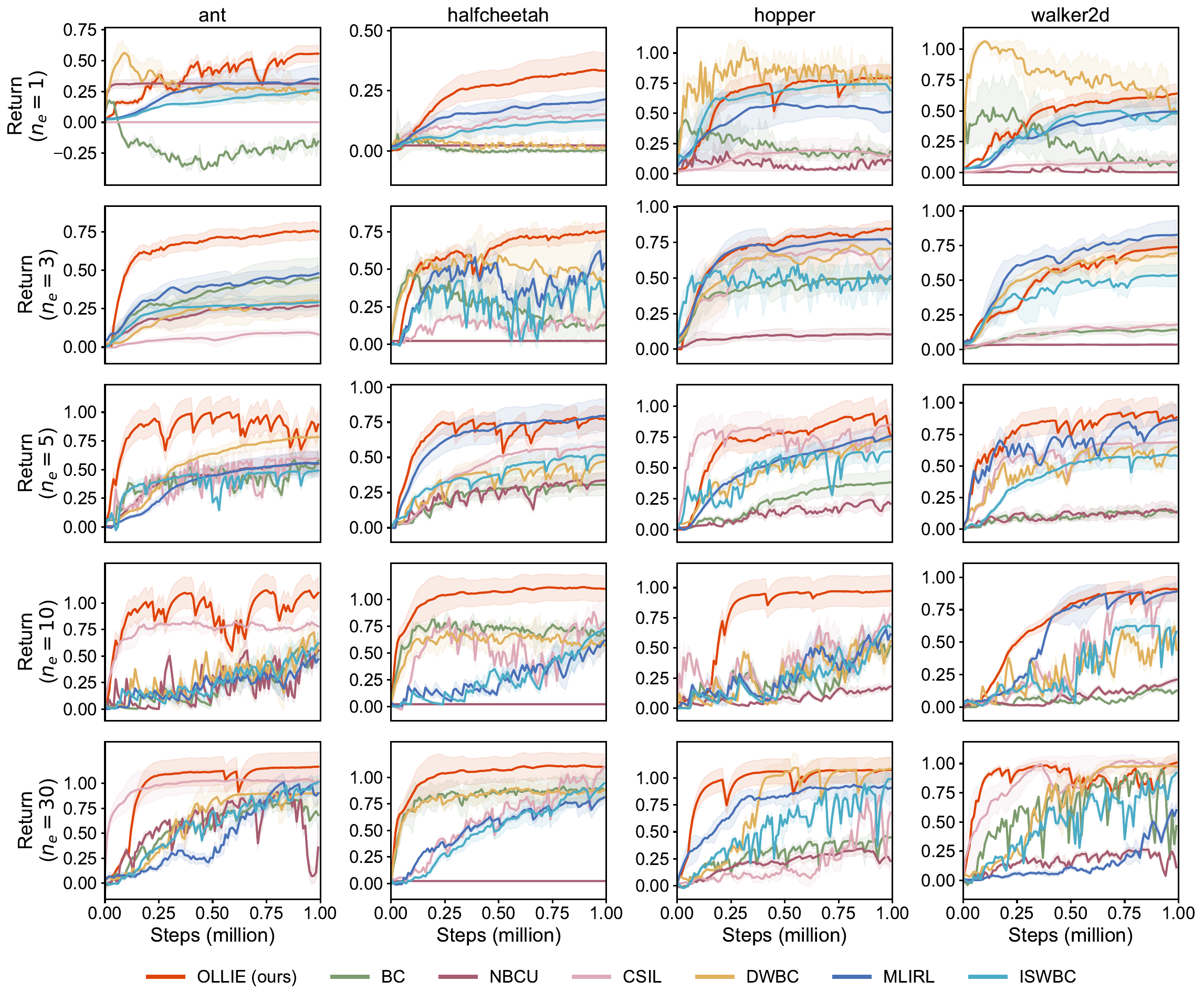}}
	\vskip -0.1in
	\caption{Learning curves in offline IL with varying quantities of expert trajectories in \textbf{\textit{MuJoCo}}. Uncertainty intervals depict standard deviation over five seeds. $n_e$ represents expert trajectories. \texttt{OLLIE} consistently exhibits fast and stabilized convergence.}
	\label{fig:curve_mujoco}
\end{figure*}

\clearpage

\begin{figure*}[t]
	\centering
	{\includegraphics[width=\textwidth]{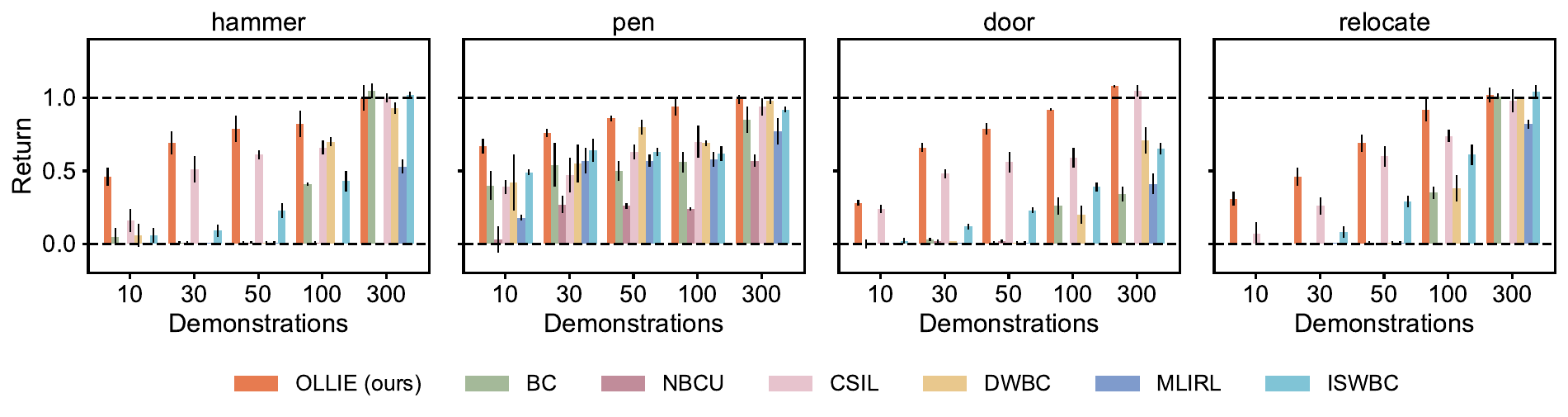}}
	\vskip -0.1in
	\caption{Performance in offline IL with varying quantities of expert trajectories in \textbf{\textit{Adroit}}. Uncertainty intervals depict standard deviation over five seeds. \texttt{OLLIE} uses fewer expert demonstrations to attain expert performance, demonstrating its great demonstration efficiency in comparison with existing methods.}
	\label{fig:performance_adroit}
\end{figure*}

\begin{figure*}[t]
	\centering
	{\includegraphics[width=\textwidth]{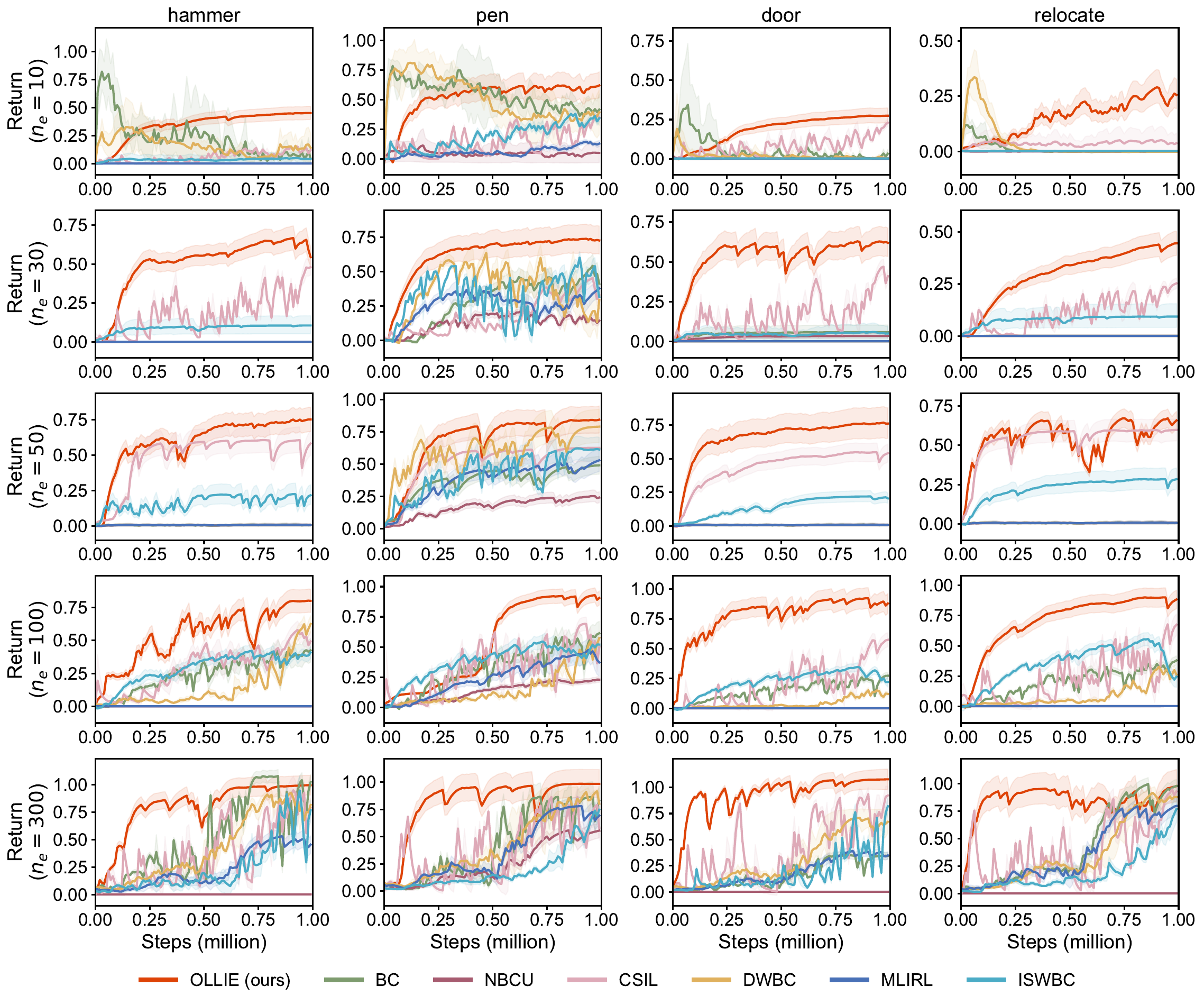}}
	\vskip -0.1in
	\caption{Learning curves in offline IL with varying quantities of expert trajectories in \textbf{\textit{Adroit}}. Uncertainty intervals depict standard deviation over five seeds. $n_e$ represents expert trajectories. \texttt{OLLIE} consistently exhibits fast and stabilized convergence.}
	\label{fig:curve_adroit}
\end{figure*}

\clearpage
\begin{figure*}[t]
	\centering
	{\includegraphics[width=0.825\textwidth]{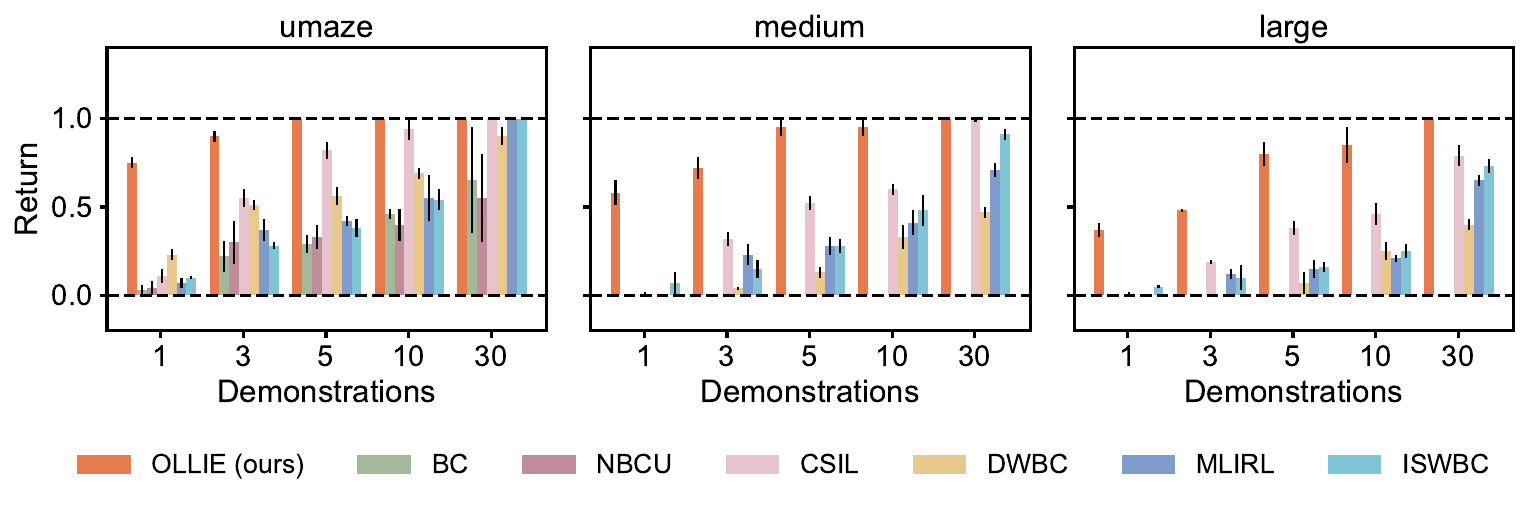}}
	\vskip -0.1in		
	\caption{Performance in offline IL with varying quantities of expert trajectories in \textbf{\textit{AntMaze}}. Uncertainty intervals depict standard deviation over five seeds. $n_e$ represents expert trajectories. \texttt{OLLIE} uses much fewer expert demonstrations to attain expert performance, demonstrating its great demonstration efficiency in comparison with existing methods. Of note, AntMaze is challenging because it requires precise long-horizon control. The outperformance of \texttt{OLLIE} reveals its capability of extracting environmental information from offline data.}
	\label{fig:performance_antmaze}
\end{figure*}

\begin{figure*}[t]
	\centering
	{\includegraphics[width=0.85\textwidth]{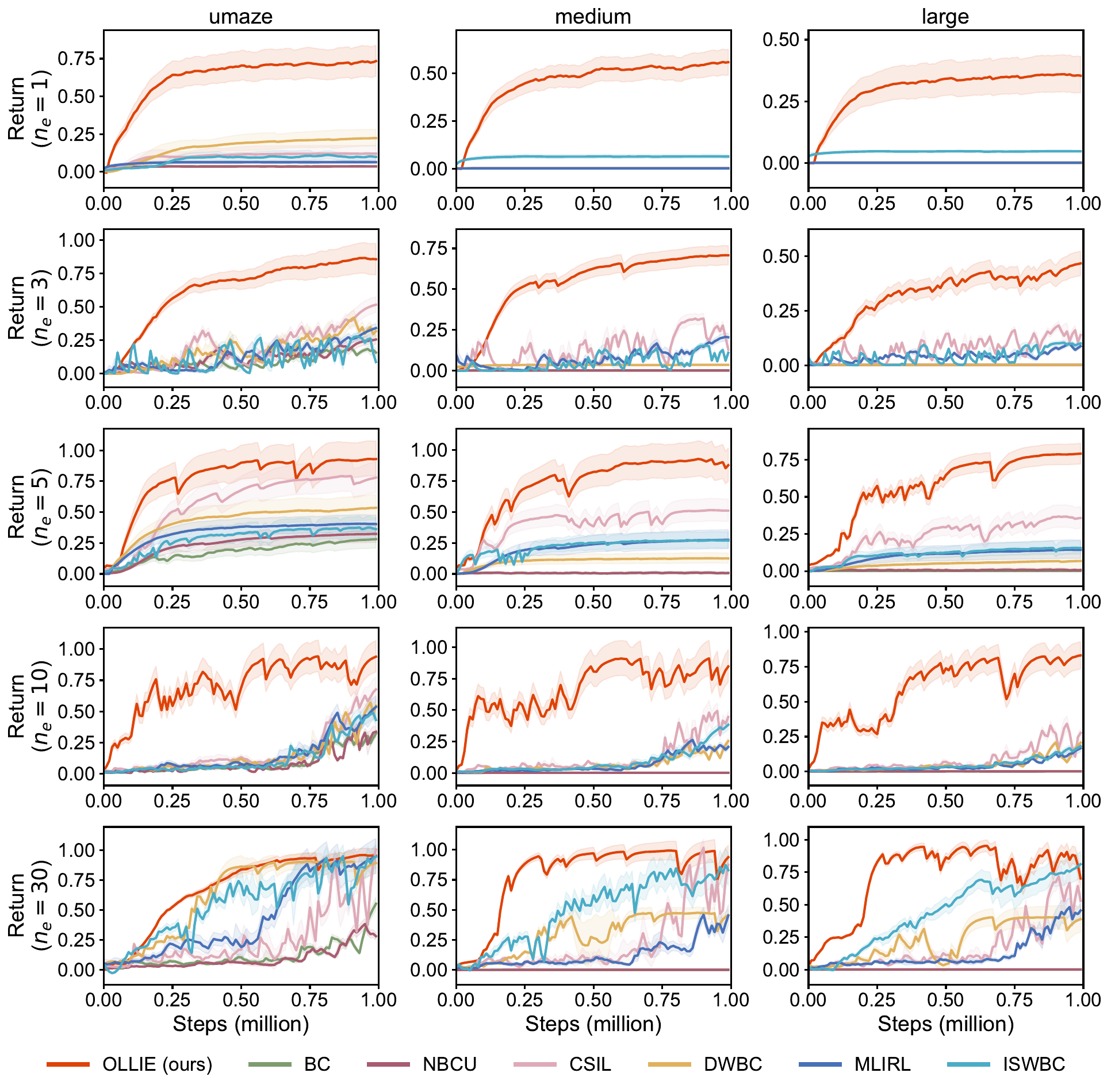}}
	\vskip -0.1in
	\caption{Learning curves in offline IL with varying quantities of expert trajectories in \textbf{\textit{AntMaze}}. Uncertainty intervals depict standard deviation over five seeds. $n_e$ represents expert trajectories. \texttt{OLLIE} consistently and significantly surpasses existing methods in terms of convergence speed and stability.}
	\label{fig:curve_antmaze}
\end{figure*}

\clearpage
\begin{figure*}[t]
	\centering
	{\includegraphics[width=0.825\textwidth]{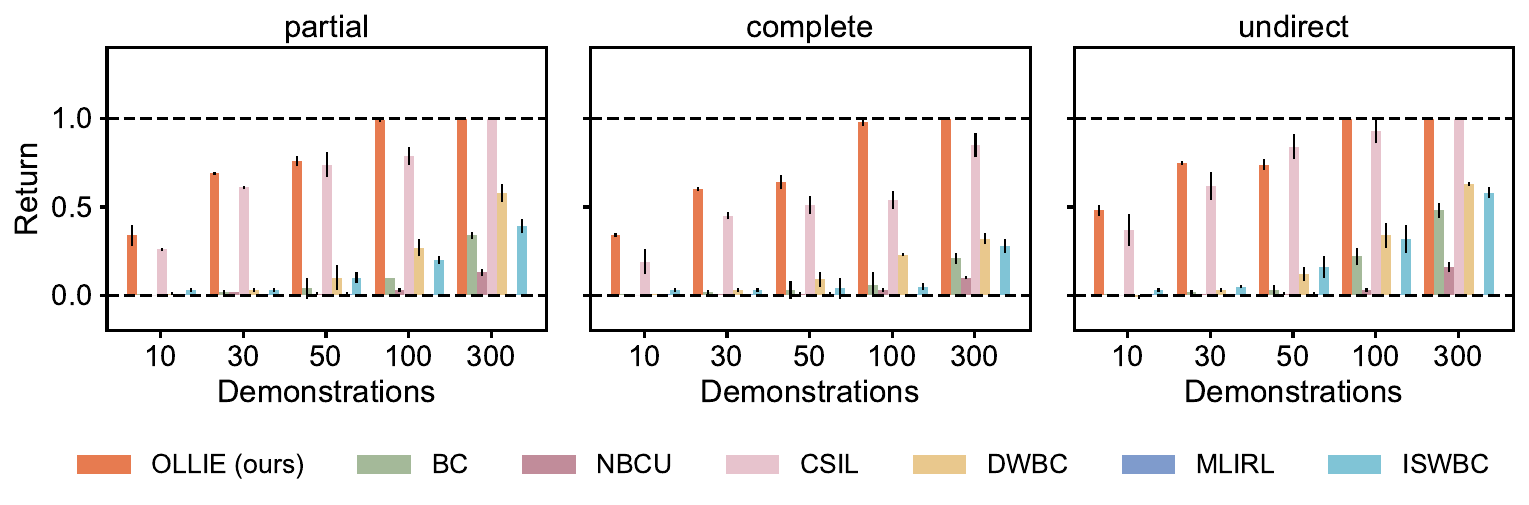}}
	\vskip -0.1in
	\caption{Performance in offline IL with varying quantities of expert trajectories in \textbf{\textit{FrankaKitchen}}. Uncertainty intervals depict standard deviation over five seeds. \texttt{OLLIE} uses fewer expert demonstrations to attain expert performance, demonstrating its great demonstration efficiency in comparison with existing methods.}
	\label{fig:performance_kitchen}
\end{figure*}

\begin{figure*}[t]
	\centering
	{\includegraphics[width=0.85\textwidth]{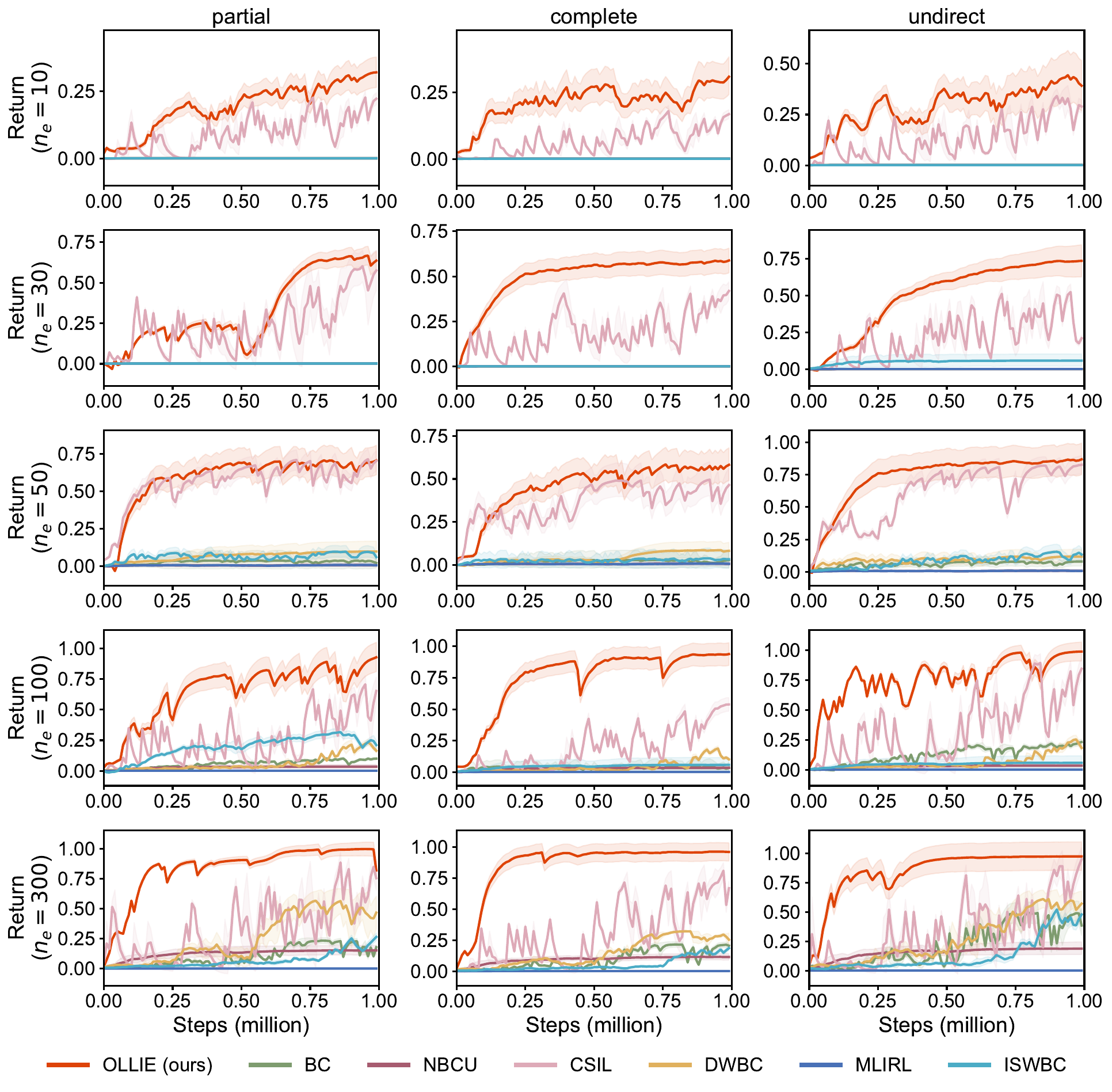}}
	\vskip -0.1in
	\caption{Learning curves in offline IL with varying quantities of expert trajectories in \textbf{\textit{FrankaKitchen}}. Uncertainty intervals depict standard deviation over five seeds. $n_e$ represents expert trajectories. \texttt{OLLIE} consistently exhibits fast and stabilized convergence. \texttt{CSIL} also works well in this domain.}
	\label{fig:curve_kitchen}
\end{figure*}

\clearpage
\begin{figure*}[t]
	\centering
	{\includegraphics[width=\textwidth]{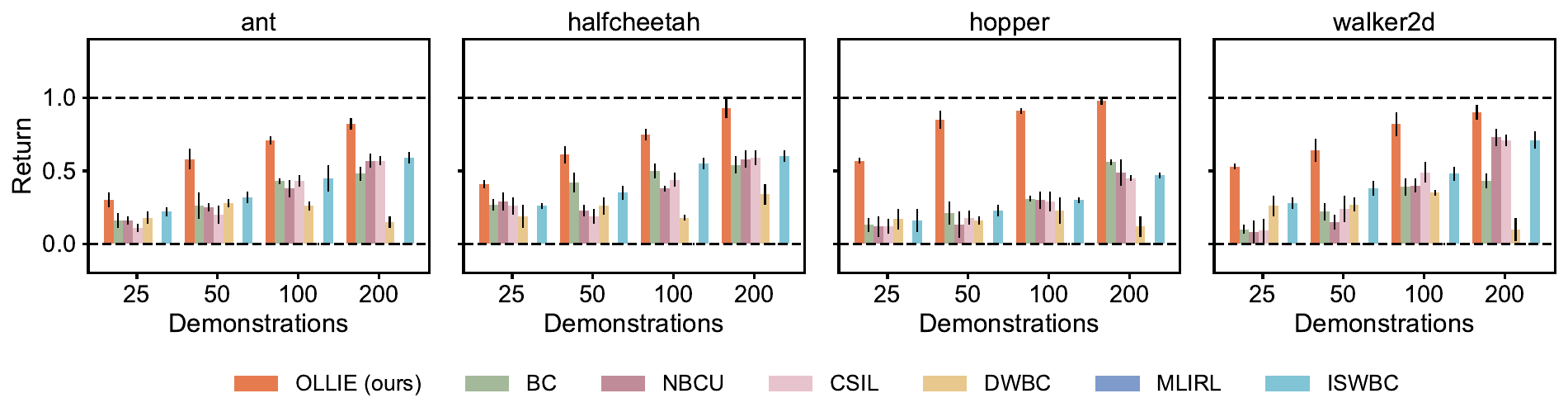}}
	\vskip -0.1in
	\caption{Performance in offline IL with varying quantities of expert trajectories in \textbf{\textit{vision-based MuJoCo}}. Uncertainty intervals depict standard deviation over five seeds. \texttt{OLLIE} uses much fewer expert demonstrations to attain expert performance, demonstrating its great demonstration efficiency in high-dimensional environments.}
	\label{fig:performance_mujoco_image}
\end{figure*}

\begin{figure*}[t]
	\centering
	{\includegraphics[width=\textwidth]{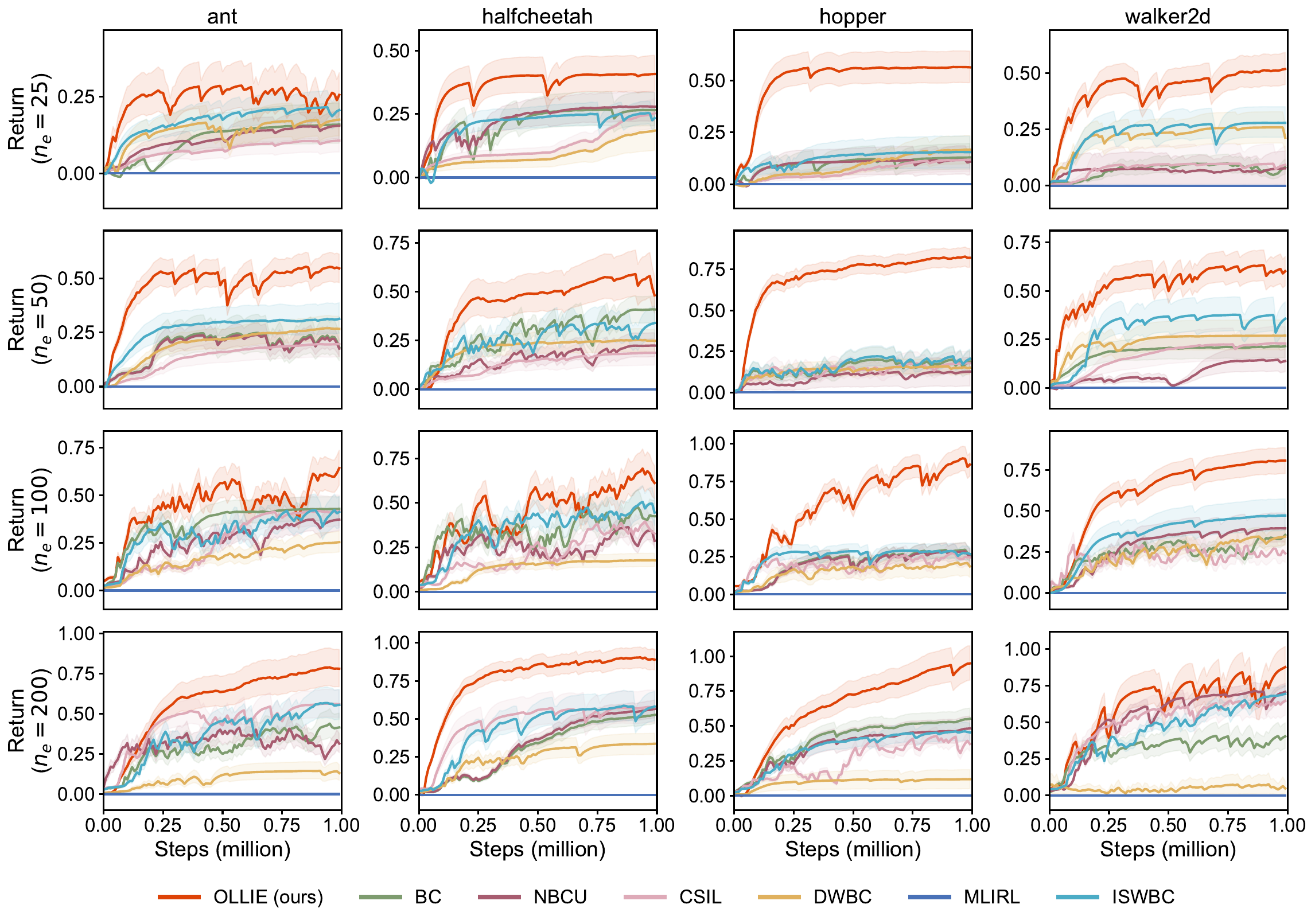}}
	\vskip -0.1in
	\caption{Learning curves in offline IL with varying quantities of expert trajectories in \textbf{\textit{vision-based MuJoCo}}. Uncertainty intervals depict standard deviation over five seeds. $n_e$ represents expert trajectories. \texttt{OLLIE} consistently and significantly surpasses existing methods in terms of convergence speed and stability.}
	\label{fig:curve_mujoco_image}
\end{figure*}

\clearpage
\begin{figure*}[t]
	\centering
	{\includegraphics[width=0.8\textwidth]{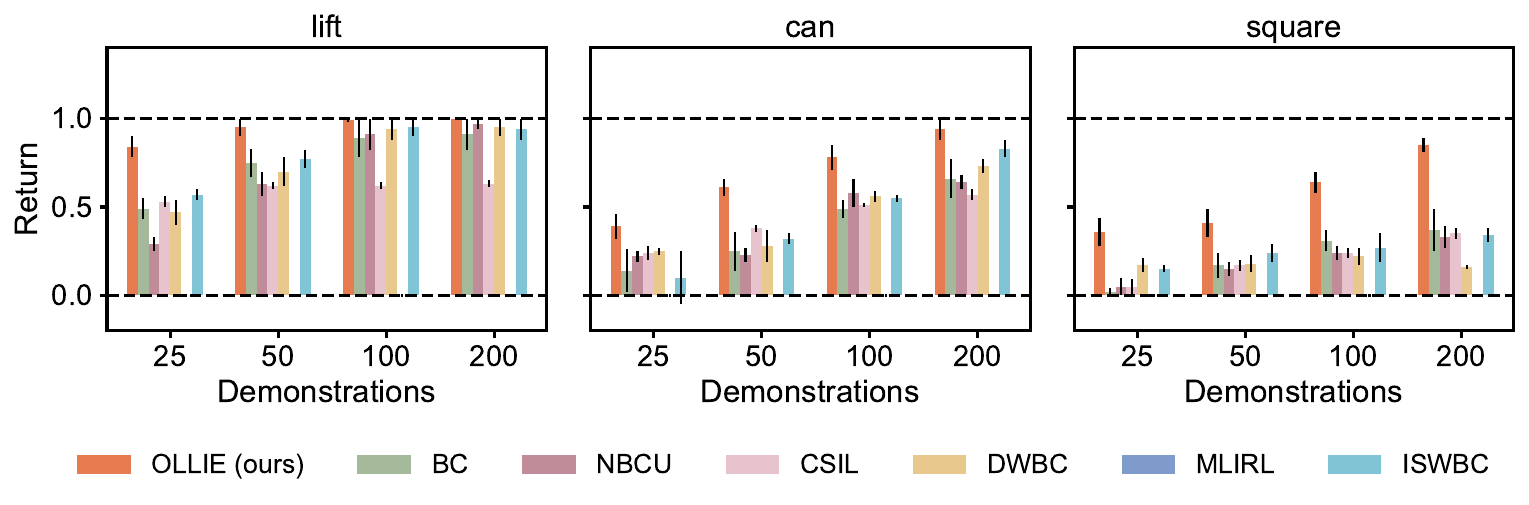}}
	\vskip -0.1in
	\caption{Performance in offline IL with varying quantities of expert trajectories in \textbf{\textit{vision-based Robomimic}}. Uncertainty intervals depict standard deviation over five seeds. \texttt{OLLIE} uses much fewer expert demonstrations to attain expert performance, demonstrating its great demonstration efficiency in high-dimensional environments.}
	\label{fig:performance_robomimic}
\end{figure*}

\begin{figure*}[t]
	\centering
	{\includegraphics[width=0.85\textwidth]{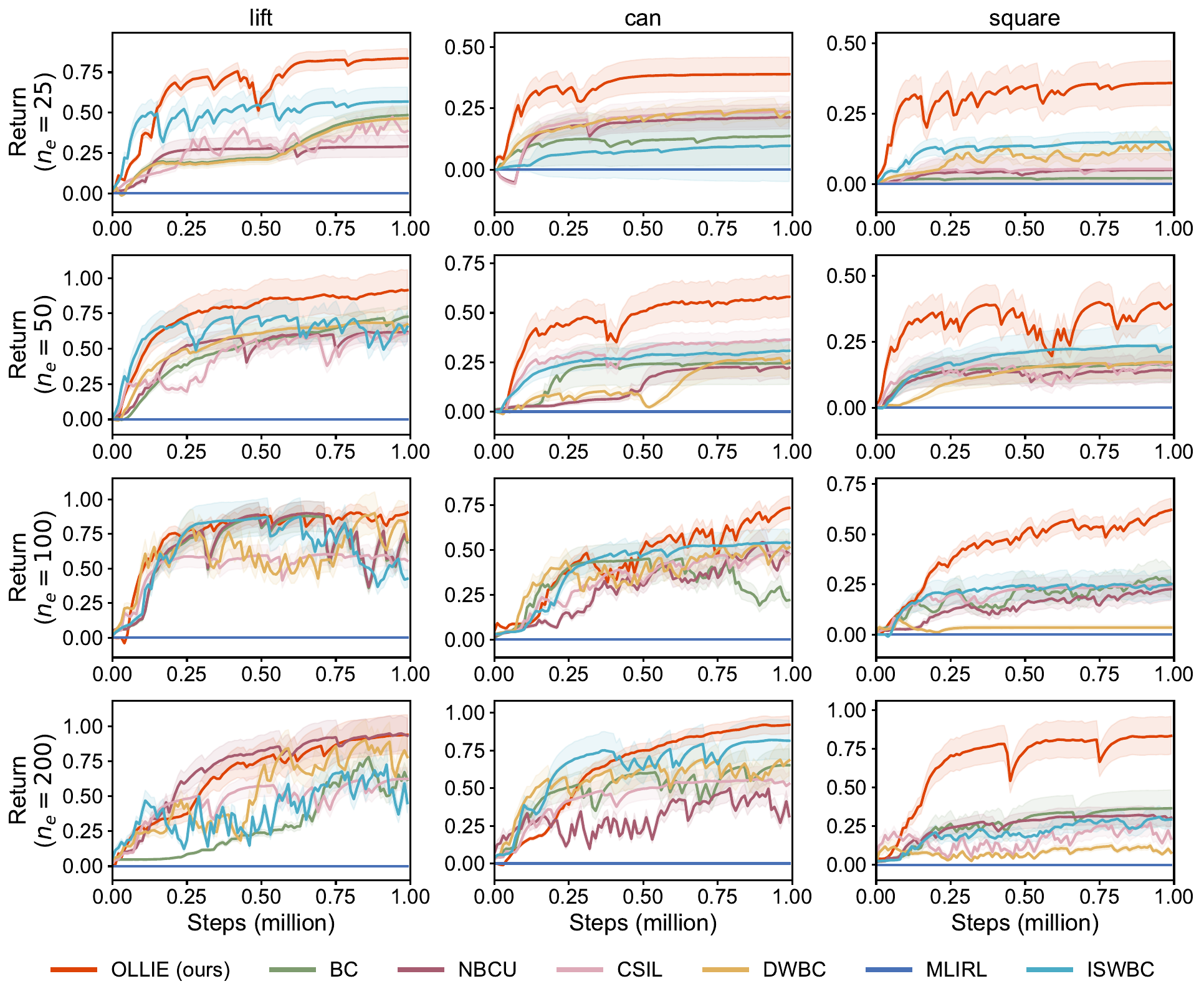}}
	\vskip -0.1in	
	\caption{Learning curves in offline IL with varying quantities of expert trajectories in \textbf{\textit{vision-based Robomimic}}. Uncertainty intervals depict standard deviation over five seeds. $n_e$ represents expert trajectories. \texttt{OLLIE} consistently exhibits fast and stabilized convergence.}
	\label{fig:curve_robomimic}
\end{figure*}

\clearpage

\subsubsection{Data Quality}
\label{sec:data_quality}

\begin{figure*}[ht]
	\centering
	{\includegraphics[width=\textwidth]{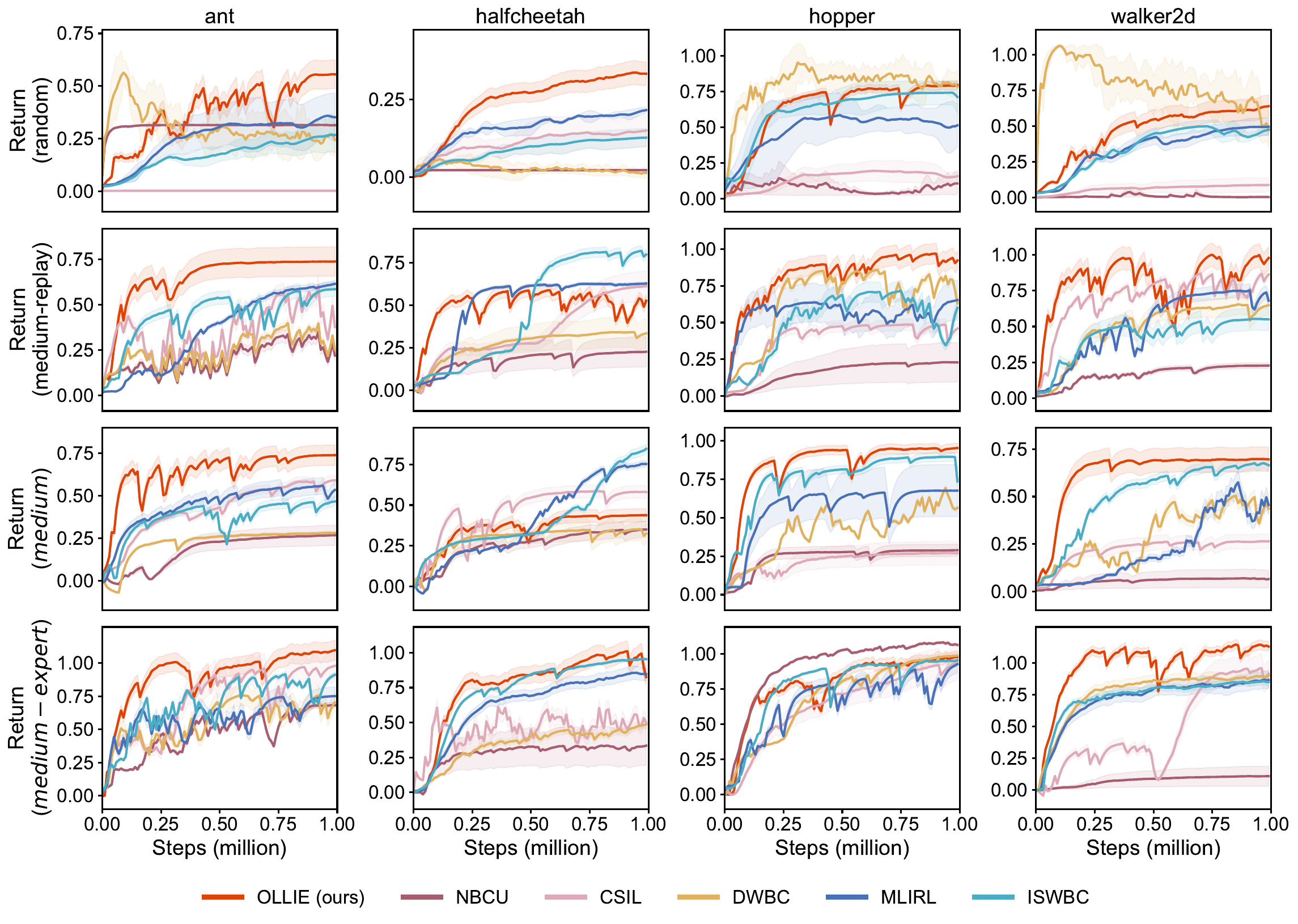}}
	\vskip -0.1in
	\caption{Learning curves in offline IL with varying qualities of imperfect trajectories in \textbf{\textit{MuJoCo}}. Uncertainty intervals depict standard deviation over five seeds. We use 1 expert trajectory, sampled from \texttt{expert} of \texttt{D4RL}, and 1000 imperfect trajectories, sampled from the corresponding datasets listed in \cref{tab:diverse_data_quality}. The length of each trajectory is less than 1000 time steps. \texttt{OLLIE} consistently exhibits fast and stabilized convergence. A higher quality of imperfect demonstrations often speeds up the convergence.}
	\label{fig:curve_data_quality_mujoco}
\end{figure*}

\begin{figure*}[t]
	\centering
	{\includegraphics[width=\textwidth]{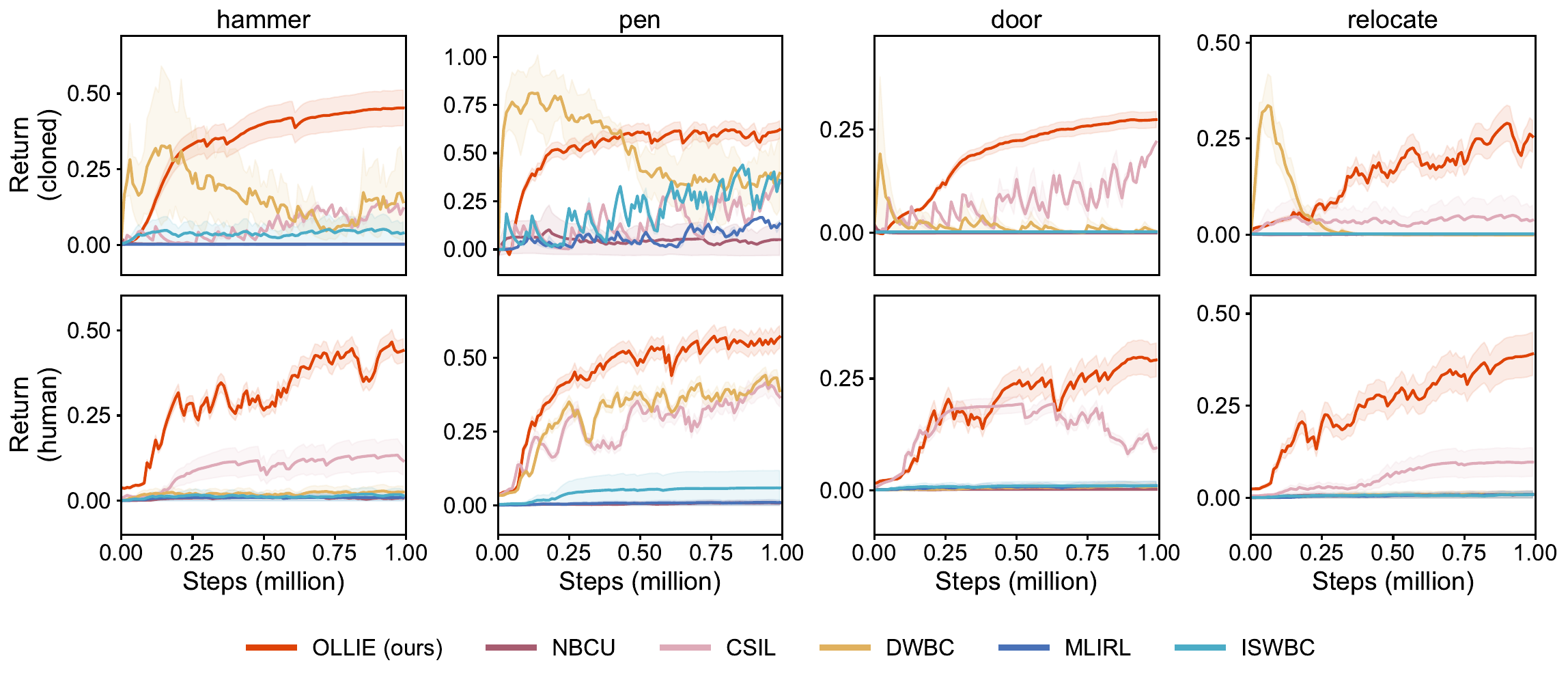}}
	\vskip -0.1in
	\caption{Learning curves in offline IL with varying qualities of imperfect trajectories in \textbf{\textit{Adroid}}. Uncertainty intervals depict standard deviation over five seeds. We use 10 expert trajectories, sampled from \texttt{expert} of \texttt{D4RL}, and 1000 imperfect trajectories, sampled from the corresponding datasets listed in \cref{tab:diverse_data_quality}. The length of each trajectory is less than 100 time steps. In contrast with the results in relatively low-dimensional MuJoCo, \texttt{OLLIE} significantly outperforms the baselines in this domain, demonstrating its robustness in complex and high-dimensional environments.}
	\label{fig:curve_data_quality_adroit}
\end{figure*}

Then, we conduct experiments using imperfect demonstrations with varying qualities to test the robustness of \texttt{ILID}'s performance. We present the result and data setup in \cref{tab:diverse_data_quality}. \texttt{OLLIE} outperforms the baselines in 20 out of 24 settings. The corresponding learning curves of \cref{tab:diverse_data_quality} are depicted in \cref{fig:curve_data_quality_mujoco,fig:curve_data_quality_adroit}.

\clearpage
\subsection{Performance in Online Finetuning}
\label{sec:performance_online}

After obtaining pretraining policies, we examine the finetuning performance under different quantities of expert demonstrations.  In \cref{tab:online_d_1,tab:online_d_3,tab:online_d_5,tab:online_d_10,tab:online_d_30,tab:online_d_25,tab:online_d_50,tab:online_d_100,tab:online_d_200}, we provide 
the performance before and after online finetuning with 10 episodes; subsequently, we present the end-to-end learning curves from offline pretraining to online finetuning across all tasks. 

\begin{figure*}[t]
	\centering
	\vskip -0.1in
	{\includegraphics[width=0.85\textwidth]{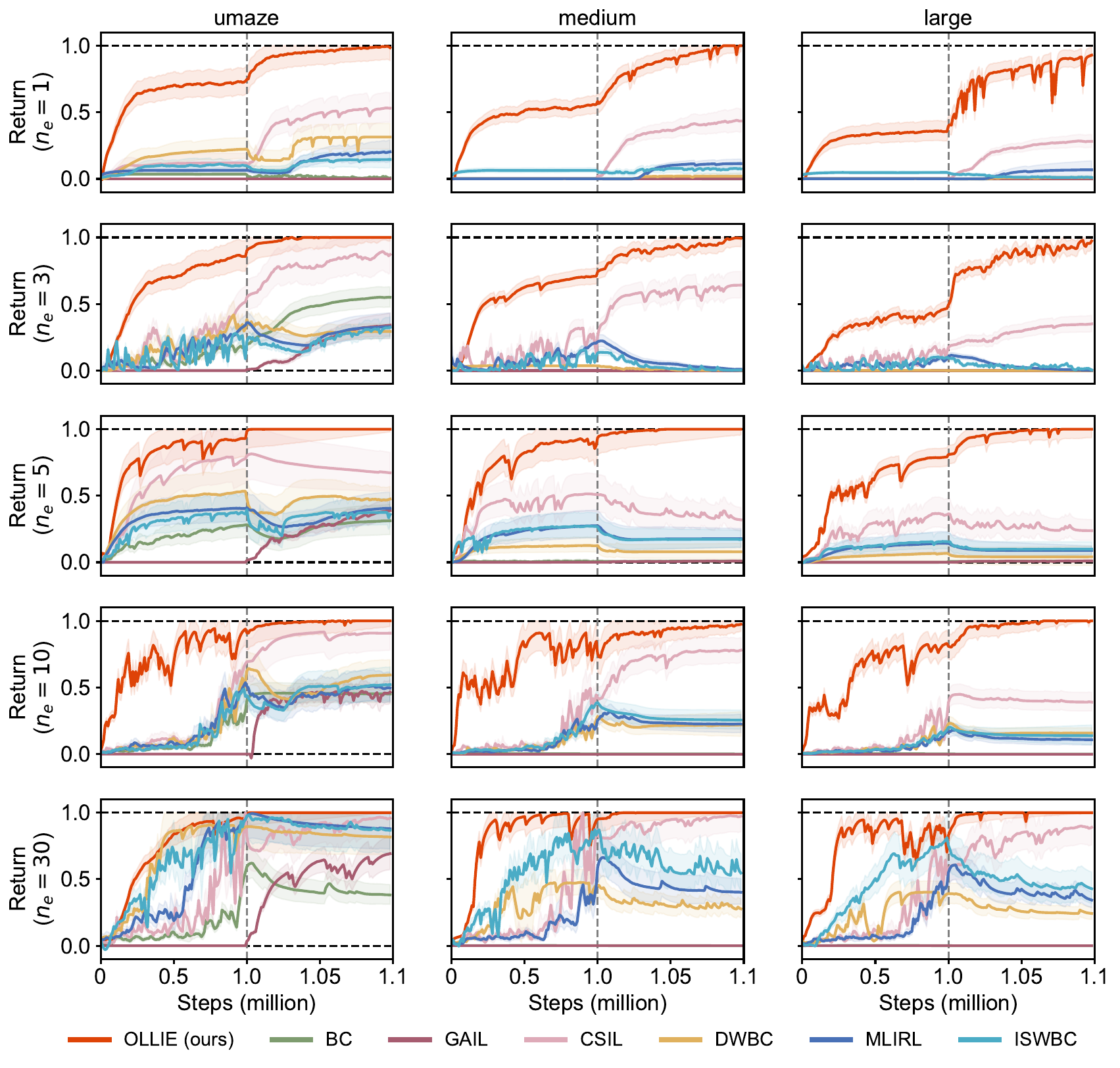}}
	\vskip -0.15in
	\caption{End-to-end learning curves from offline pretraining to online finetuning under varying quantities of expert trajectories in \textbf{\textit{AntMaze}}. Uncertainty intervals depict standard deviation over five seeds. $n_e$ represents expert trajectories. The simple combination of existing offline IL and \texttt{GAIL} suffers from unlearning pretrained knowledge. \texttt{OLLIE} not only avoids this issue but also expedites online training. This is attributed to \texttt{OLLIE}'s initial discriminator aligning with the well-performed policy initialization, thereby acting as a good local reward function capable of guiding fast policy search. Importantly, in the \texttt{medium} and \texttt{large} layouts, \texttt{GAIL} from scratch fails even with sufficient expert demonstrations. This may stem from the ineffective exploration of random policies in this domain, revealing the significance of effective pretraining in IL.}
	\label{fig:online_curve_antmaze}
\end{figure*}

\begin{figure*}[t]
	\centering
	{\includegraphics[width=0.85\textwidth]{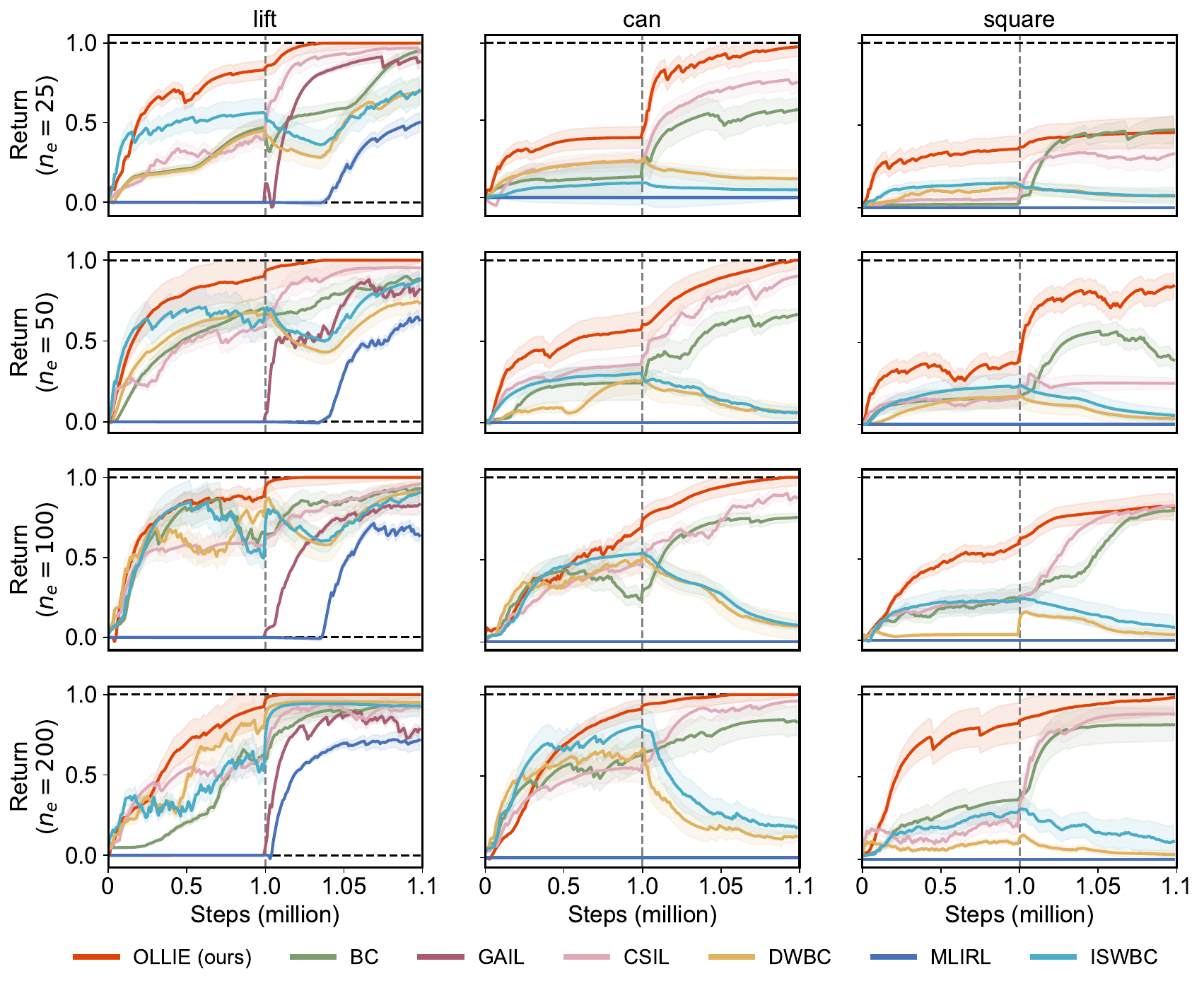}}
	\vskip -0.1in
	\caption{End-to-end learning curves from offline pretraining to online finetuning under varying quantities of expert trajectories in \textbf{\textit{vision-based Romomimic}}. Uncertainty intervals depict standard deviation over five seeds. $n_e$ represents expert trajectories. $n_e$ represents expert trajectories. The simple combination of existing offline IL and \texttt{GAIL} suffers from unlearning pretrained knowledge. \texttt{OLLIE} not only avoids this issue but also expedites online training. This is attributed to \texttt{OLLIE}'s initial discriminator aligning with the well-performed policy initialization, thereby acting as a good local reward function capable of guiding fast policy search. The results reveal \texttt{OLLIE}'s potential in practical, high-dimensional scenarios. In addition, analogously to AntMaze, \texttt{GAIL} from scratch proves unsuccessful even with sufficient expert demonstrations in \texttt{can} and \texttt{square}, which implies the importance of effective pretraining in IL.}
	\label{fig:online_curve_robomimic}
\end{figure*}

\textbf{Summary of key findings.} \texttt{OLLIE} successfully overcomes the unlearning problem and enables fast online finetuning that enables substantial performance improvement within a limited number of episodes. Importantly, \texttt{OLLIE} can work well in the cases where \texttt{GAIL} fails (as illustrated in \cref{fig:online_curve_antmaze,fig:online_curve_robomimic}), highlighting the importance of pretraining IL.

\begin{table*}[ht]
	\centering
	\caption{Normalized performance before and after online finetuning with \textbf{10} episodes.}
	{(\# expert trajectories: \textbf{1} in AntMaze/MuJoCo and \textbf{10} in Adroit/FrankaKitchen)}
	\label{tab:online_d_1}
	\vskip 0.1in
	{\small \begin{tabular}{lccccccc} 
			\toprule
			Task                 & \texttt{OLLIE}    & \texttt{BC}       & \texttt{NBCU}     & \texttt{CSIL}     & \texttt{DWBC}     & \texttt{MLIRL}    & \texttt{ISWBC}     \\ 
			\midrule
			\texttt{ant}         & $55\rightarrow\bm{83}$ & $31\rightarrow00$ & $00\rightarrow36$ & $00\rightarrow42$ & $23\rightarrow13$ & $34\rightarrow24$ & $26\rightarrow16$  \\
			\texttt{halfcheetah} & $33\rightarrow\bm{47}$ & $02\rightarrow02$ & $00\rightarrow12$ & $15\rightarrow40$ & $00\rightarrow02$ & $22\rightarrow14$ & $12\rightarrow07$  \\
			\texttt{hopper}      & $80\rightarrow\bm{90}$ & $07\rightarrow21$ & $00\rightarrow14$ & $16\rightarrow69$ & $76\rightarrow44$ & $52\rightarrow32$ & $67\rightarrow39$  \\
			\texttt{walker2d}    & $64\rightarrow\bm{64}$ & $00\rightarrow02$ & $00\rightarrow10$ & $08\rightarrow20$ & $57\rightarrow31$ & $50\rightarrow29$ & $47\rightarrow35$  \\
			\texttt{hammer}      & $45\rightarrow\bm{53}$ & $00\rightarrow00$ & $00\rightarrow00$ & $15\rightarrow24$ & $08\rightarrow02$ & $00\rightarrow00$ & $04\rightarrow02$  \\
			\texttt{pen}         & $64\rightarrow\bm{72}$ & $05\rightarrow35$ & $00\rightarrow06$ & $34\rightarrow41$ & $38\rightarrow23$ & $17\rightarrow09$ & $47\rightarrow25$  \\
			\texttt{door}        & $27\rightarrow\bm{48}$ & $00\rightarrow02$ & $00\rightarrow00$ & $05\rightarrow92$ & $00\rightarrow01$ & $00\rightarrow01$ & $00\rightarrow01$  \\
			\texttt{relocate}    & $24\rightarrow\bm{38}$ & $00\rightarrow03$ & $00\rightarrow07$ & $04\rightarrow09$ & $00\rightarrow03$ & $00\rightarrow02$ & $00\rightarrow00$  \\
			\texttt{umaze}       & $75\rightarrow\bm{85}$ & $03\rightarrow02$ & $00\rightarrow00$ & $12\rightarrow22$ & $22\rightarrow14$ & $06\rightarrow04$ & $09\rightarrow06$  \\
			\texttt{medium}      & $56\rightarrow\bm{67}$ & $00\rightarrow00$ & $00\rightarrow00$ & $00\rightarrow15$ & $00\rightarrow00$ & $00\rightarrow00$ & $06\rightarrow04$  \\
			\texttt{large}       & $35\rightarrow\bm{61}$ & $00\rightarrow00$ & $00\rightarrow00$ & $00\rightarrow06$ & $00\rightarrow00$ & $00\rightarrow00$ & $04\rightarrow03$  \\
			\texttt{complete}    & $33\rightarrow\bm{43}$ & $00\rightarrow03$ & $00\rightarrow02$ & $16\rightarrow15$ & $00\rightarrow00$ & $00\rightarrow00$ & $00\rightarrow02$  \\
			\texttt{partial}     & $32\rightarrow\bm{44}$ & $00\rightarrow01$ & $00\rightarrow07$ & $25\rightarrow27$ & $00\rightarrow00$ & $00\rightarrow00$ & $00\rightarrow02$  \\
			\texttt{undirect}    & $36\rightarrow\bm{58}$ & $00\rightarrow11$ & $00\rightarrow23$ & $36\rightarrow35$ & $00\rightarrow00$ & $00\rightarrow00$ & $00\rightarrow01$  \\
			\bottomrule
	\end{tabular}}
\end{table*}

\begin{table*}[ht]
	\centering
	\caption{Normalized performance before and after online finetuning with \textbf{10} episodes.} 
	{(\# expert trajectories: \textbf{3} in AntMaze/MuJoCo and \textbf{30} in Adroit/FrankaKitchen)}
	\vskip 0.1in
		{\small \begin{tabular}{lccccccc} 
				\toprule
				Task                 & \texttt{OLLIE}    & \texttt{BC}       & \texttt{NBCU}     & \texttt{CSIL}     & \texttt{DWBC}     & \texttt{MLIRL}    & \texttt{ISWBC}     \\ 
				\midrule
				\texttt{ant}         & $75\rightarrow\bm{82}$ & $27\rightarrow47$ & $00\rightarrow19$ & $08\rightarrow42$ & $30\rightarrow23$ & $49\rightarrow31$ & $29\rightarrow18$  \\
				\texttt{halfcheetah} & $75\rightarrow\bm{84}$ & $02\rightarrow17$ & $00\rightarrow26$ & $21\rightarrow58$ & $45\rightarrow32$ & $56\rightarrow45$ & $28\rightarrow29$  \\
				\texttt{hopper}      & $85\rightarrow\bm{91}$ & $10\rightarrow53$ & $00\rightarrow06$ & $64\rightarrow82$ & $70\rightarrow54$ & $66\rightarrow53$ & $38\rightarrow43$  \\
				\texttt{walker2d}    & $74\rightarrow\bm{89}$ & $03\rightarrow21$ & $00\rightarrow63$ & $18\rightarrow67$ & $70\rightarrow52$ & $83\rightarrow36$ & $53\rightarrow35$  \\
				\texttt{hammer}      & $44\rightarrow\bm{85}$ & $00\rightarrow00$ & $00\rightarrow00$ & $22\rightarrow46$ & $01\rightarrow00$ & $00\rightarrow00$ & $10\rightarrow06$  \\
				\texttt{pen}         & $71\rightarrow\bm{89}$ & $15\rightarrow54$ & $00\rightarrow11$ & $47\rightarrow50$ & $48\rightarrow60$ & $39\rightarrow38$ & $36\rightarrow46$  \\
				\texttt{door}        & $61\rightarrow\bm{89}$ & $03\rightarrow44$ & $00\rightarrow00$ & $45\rightarrow46$ & $00\rightarrow01$ & $00\rightarrow00$ & $05\rightarrow08$  \\
				\texttt{relocate}    & $44\rightarrow\bm{52}$ & $00\rightarrow12$ & $00\rightarrow06$ & $26\rightarrow26$ & $00\rightarrow00$ & $00\rightarrow00$ & $09\rightarrow05$  \\
				\texttt{umaze}       & $85\rightarrow\bm{95}$ & $29\rightarrow28$ & $00\rightarrow04$ & $50\rightarrow69$ & $28\rightarrow42$ & $32\rightarrow23$ & $02\rightarrow21$  \\
				\texttt{medium}      & $70\rightarrow\bm{81}$ & $00\rightarrow00$ & $00\rightarrow00$ & $12\rightarrow51$ & $03\rightarrow03$ & $22\rightarrow17$ & $15\rightarrow13$  \\
				\texttt{large}       & $47\rightarrow\bm{73}$ & $00\rightarrow00$ & $00\rightarrow00$ & $18\rightarrow20$ & $00\rightarrow00$ & $11\rightarrow09$ & $09\rightarrow04$  \\
				\texttt{complete}    & $60\rightarrow\bm{61}$ & $00\rightarrow28$ & $00\rightarrow07$ & $04\rightarrow44$ & $00\rightarrow02$ & $00\rightarrow00$ & $00\rightarrow02$  \\
				\texttt{partial}     & $66\rightarrow\bm{72}$ & $00\rightarrow36$ & $00\rightarrow20$ & $24\rightarrow50$ & $00\rightarrow02$ & $00\rightarrow00$ & $00\rightarrow02$  \\
				\texttt{undirect}    & $73\rightarrow\bm{84}$ & $00\rightarrow18$ & $00\rightarrow07$ & $62\rightarrow48$ & $00\rightarrow01$ & $00\rightarrow00$ & $05\rightarrow03$  \\
				\bottomrule
		\end{tabular}}
	\label{tab:online_d_3}
\end{table*}

\begin{table*}[ht]
	\centering
	\caption{Normalized performance before and after online finetuning with \textbf{10} episodes.} 
	{(\# expert trajectories: \textbf{5} in AntMaze/MuJoCo and \textbf{50} in Adroit/FrankaKitchen)}
	\vskip 0.1in
		{\small \begin{tabular}{lccccccc} 
				\toprule
				Task                 & \texttt{OLLIE}      & \texttt{BC}       & \texttt{NBCU}     & \texttt{CSIL}     & \texttt{DWBC}     & \texttt{MLIRL}    & \texttt{ISWBC}     \\ 
				\midrule
				\texttt{ant}         & $103\rightarrow\bm{112}$ & $55\rightarrow40$ & $00\rightarrow13$ & $59\rightarrow65$ & $78\rightarrow53$ & $56\rightarrow40$ & $49\rightarrow31$  \\
				\texttt{halfcheetah} & $075\rightarrow\bm{097}$ & $33\rightarrow19$ & $00\rightarrow41$ & $59\rightarrow61$ & $47\rightarrow29$ & $80\rightarrow61$ & $51\rightarrow34$  \\
				\texttt{hopper}      & $061\rightarrow\bm{099}$ & $19\rightarrow27$ & $00\rightarrow21$ & $85\rightarrow87$ & $74\rightarrow46$ & $76\rightarrow52$ & $63\rightarrow41$  \\
				\texttt{walker2d}    & $092\rightarrow\bm{109}$ & $12\rightarrow11$ & $00\rightarrow54$ & $68\rightarrow71$ & $66\rightarrow43$ & $87\rightarrow56$ & $60\rightarrow44$  \\
				\texttt{hammer}      & $075\rightarrow\bm{093}$ & $00\rightarrow00$ & $00\rightarrow00$ & $60\rightarrow58$ & $00\rightarrow00$ & $00\rightarrow00$ & $22\rightarrow15$  \\
				\texttt{pen}         & $085\rightarrow\bm{094}$ & $25\rightarrow35$ & $00\rightarrow12$ & $62\rightarrow63$ & $79\rightarrow75$ & $56\rightarrow41$ & $61\rightarrow41$  \\
				\texttt{door}        & $076\rightarrow\bm{096}$ & $00\rightarrow00$ & $00\rightarrow03$ & $55\rightarrow54$ & $00\rightarrow00$ & $00\rightarrow00$ & $21\rightarrow15$  \\
				\texttt{relocate}    & $068\rightarrow\bm{084}$ & $00\rightarrow00$ & $00\rightarrow33$ & $59\rightarrow60$ & $00\rightarrow00$ & $00\rightarrow00$ & $28\rightarrow19$  \\
				\texttt{umaze}       & $098\rightarrow\bm{099}$ & $32\rightarrow20$ & $00\rightarrow10$ & $81\rightarrow78$ & $54\rightarrow36$ & $39\rightarrow31$ & $35\rightarrow24$  \\
				\texttt{medium}      & $093\rightarrow\bm{096}$ & $00\rightarrow00$ & $00\rightarrow00$ & $51\rightarrow44$ & $12\rightarrow09$ & $27\rightarrow18$ & $26\rightarrow20$  \\
				\texttt{large}       & $079\rightarrow\bm{088}$ & $00\rightarrow00$ & $00\rightarrow00$ & $37\rightarrow26$ & $06\rightarrow04$ & $14\rightarrow10$ & $15\rightarrow11$  \\
				\texttt{complete}    & $060\rightarrow\bm{078}$ & $00\rightarrow02$ & $00\rightarrow05$ & $49\rightarrow52$ & $08\rightarrow05$ & $00\rightarrow00$ & $03\rightarrow02$  \\
				\texttt{partial}     & $072\rightarrow\bm{088}$ & $00\rightarrow02$ & $00\rightarrow45$ & $72\rightarrow75$ & $09\rightarrow06$ & $00\rightarrow00$ & $04\rightarrow06$  \\
				\texttt{undirect}    & $074\rightarrow\bm{085}$ & $00\rightarrow05$ & $00\rightarrow10$ & $83\rightarrow83$ & $11\rightarrow07$ & $00\rightarrow00$ & $12\rightarrow10$  \\
				\bottomrule
		\end{tabular}}
	\label{tab:online_d_5}
\end{table*}

\begin{table*}[ht]
	\centering
	\caption{Normalized performance before and after online finetuning with \textbf{10} episodes.} 
	{(\# expert trajectories: \textbf{10} in AntMaze/MuJoCo and \textbf{100} in Adroit/FrankaKitchen)}
	\vskip 0.1in
		{\small \begin{tabular}{lccccccc} 
				\toprule
				Task                 & \texttt{OLLIE}      & \texttt{BC}       & \texttt{NBCU}     & \texttt{CSIL}     & \texttt{DWBC}     & \texttt{MLIRL}    & \texttt{ISWBC}     \\ 
				\midrule
				\texttt{ant}         & $114\rightarrow\bm{119}$ & $56\rightarrow64$ & $00\rightarrow33$ & $74\rightarrow94$ & $83\rightarrow66$ & $65\rightarrow39$ & $74\rightarrow55$  \\
				\texttt{halfcheetah} & $110\rightarrow\bm{106}$ & $02\rightarrow84$ & $00\rightarrow70$ & $78\rightarrow82$ & $58\rightarrow47$ & $73\rightarrow50$ & $74\rightarrow51$  \\
				\texttt{hopper}      & $097\rightarrow\bm{105}$ & $05\rightarrow75$ & $00\rightarrow70$ & $94\rightarrow78$ & $59\rightarrow48$ & $82\rightarrow60$ & $03\rightarrow59$  \\
				\texttt{walker2d}    & $094\rightarrow\bm{108}$ & $20\rightarrow81$ & $00\rightarrow23$ & $90\rightarrow92$ & $69\rightarrow45$ & $91\rightarrow64$ & $59\rightarrow48$  \\
				\texttt{hammer}      & $079\rightarrow\bm{099}$ & $00\rightarrow43$ & $00\rightarrow00$ & $65\rightarrow67$ & $64\rightarrow59$ & $00\rightarrow00$ & $45\rightarrow28$  \\
				\texttt{pen}         & $093\rightarrow\bm{098}$ & $23\rightarrow57$ & $00\rightarrow15$ & $69\rightarrow65$ & $60\rightarrow53$ & $66\rightarrow37$ & $51\rightarrow39$  \\
				\texttt{door}        & $090\rightarrow\bm{097}$ & $00\rightarrow31$ & $00\rightarrow59$ & $06\rightarrow93$ & $16\rightarrow15$ & $00\rightarrow00$ & $36\rightarrow27$  \\
				\texttt{relocate}    & $090\rightarrow\bm{095}$ & $00\rightarrow60$ & $00\rightarrow52$ & $69\rightarrow73$ & $36\rightarrow25$ & $00\rightarrow00$ & $58\rightarrow41$  \\
				\texttt{umaze}       & $099\rightarrow\bm{095}$ & $12\rightarrow46$ & $00\rightarrow32$ & $30\rightarrow82$ & $67\rightarrow59$ & $64\rightarrow34$ & $52\rightarrow33$  \\
				\texttt{medium}      & $091\rightarrow\bm{085}$ & $00\rightarrow00$ & $00\rightarrow00$ & $18\rightarrow50$ & $32\rightarrow21$ & $29\rightarrow35$ & $44\rightarrow31$  \\
				\texttt{large}       & $084\rightarrow\bm{088}$ & $00\rightarrow00$ & $00\rightarrow00$ & $32\rightarrow45$ & $24\rightarrow16$ & $20\rightarrow13$ & $22\rightarrow16$  \\
				\texttt{complete}    & $097\rightarrow\bm{100}$ & $03\rightarrow15$ & $00\rightarrow66$ & $52\rightarrow67$ & $22\rightarrow15$ & $00\rightarrow00$ & $05\rightarrow03$  \\
				\texttt{partial}     & $098\rightarrow\bm{100}$ & $03\rightarrow23$ & $00\rightarrow54$ & $71\rightarrow80$ & $25\rightarrow18$ & $00\rightarrow00$ & $29\rightarrow14$  \\
				\texttt{undirect}    & $099\rightarrow\bm{100}$ & $03\rightarrow75$ & $00\rightarrow28$ & $89\rightarrow79$ & $28\rightarrow22$ & $00\rightarrow00$ & $05\rightarrow23$  \\
				\bottomrule
		\end{tabular}}
	\label{tab:online_d_10}
\end{table*}

\begin{table*}[ht]
	\centering
	\caption{Normalized performance before and after online finetuning with \textbf{10} episodes.} 
	{(\# expert trajectories: \textbf{30} in AntMaze/MuJoCo and \textbf{300} in Adroit/FrankaKitchen)}
	\vskip 0.1in
		{\small \begin{tabular}{lrrrrrrr} 
				\toprule
				Task                 & \texttt{OLLIE}      & \texttt{BC}         & \texttt{NBCU}     & \texttt{CSIL}       & \texttt{DWBC}       & \texttt{MLIRL}      & \texttt{ISWBC}       \\ 
				\midrule
				\texttt{ant}         & $117\rightarrow\bm{119}$ & $091\rightarrow086$ & $00\rightarrow34$ & $104\rightarrow104$ & $090\rightarrow088$ & $109\rightarrow095$ & $105\rightarrow100$  \\
				\texttt{halfcheetah} & $110\rightarrow\bm{108}$ & $002\rightarrow093$ & $00\rightarrow54$ & $112\rightarrow100$ & $088\rightarrow090$ & $084\rightarrow094$ & $105\rightarrow101$  \\
				\texttt{hopper}      & $107\rightarrow\bm{108}$ & $031\rightarrow070$ & $00\rightarrow99$ & $078\rightarrow093$ & $109\rightarrow107$ & $099\rightarrow104$ & $108\rightarrow099$  \\
				\texttt{walker2d}    & $102\rightarrow\bm{107}$ & $023\rightarrow090$ & $00\rightarrow34$ & $099\rightarrow100$ & $096\rightarrow097$ & $079\rightarrow092$ & $095\rightarrow092$  \\
				\texttt{hammer}      & $099\rightarrow\bm{109}$ & $000\rightarrow095$ & $00\rightarrow16$ & $100\rightarrow091$ & $075\rightarrow093$ & $053\rightarrow059$ & $088\rightarrow101$  \\
				\texttt{pen}         & $098\rightarrow\bm{101}$ & $057\rightarrow094$ & $00\rightarrow70$ & $093\rightarrow072$ & $053\rightarrow098$ & $074\rightarrow078$ & $103\rightarrow092$  \\
				\texttt{door}        & $107\rightarrow\bm{106}$ & $000\rightarrow044$ & $00\rightarrow16$ & $010\rightarrow096$ & $073\rightarrow062$ & $042\rightarrow038$ & $087\rightarrow052$  \\
				\texttt{relocate}    & $099\rightarrow\bm{100}$ & $000\rightarrow100$ & $00\rightarrow83$ & $094\rightarrow084$ & $038\rightarrow100$ & $045\rightarrow084$ & $099\rightarrow103$  \\
				\texttt{umaze}       & $099\rightarrow\bm{100}$ & $053\rightarrow051$ & $00\rightarrow42$ & $119\rightarrow070$ & $090\rightarrow088$ & $055\rightarrow097$ & $103\rightarrow096$  \\
				\texttt{medium}      & $099\rightarrow\bm{095}$ & $000\rightarrow000$ & $00\rightarrow00$ & $082\rightarrow080$ & $047\rightarrow034$ & $049\rightarrow057$ & $082\rightarrow070$  \\
				\texttt{large}       & $046\rightarrow\bm{097}$ & $000\rightarrow000$ & $00\rightarrow00$ & $075\rightarrow060$ & $040\rightarrow037$ & $020\rightarrow051$ & $090\rightarrow065$  \\
				\texttt{complete}    & $098\rightarrow\bm{100}$ & $011\rightarrow075$ & $00\rightarrow18$ & $082\rightarrow089$ & $032\rightarrow048$ & $000\rightarrow034$ & $020\rightarrow052$  \\
				\texttt{partial}     & $040\rightarrow\bm{100}$ & $015\rightarrow070$ & $00\rightarrow65$ & $004\rightarrow092$ & $040\rightarrow059$ & $000\rightarrow031$ & $038\rightarrow059$  \\
				\texttt{undirect}    & $099\rightarrow\bm{100}$ & $018\rightarrow068$ & $00\rightarrow14$ & $094\rightarrow095$ & $063\rightarrow066$ & $000\rightarrow027$ & $054\rightarrow073$  \\
				\bottomrule
		\end{tabular}}
	\label{tab:online_d_30}
\end{table*}

\begin{table*}[ht]
	\centering
	\caption{Normalized performance before \& after online finetuning with \textbf{10} episodes in vision-based tasks.} 
	{(\# expert trajectories: \textbf{25})}
	\vskip 0.1in
		{\small \begin{tabular}{lccccccc} 
				\toprule
				Task                 & \texttt{OLLIE}    & \texttt{BC}       & \texttt{NBCU}     & \texttt{CSIL}     & \texttt{DWBC}     & \texttt{MLIRL}    & \texttt{ISWBC}     \\ 
				\midrule
				\texttt{ant}         & $29\rightarrow\bm{59}$ & $15\rightarrow20$ & $00\rightarrow02$ & $10\rightarrow16$ & $17\rightarrow14$ & $00\rightarrow00$ & $21\rightarrow13$  \\
				\texttt{halfcheetah} & $40\rightarrow\bm{44}$ & $27\rightarrow28$ & $00\rightarrow02$ & $26\rightarrow30$ & $19\rightarrow13$ & $00\rightarrow00$ & $25\rightarrow19$  \\
				\texttt{hopper}      & $56\rightarrow\bm{68}$ & $11\rightarrow15$ & $00\rightarrow06$ & $11\rightarrow19$ & $16\rightarrow13$ & $00\rightarrow00$ & $15\rightarrow11$  \\
				\texttt{walker2d}    & $52\rightarrow\bm{57}$ & $07\rightarrow14$ & $00\rightarrow20$ & $09\rightarrow13$ & $10\rightarrow16$ & $00\rightarrow00$ & $27\rightarrow20$  \\
				\texttt{lift}        & $83\rightarrow\bm{96}$ & $29\rightarrow52$ & $00\rightarrow62$ & $39\rightarrow80$ & $46\rightarrow30$ & $00\rightarrow00$ & $56\rightarrow39$  \\
				\texttt{can}         & $38\rightarrow\bm{90}$ & $21\rightarrow45$ & $00\rightarrow00$ & $23\rightarrow55$ & $24\rightarrow17$ & $00\rightarrow00$ & $09\rightarrow06$  \\
				\texttt{square}      & $35\rightarrow\bm{38}$ & $05\rightarrow35$ & $00\rightarrow00$ & $05\rightarrow33$ & $12\rightarrow10$ & $00\rightarrow00$ & $06\rightarrow11$  \\
				\bottomrule
		\end{tabular}}
	\label{tab:online_d_25}
\end{table*}

\begin{table*}[ht]
	\centering
	\caption{Normalized performance before \& after online finetuning with \textbf{10} episodes in vision-based tasks.} 
	{(\# expert trajectories: \textbf{50})}
	\vskip 0.1in
		{\small \begin{tabular}{lccccccc} 
				\toprule
				Task                 & \texttt{OLLIE}    & \texttt{BC}       & \texttt{NBCU}     & \texttt{CSIL}     & \texttt{DWBC}     & \texttt{MLIRL}    & \texttt{ISWBC}     \\ 
				\midrule
				\texttt{ant}         & $29\rightarrow\bm{59}$ & $15\rightarrow20$ & $00\rightarrow02$ & $10\rightarrow16$ & $17\rightarrow14$ & $00\rightarrow00$ & $21\rightarrow13$  \\
				\texttt{halfcheetah} & $40\rightarrow\bm{44}$ & $27\rightarrow28$ & $00\rightarrow02$ & $26\rightarrow30$ & $19\rightarrow13$ & $00\rightarrow00$ & $25\rightarrow19$  \\
				\texttt{hopper}      & $56\rightarrow\bm{68}$ & $11\rightarrow15$ & $00\rightarrow06$ & $11\rightarrow19$ & $16\rightarrow13$ & $00\rightarrow00$ & $15\rightarrow11$  \\
				\texttt{walker2d}    & $52\rightarrow\bm{57}$ & $07\rightarrow14$ & $00\rightarrow20$ & $09\rightarrow13$ & $10\rightarrow16$ & $00\rightarrow00$ & $27\rightarrow20$  \\
				\texttt{lift}        & $83\rightarrow\bm{96}$ & $29\rightarrow52$ & $00\rightarrow62$ & $39\rightarrow80$ & $46\rightarrow30$ & $00\rightarrow00$ & $56\rightarrow39$  \\
				\texttt{can}         & $38\rightarrow\bm{90}$ & $21\rightarrow45$ & $00\rightarrow00$ & $23\rightarrow55$ & $24\rightarrow17$ & $00\rightarrow00$ & $09\rightarrow06$  \\
				\texttt{square}      & $35\rightarrow\bm{38}$ & $05\rightarrow35$ & $00\rightarrow00$ & $05\rightarrow33$ & $12\rightarrow10$ & $00\rightarrow00$ & $06\rightarrow11$  \\
				\bottomrule
		\end{tabular}}
	\label{tab:online_d_50}
\end{table*}

\begin{table*}[ht]
	\centering
	\caption{Normalized performance before \& after online finetuning with \textbf{10} episodes in vision-based tasks.} 
	{(\# expert trajectories: \textbf{100})}
	\vskip 0.1in
		{\small \begin{tabular}{lrrrrrrr} 
				\toprule
				Task                 & \texttt{OLLIE}    & \texttt{BC}       & \texttt{NBCU}     & \texttt{CSIL}     & \texttt{DWBC}     & \texttt{MLIRL}    & \texttt{ISWBC}     \\ 
				\midrule
				\texttt{ant}         & $70\rightarrow\bm{74}$ & $37\rightarrow42$ & $00\rightarrow38$ & $41\rightarrow45$ & $25\rightarrow17$ & $00\rightarrow00$ & $43\rightarrow31$  \\
				\texttt{halfcheetah} & $56\rightarrow\bm{86}$ & $16\rightarrow42$ & $00\rightarrow09$ & $30\rightarrow35$ & $17\rightarrow13$ & $00\rightarrow00$ & $32\rightarrow40$  \\
				\texttt{hopper}      & $88\rightarrow\bm{95}$ & $28\rightarrow49$ & $00\rightarrow04$ & $24\rightarrow49$ & $13\rightarrow15$ & $00\rightarrow00$ & $25\rightarrow22$  \\
				\texttt{walker2d}    & $80\rightarrow\bm{88}$ & $39\rightarrow45$ & $00\rightarrow19$ & $20\rightarrow22$ & $34\rightarrow27$ & $00\rightarrow00$ & $47\rightarrow35$  \\
				\texttt{lift}        & $94\rightarrow\bm{99}$ & $60\rightarrow79$ & $00\rightarrow53$ & $52\rightarrow71$ & $40\rightarrow70$ & $00\rightarrow00$ & $44\rightarrow75$  \\
				\texttt{can}         & $74\rightarrow\bm{81}$ & $45\rightarrow55$ & $00\rightarrow00$ & $49\rightarrow61$ & $53\rightarrow39$ & $00\rightarrow00$ & $53\rightarrow43$  \\
				\texttt{square}      & $63\rightarrow\bm{69}$ & $22\rightarrow32$ & $00\rightarrow00$ & $23\rightarrow32$ & $03\rightarrow14$ & $00\rightarrow00$ & $25\rightarrow26$  \\
				\bottomrule
		\end{tabular}}
	\label{tab:online_d_100}
\end{table*}

\begin{table*}[ht]
	\centering
	\caption{Normalized performance before \& after online finetuning with \textbf{10} episodes in vision-based tasks.} 
	{(\# expert trajectories: \textbf{200})}
	\vskip 0.1in
		{\small \begin{tabular}{lccccccc} 
				\toprule
				Task                 & \texttt{OLLIE}      & \texttt{BC}       & \texttt{NBCU}     & \texttt{CSIL}     & \texttt{DWBC}     & \texttt{MLIRL}    & \texttt{ISWBC}     \\ 
				\midrule
				\texttt{ant}         & $077\rightarrow\bm{098}$ & $26\rightarrow42$ & $00\rightarrow08$ & $55\rightarrow69$ & $14\rightarrow45$ & $00\rightarrow35$ & $57\rightarrow58$  \\
				\texttt{halfcheetah} & $088\rightarrow\bm{101}$ & $57\rightarrow65$ & $00\rightarrow34$ & $57\rightarrow66$ & $33\rightarrow64$ & $00\rightarrow17$ & $59\rightarrow64$  \\
				\texttt{hopper}      & $095\rightarrow\bm{100}$ & $48\rightarrow58$ & $00\rightarrow30$ & $31\rightarrow57$ & $11\rightarrow32$ & $00\rightarrow14$ & $44\rightarrow49$  \\
				\texttt{walker2d}    & $089\rightarrow\bm{092}$ & $72\rightarrow54$ & $00\rightarrow64$ & $66\rightarrow86$ & $05\rightarrow48$ & $00\rightarrow07$ & $70\rightarrow71$  \\
				\texttt{lift}        & $098\rightarrow\bm{100}$ & $91\rightarrow83$ & $00\rightarrow81$ & $62\rightarrow92$ & $67\rightarrow95$ & $00\rightarrow50$ & $51\rightarrow95$  \\
				\texttt{can}         & $092\rightarrow\bm{094}$ & $35\rightarrow66$ & $00\rightarrow00$ & $56\rightarrow84$ & $71\rightarrow28$ & $00\rightarrow00$ & $81\rightarrow43$  \\
				\texttt{square}      & $084\rightarrow\bm{088}$ & $32\rightarrow71$ & $00\rightarrow00$ & $19\rightarrow51$ & $08\rightarrow08$ & $00\rightarrow00$ & $29\rightarrow31$  \\
				\bottomrule
		\end{tabular}}
	\label{tab:online_d_200}
\end{table*}

\clearpage

\begin{figure*}[t]
    \vskip -0.15in
	\includegraphics[width=\textwidth]{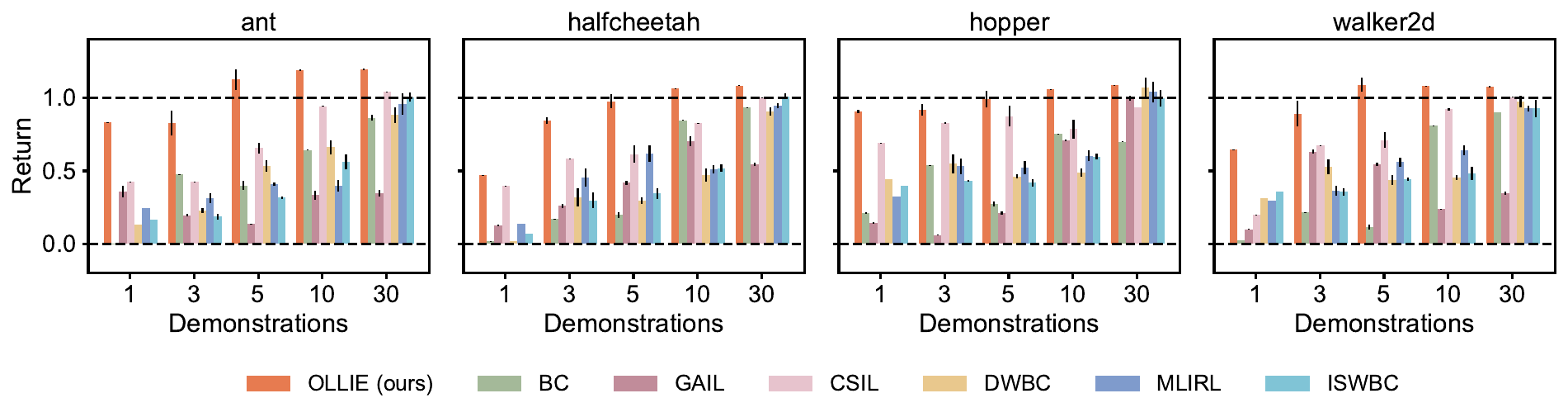}
	\vskip -0.15in
	\caption{End-to-end performance from offline pretraining to 10-episode finetuning under varying quantities of expert trajectories in \textbf{\textit{MuJoCo}}. Uncertainty intervals depict standard deviation over five seeds. The results demonstrate \texttt{OLLIE}'s \textit{overall} efficiency in both sampling/interaction and expert demonstrations, and underscore the great potential of pretraining and fintuning paradigm in IL.}
	\label{fig:online_performance_mujoco}
\end{figure*}

\begin{figure*}[t]
\centering
\vskip -0.25in
{\includegraphics[width=\textwidth]{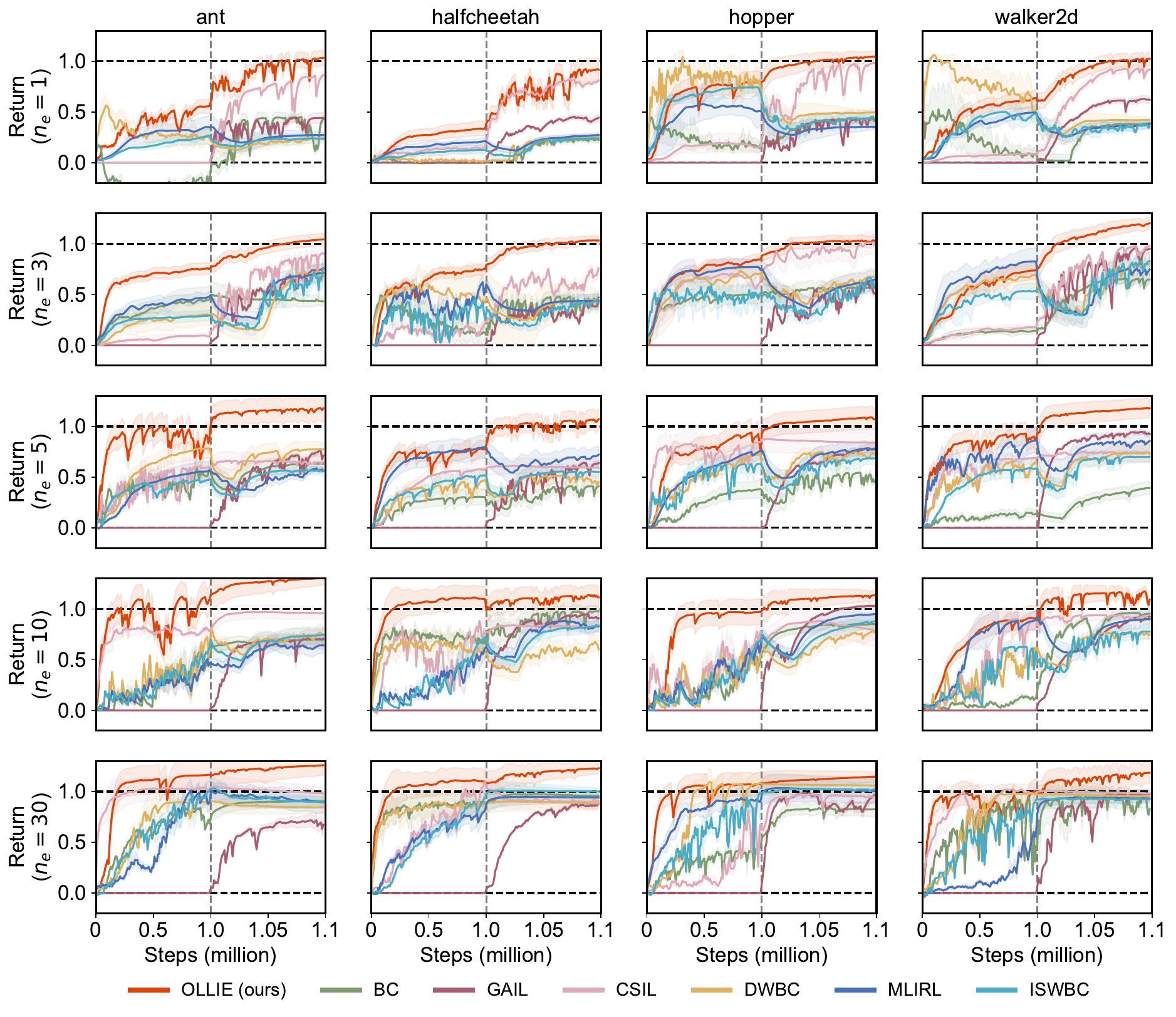}}
\vskip -0.2in
\caption{End-to-end learning curves from offline pretraining to online finetuning under varying quantities of expert trajectories in \textbf{\textit{MuJoCo}}. The dashed line separates the offline and online phases. Uncertainty intervals depict standard deviation over five seeds. $n_e$ represents expert trajectories. The simple combination of existing offline IL and \texttt{GAIL} suffers from unlearning pretrained knowledge. \texttt{OLLIE} not only avoids this issue but also expedites online training. This is attributed to \texttt{OLLIE}'s initial discriminator aligning with the well-performed policy initialization, thereby acting as a good local reward function capable of guiding fast policy search. In addition, while \texttt{CSIL} can alleviate the unlearning issue to some extent, its IRL procesure exhibits inefficiency compared to \texttt{OLLIE}.}
	\label{fig:online_curve_mujoco}
\end{figure*}

\clearpage
\begin{figure*}[t]
	\vskip -0.1in
	\centering
	{\includegraphics[width=\textwidth]{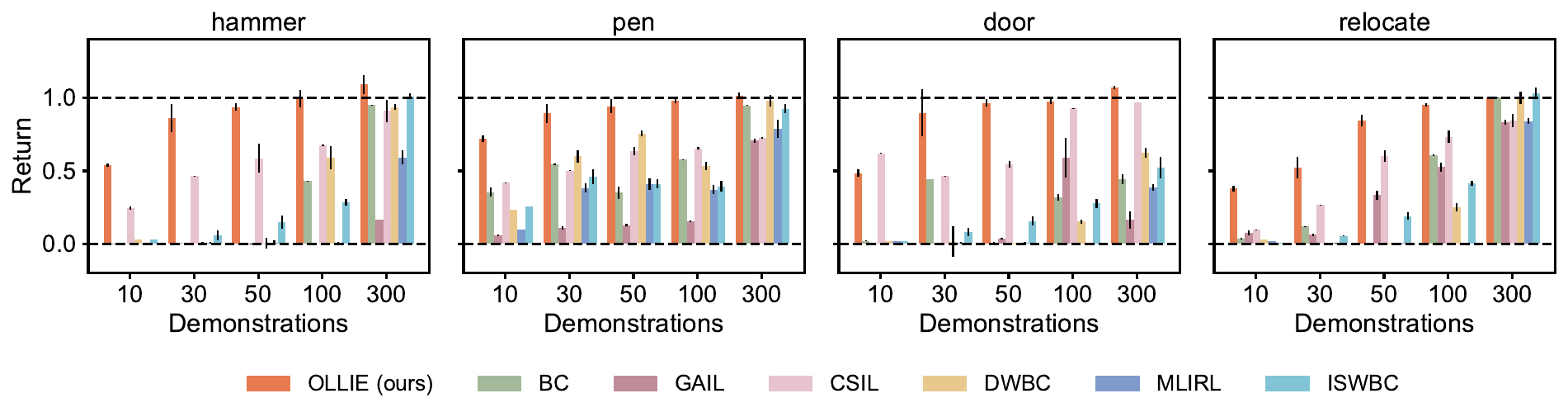}}
	\vskip -0.15in
	\caption{End-to-end performance from offline pretraining to 10-episode finetuning under varying quantities of expert trajectories in \textbf{\textit{Adroit}}. Uncertainty intervals depict standard deviation over five seeds. The results demonstrate \texttt{OLLIE}'s remarkable overall efficiency in both sampling/interaction and expert demonstrations and underscore the great potential of pretraining and fintuning paradigm in IL.}
	\label{fig:online_performance_adroit}
\end{figure*}

\begin{figure*}[t]
	\centering
	{\includegraphics[width=\textwidth]{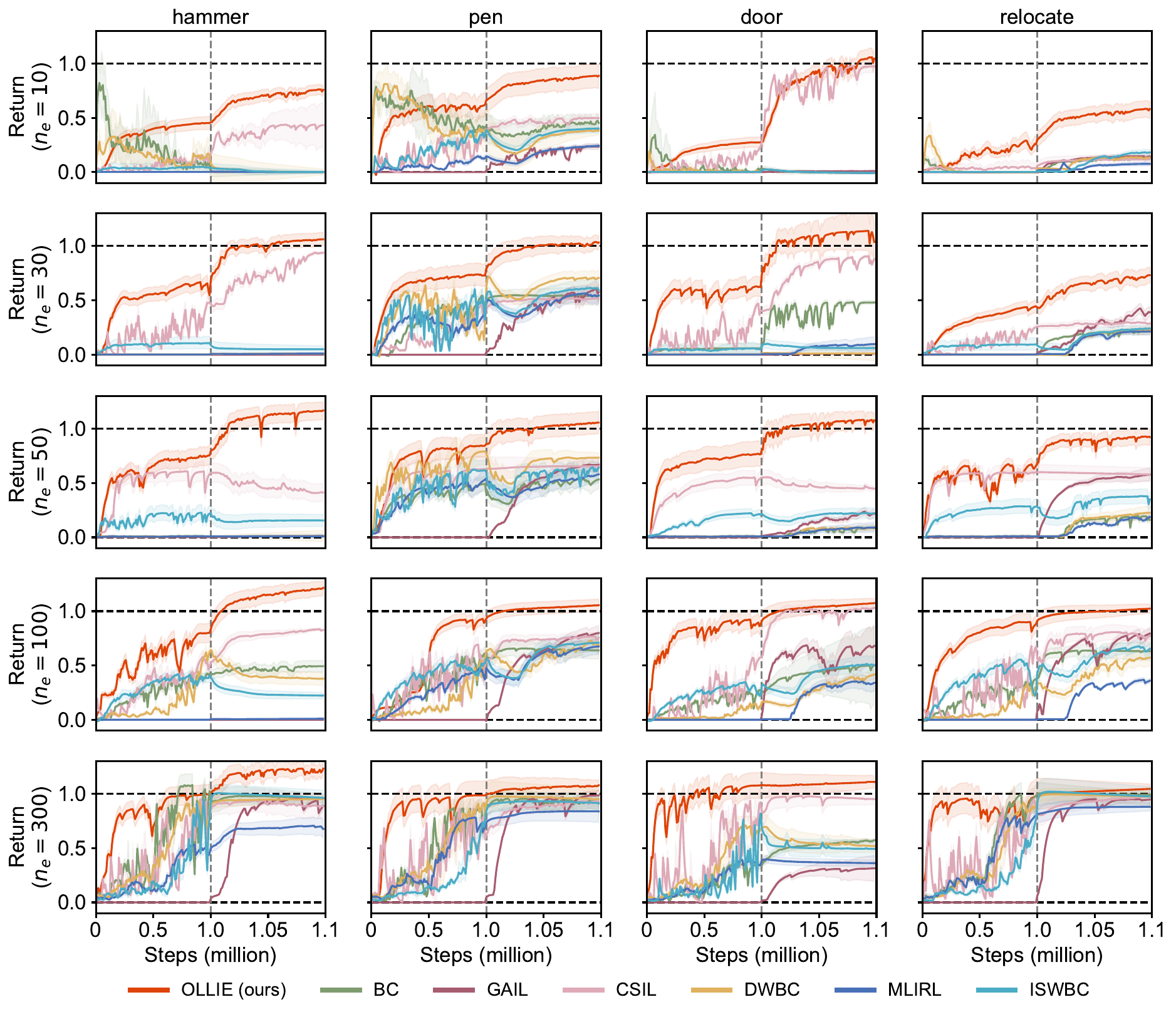}}
	\vskip -0.2in
	\caption{End-to-end learning curves from offline pretraining to online finetuning under varying quantities of expert trajectories in \textbf{\textit{Adroit}}. Uncertainty intervals depict standard deviation over five seeds. $n_e$ represents expert trajectories. Simple combination of existing offline IL and \texttt{GAIL} suffers from unlearning pretrained knowledge. \texttt{OLLIE} not only avoids this issue but also expedites online training. This is attributed to \texttt{OLLIE}'s initial discriminator aligning with the well-performed policy initialization, thereby acting as a good local reward function capable of guiding fast policy search. In comparison with MuJoCo, 
	high-dimensional environments make reward learning more challenging for IRL methods.}
	\label{fig:online_curve_adroit}
\end{figure*}

\clearpage
\begin{figure*}[t]
	\centering
	{\includegraphics[width=0.825\textwidth]{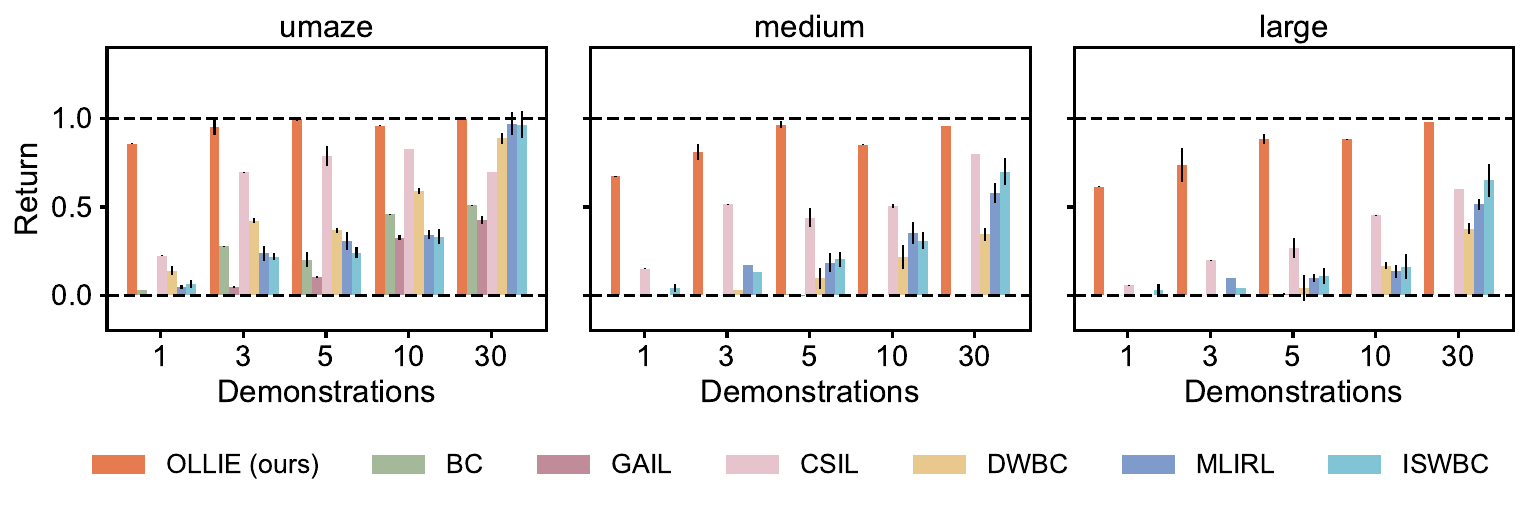}}
	\vskip -0.15in
	\caption{End-to-end performance from offline pretraining to 10-episode finetuning under varying quantities of expert trajectories in \textbf{\textit{AntMaze}}. Uncertainty intervals depict standard deviation over five seeds. It demonstrates the remarkable overall efficiency of \texttt{OLLIE} in both sampling/interaction and expert demonstrations and underscores great potentials of the pretraining and fintuning paradigm in IL.}
	\label{fig:online_performance_antmaze}
\end{figure*}

\clearpage
\begin{figure*}[t]
	\centering
	{\includegraphics[width=0.825\textwidth]{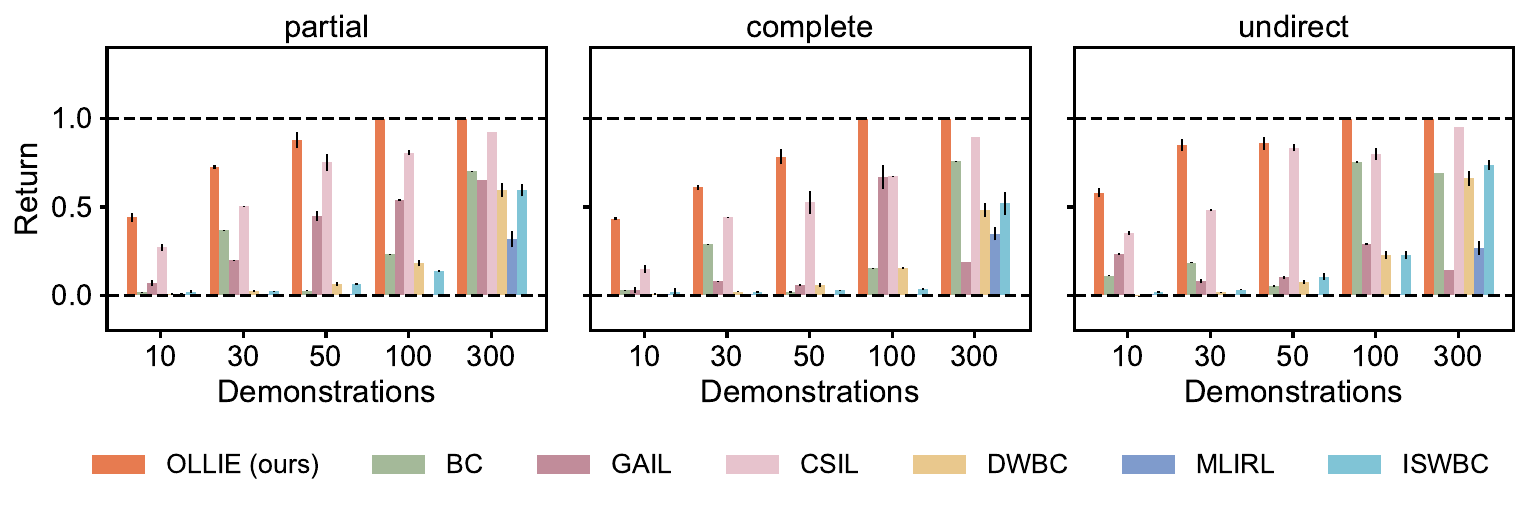}}
	\vskip -0.1in
	\caption{End-to-end performance from offline pretraining to 10-episode finetuning under varying quantities of expert trajectories in \textbf{\textit{FrankaKitchen}}. Uncertainty intervals depict standard deviation over five seeds. The results clearly demonstrate the remarkable overall efficiency of \texttt{OLLIE} in both sampling/interaction and expert demonstrations and underscore the great potential of pretraining and fintuning paradigm in IL.}
	\label{fig:online_performance_kitchen}
\end{figure*}

\begin{figure*}[t]
	\centering
        \vskip -0.1in
	{\includegraphics[width=0.85\textwidth]{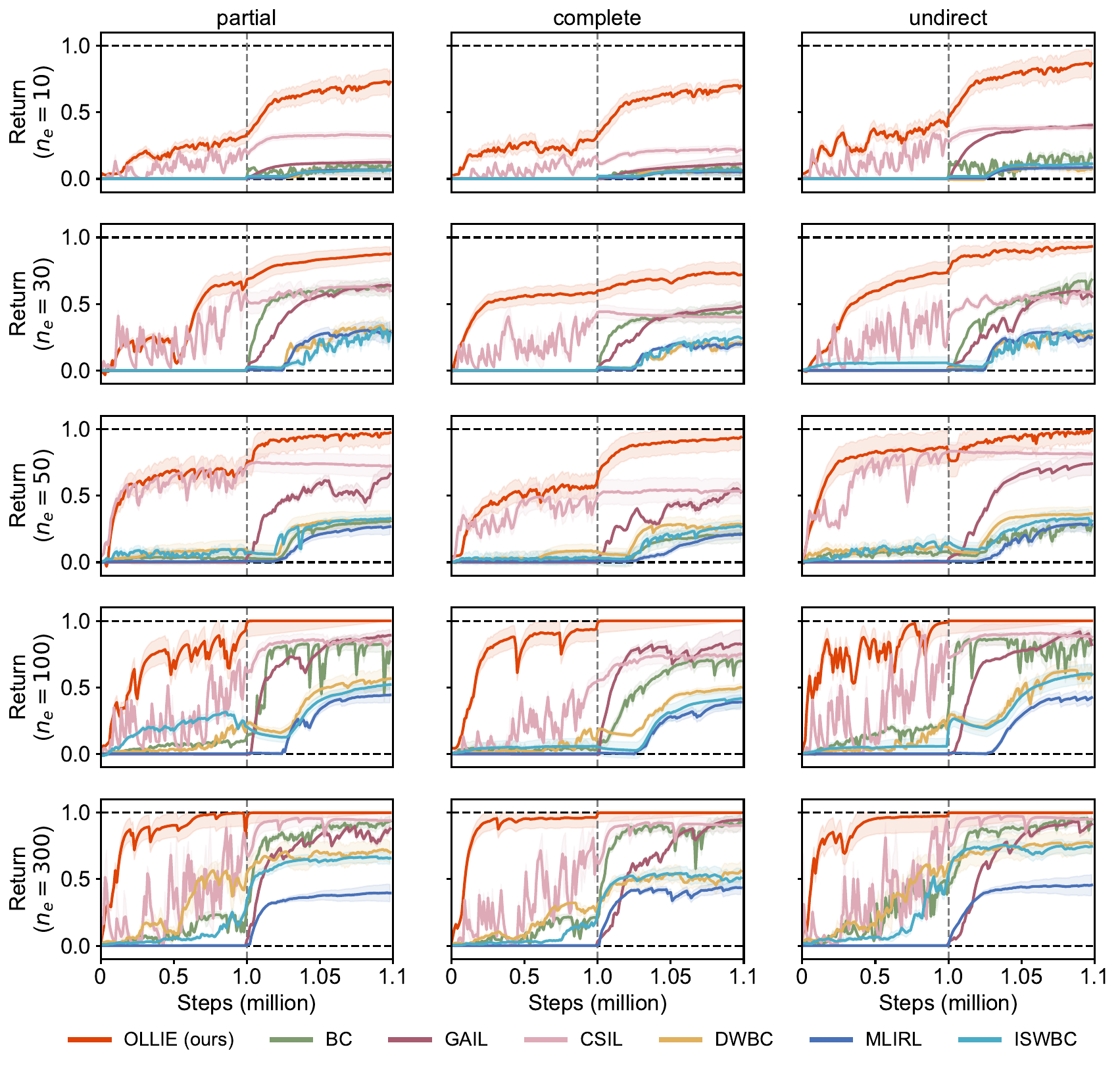}}
	\vskip -0.1in
	\caption{End-to-end learning curves from offline pretraining to online finetuning under varying quantities of expert trajectories in \textbf{\textit{FrankaKitchen}}. Uncertainty intervals depict standard deviation over five seeds. $n_e$ represents expert trajectories. The simple combination of existing offline IL and \texttt{GAIL} suffers from unlearning pretrained knowledge. \texttt{OLLIE} not only avoids this issue but also expedites online training. This is attributed to \texttt{OLLIE}'s initial discriminator aligning with the well-performed policy initialization, thereby acting as a good local reward function capable of guiding fast policy search. In comparison with MuJoCo, 
	high-dimensional environments make reward learning more challenging for IRL methods.}
	\label{fig:online_curve_kitchen}
\end{figure*}

\clearpage
\begin{figure*}[t]
	\centering
	{\includegraphics[width=\textwidth]{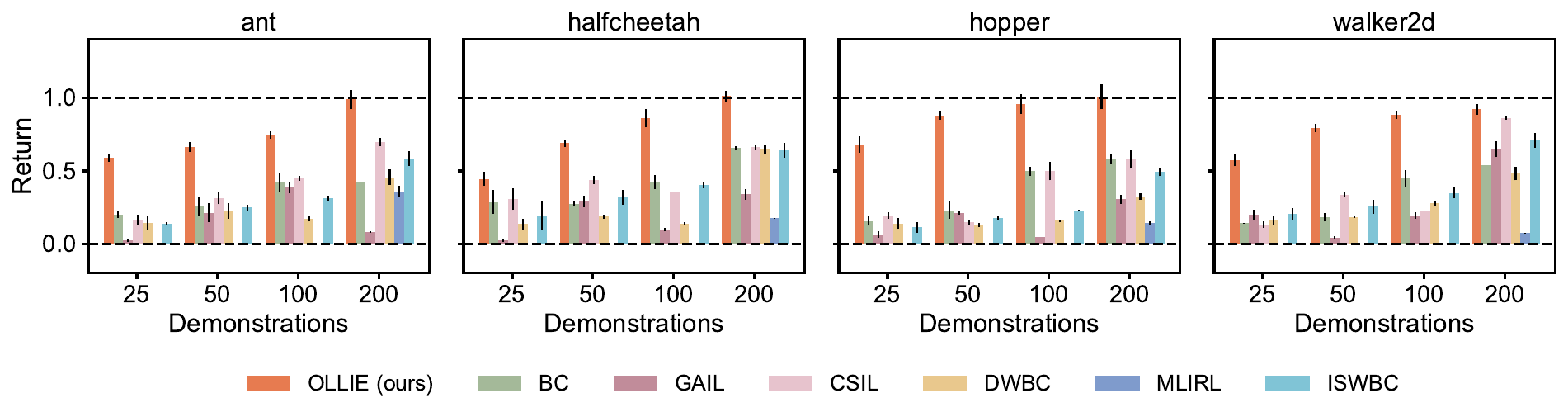}} 
	\vskip -0.1in
	\caption{End-to-end performance from offline pretraining to 10-episode finetuning under varying quantities of expert trajectories in \textbf{\textit{vision-based MuJoCo}}. Uncertainty intervals depict standard deviation over five seeds. The results clearly demonstrate the remarkable overall efficiency of \texttt{OLLIE} in both sampling/interaction in high-dimensional environments and expert demonstrations and underscore the great practical potential of pretraining and fintuning paradigm in IL.}
	\label{fig:online_performance_mujoco_image}
\end{figure*}

\begin{figure*}[t]
	\centering
	{\includegraphics[width=\textwidth]{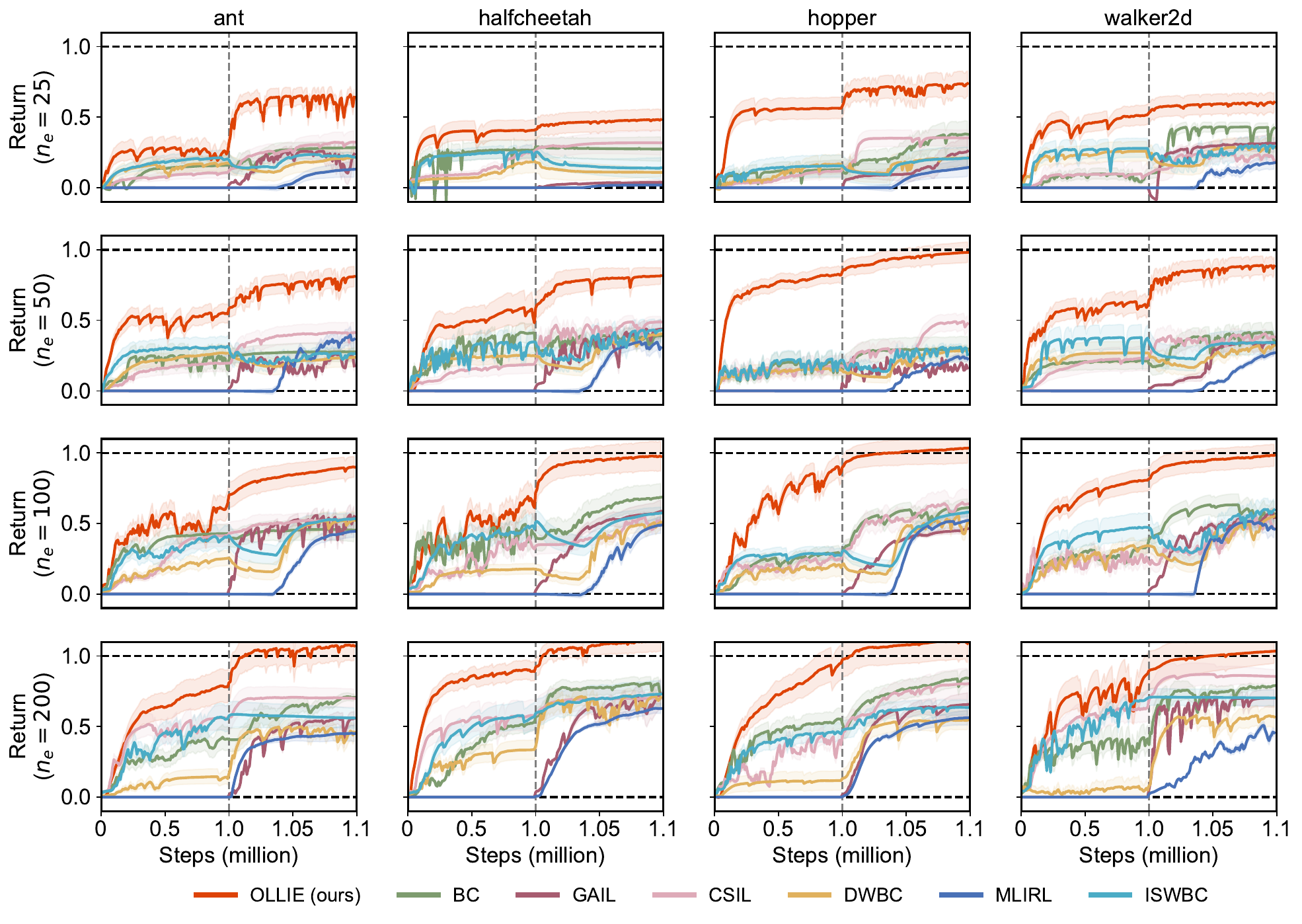}}
	\vskip -0.1in
	\caption{End-to-end learning curves from offline pretraining to online finetuning under varying quantities of expert trajectories in \textbf{\textit{vision-based MuJoCo}}. Uncertainty intervals depict standard deviation over five seeds. $n_e$ represents expert trajectories. The simple combination of existing offline IL and \texttt{GAIL} suffers from unlearning pretrained knowledge. \texttt{OLLIE} not only avoids this issue but also expedites online training. This is attributed to \texttt{OLLIE}'s initial discriminator aligning with the well-performed policy initialization, thereby acting as a good local reward function capable of guiding fast policy search. The superiority of \texttt{OLLIE} proves more remarkable in this vision-based domain, demonstrating its promise in practical scenarios.}
	\label{fig:online_curve_mujoco_image}
\end{figure*}

\clearpage
\begin{figure*}[t]
	\centering
	{\includegraphics[width=0.8\textwidth]{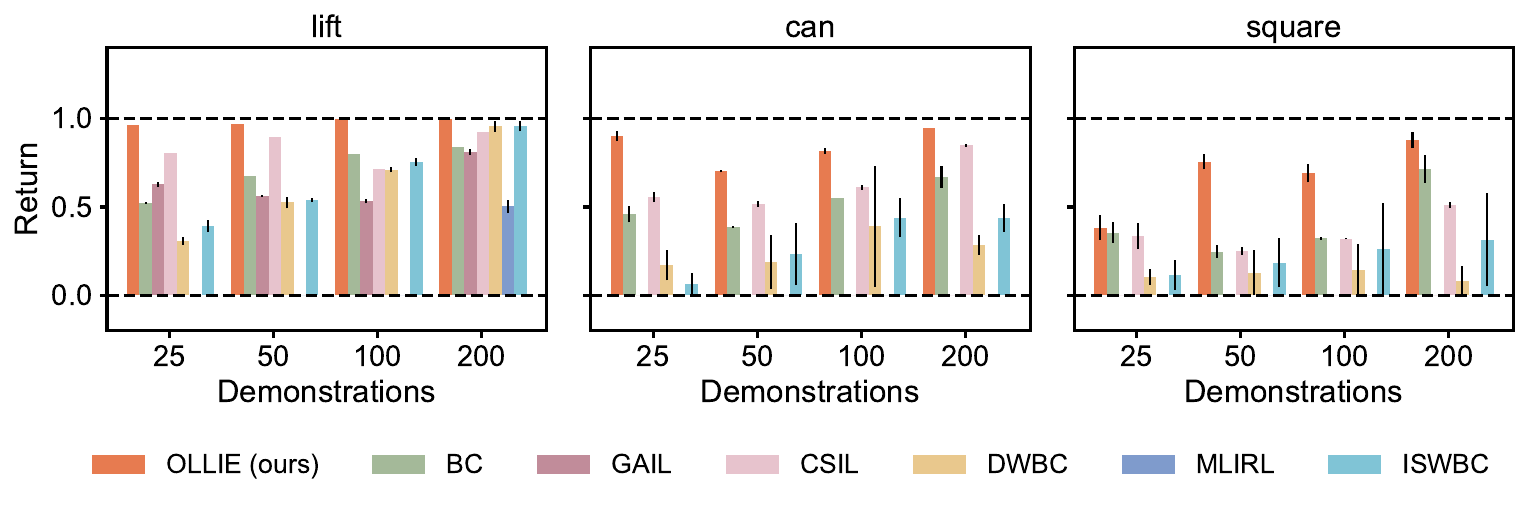}}
	\vskip -0.1in
	\caption{End-to-end performance from offline pretraining to 10-episode finetuning under varying quantities of expert trajectories in \textbf{\textit{vision-based Robomimic}}. Uncertainty intervals depict standard deviation over five seeds. The results demonstrate the remarkable overall efficiency of \texttt{OLLIE} in both sampling/interaction in high-dimensional environments and expert demonstrations and underscore the great practical potential of pretraining and fintuning paradigm in IL.}
	\label{fig:online_performance_robomimic}
\end{figure*}

\clearpage
\subsection{Ablation Studies and Complementary Experiments}
\label{sec:ablation}

This section includes ablation studies for \texttt{OLLIE} and complementary empirical studies.

\subsubsection{Importance of discriminator initialization}

\begin{figure*}[ht]
	\centering
	{\includegraphics[width=\textwidth]{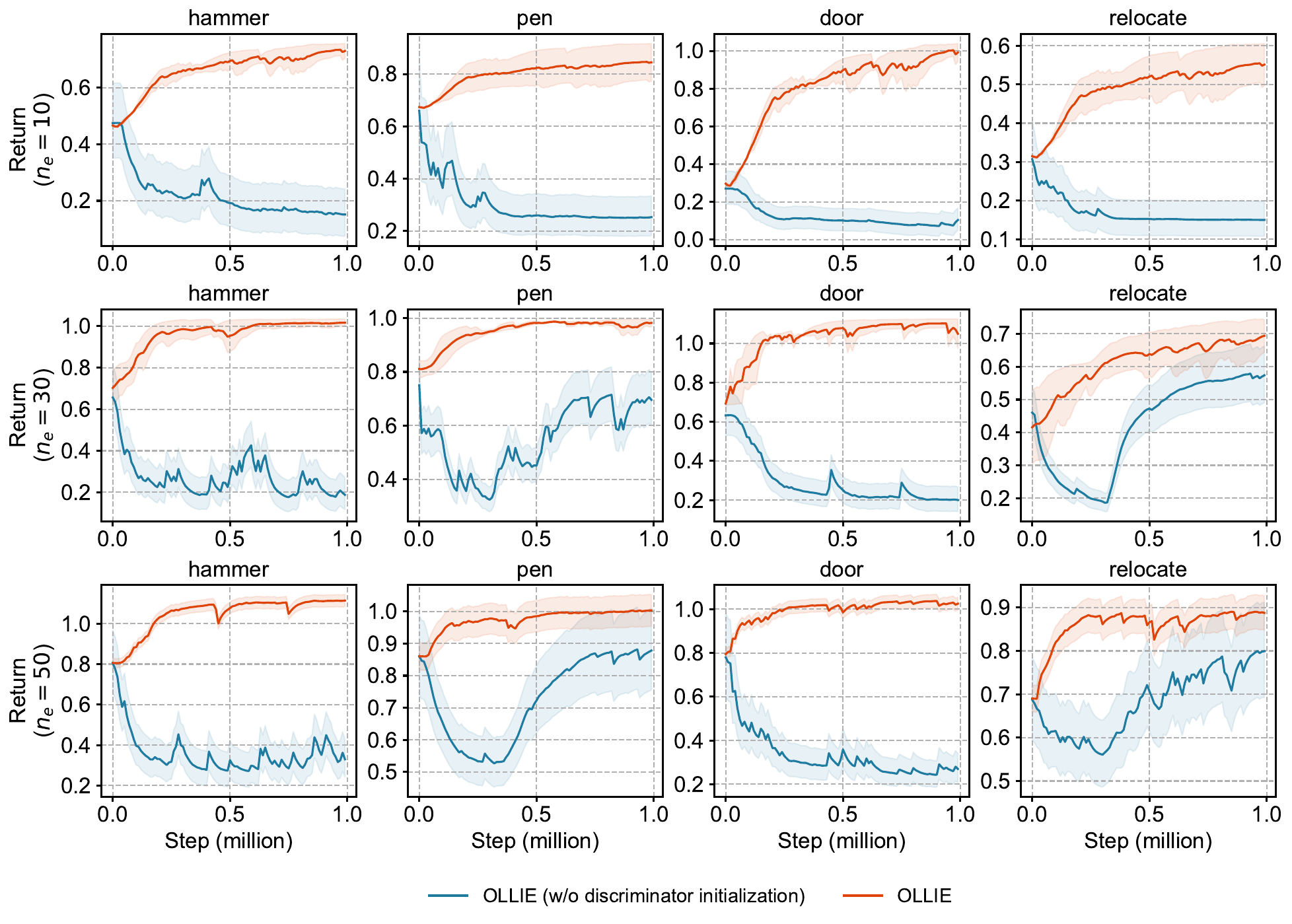}}
	\caption{Ablation on the discriminator initialization. Uncertainty intervals depict standard deviation over five seeds. Albeit with a good policy initialization, the finetuning performance of \texttt{OLLIE} experiences a significant degradation in the absence of the initialized discriminator.}
	\label{fig:non_discriminator}
\end{figure*}

We ablate the discriminator initialization by a random initialization and run experiments with the same setup as \cref{tab:online_d_1,tab:online_d_10,tab:online_d_25}. Not surprisingly, albeit with a good policy initialization, the finetuning performance degrades dramatically in the absence of the initialized discriminator. In addition, \cref{fig:non_discriminator} demonstrates that pretraining with less expert demonstrations leads to a worse finetuning performance. This could be attributed to the poor generalizability of the policy trained with less data (akin to the findings of \citet{he2019rethinking} in vision).


\clearpage
\subsubsection{Importance of Imperfect Demonstrations}

\begin{figure*}[ht]
	\centering
	{\includegraphics[width=\textwidth]{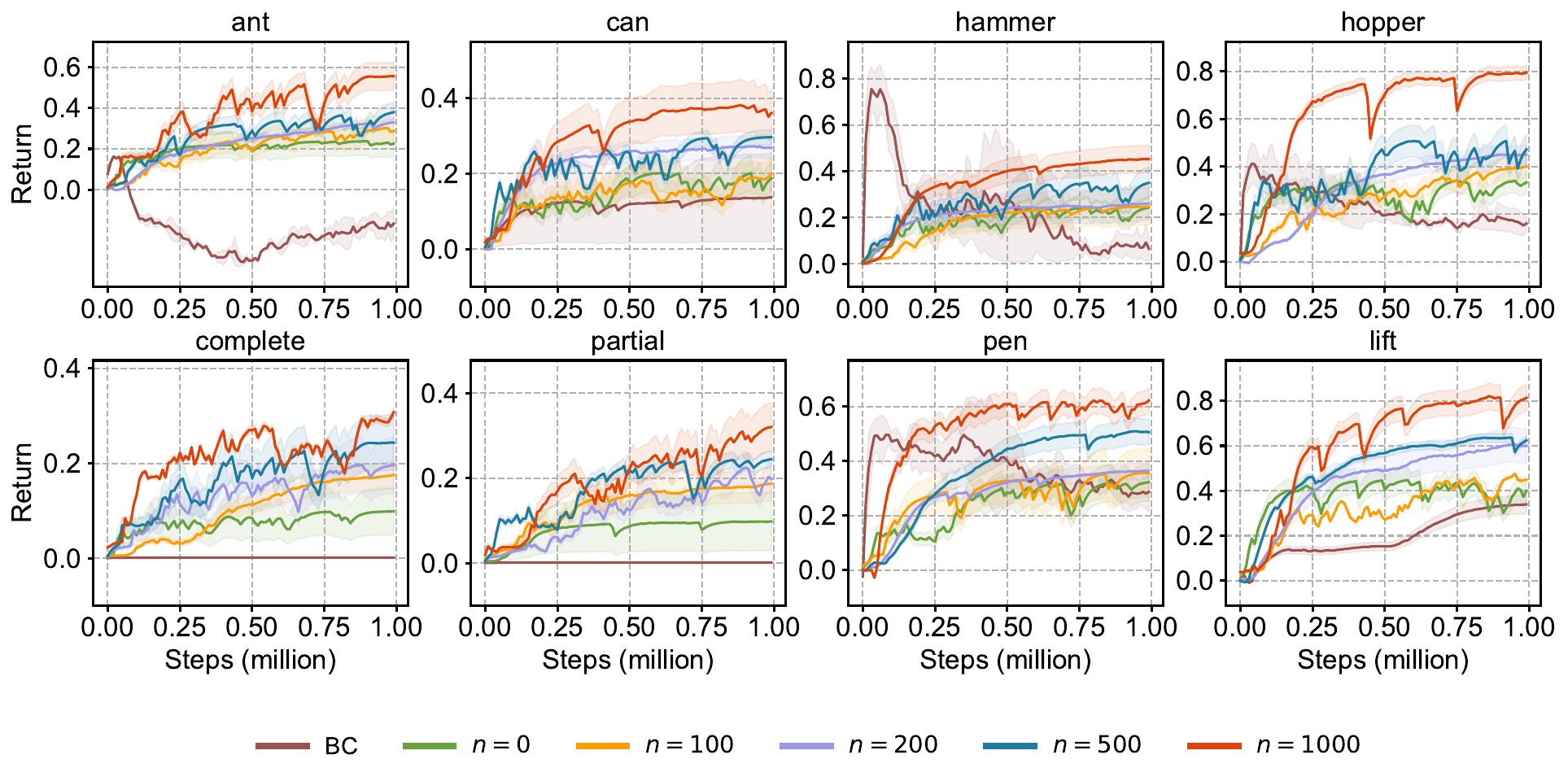}}
	\caption{Importance of complementary suboptimal data. Uncertainty intervals depict standard deviation over five seeds. Even with no complementary data, \texttt{OLLIE} outperforms \texttt{BC}, owing to the capability of utilizing dynamics information in expert demonstrations.}
	\label{fig:unlabel_data}
\end{figure*}

To assess the effect of imperfect demonstrations, we carry out experiments by varying the number of complementary trajectories from 0 to 1000. The setup follows that of \cref{tab:online_d_1,tab:online_d_10,tab:online_d_25}. As illustrated in \cref{fig:unlabel_data}, it is important to incorporate complementary data, which can remedy the limited state coverage of expert demonstrations and combat covariate shifts. Moreover, with no complementary data, \texttt{DWBC}, \texttt{ISWBC}, and \texttt{CSIL} reduce to \texttt{BC} or soft \texttt{BC}, which proves ineffective with scarce expert data; and the model-based counterparts struggle in these high-dimensional environments. In contrast, \texttt{OLLIE} works much better, as it can effectively leverage the dynamics information in expert demonstrations and enable the policy to stay in the expert data support as much as possible.

\clearpage
\subsubsection{A Visualization of Distribution Matching}
\label{sec:distribution_matching}

\begin{figure}[ht]
\centering
\subfigure[Expert data]{
\label{subfig:antmaze_expert}
\includegraphics[width=0.18\textwidth]{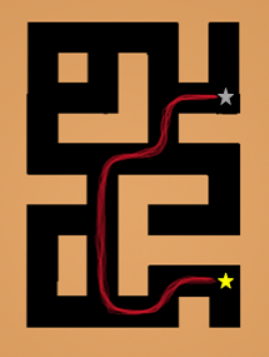}}
\subfigure[Imperfect data]{
\label{subfig:antmaze_diverse}
\includegraphics[width=0.18\textwidth]{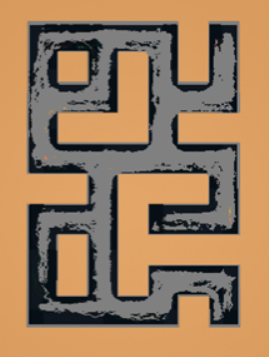}}
\subfigure[\texttt{DWBC}]{
\label{subfig:antmaze_dwbc}
\includegraphics[width=0.18\textwidth]{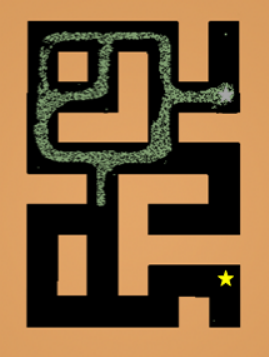}}
\subfigure[\texttt{ISWBC}]{
\label{subfig:iswbc}
\includegraphics[width=0.18\textwidth]{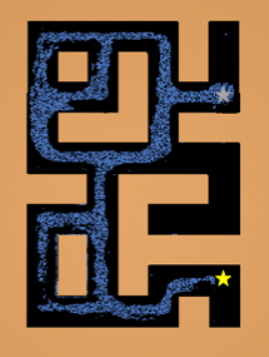}}
\subfigure[\texttt{OLLIE}]{
\label{subfig:antmaze_ollie}
\includegraphics[width=0.18\textwidth]{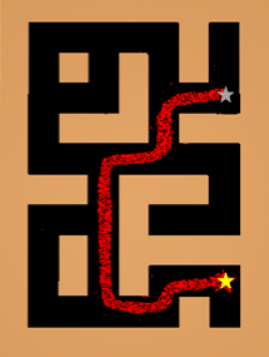}}
\caption{Visualization of the trajectories in \texttt{antmaze-large} generated by different algorithms. The policies are trained with 10 \texttt{expert} trajectory (shown in (a)) as the expert demonstration and 1000 \texttt{diverse} trajectories (shown in (b)) as imperfect demonstrations. (c)-(d) depict the trajectories sampled from the policies learned by \texttt{DWBC}, \texttt{ISWBC}, and \texttt{OLLIE}, repectively.}
\label{fig:antmaze}
\end{figure}

To validate the theory of \texttt{OLLIE} where it minimizes the discrepancy with the empirical expert distribution, we visualize the experimental result of \texttt{antmaze-large} in \cref{fig:antmaze}, where \texttt{OLLIE} nearly recovers the expert state-action distribution. From Eq.~(5) of \citet{xu2022discriminator}, \texttt{DWBC} assigns positive weights to all diverse state-actions, leading to the interference in mimicking expert behaviors. As analyzed in \cref{sec:supp_related_work}, with scarce expert data, \texttt{ISWBC} may pursue a biased objective and suffer from error compounding once leaving the expert data support.

\clearpage
\subsubsection{Minimax Optimization}
\label{sec:minimax_loss}

\begin{figure*}[ht]
	\centering
	{\includegraphics[width=\textwidth]{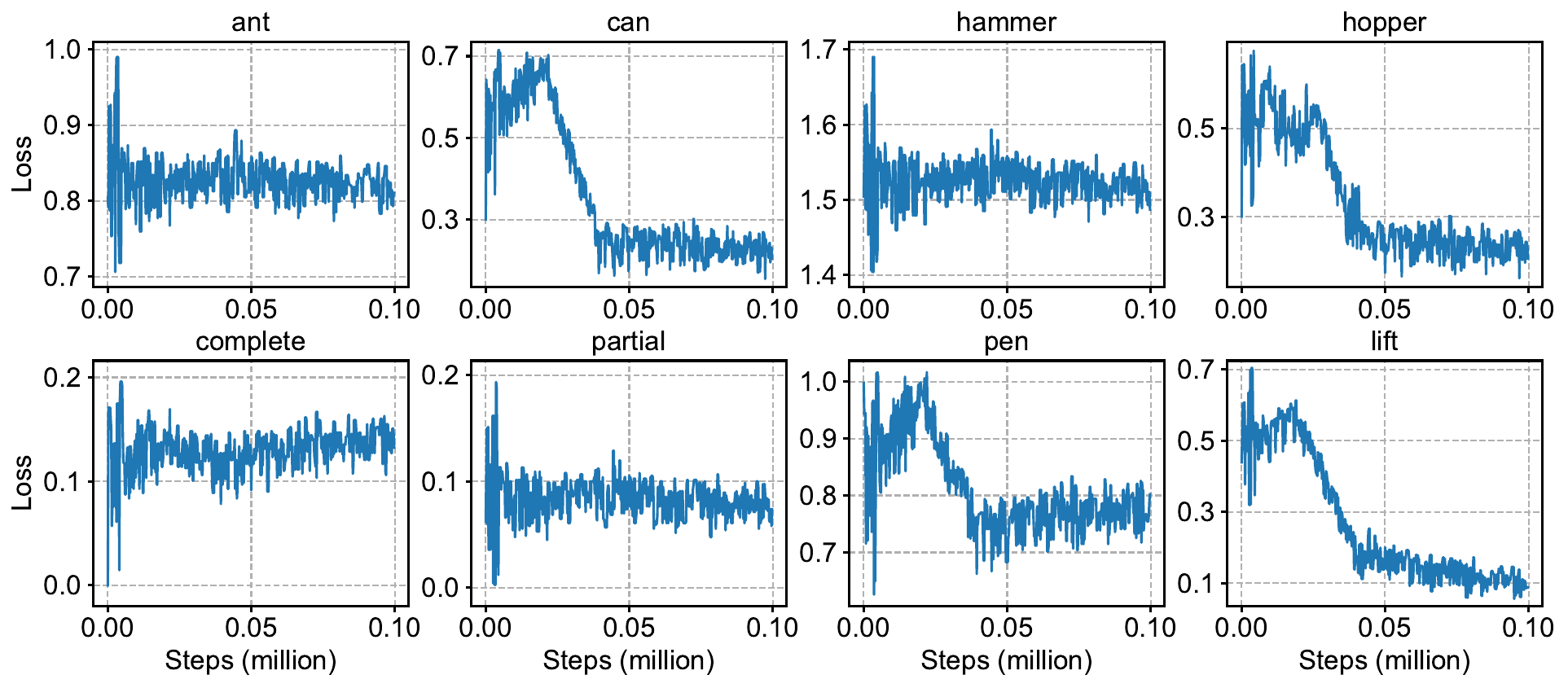}}
	\caption{The value of $\tilde F(\phi_\nu,\phi_y)$ in selected experiments.}
	\label{fig:minimax_loss}
\end{figure*}

We test the stability of the approximate dual descent in solving the SSP. As shown in \cref{fig:minimax_loss}, it works well in all environments and often converges in 50k gradient steps.

\clearpage
\subsubsection{Further Comparison with \texttt{MLIRL}}
\label{sec:comparision_mlirl}

\begin{figure*}[ht]
	\centering
	{\includegraphics[width=\textwidth]{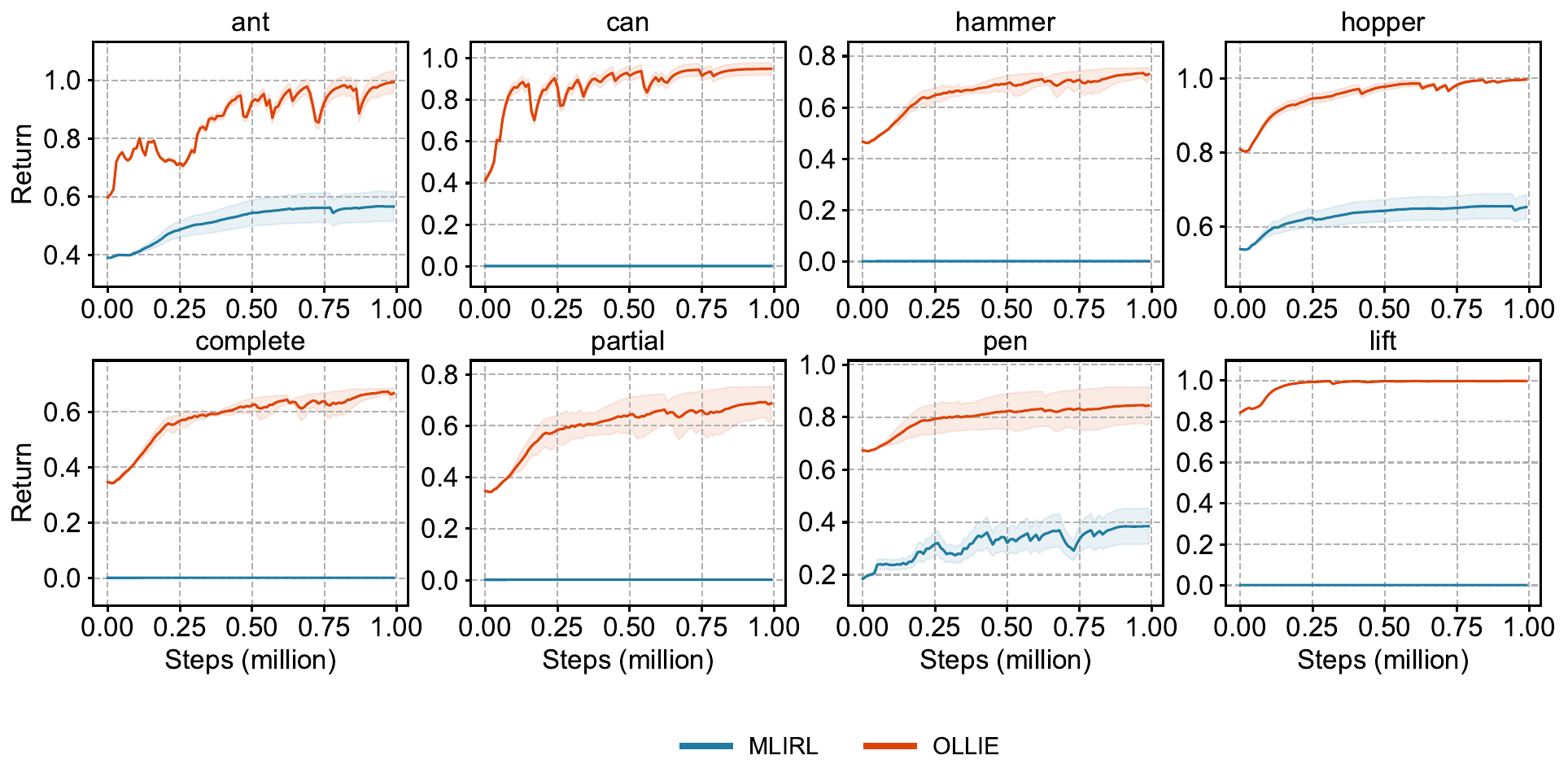}}
	\caption{Performance of finetuning \texttt{MLIRL} via its learned reward function. Uncertainty intervals depict standard deviation over five seeds.}
    \label{fig:recoverd_reward}
\end{figure*}

In our experiments, we use \texttt{GAIL} to tune the policy learned by \texttt{MLIRL}. Since \texttt{MLIRL} learns a reward function during the offline phase, another approach to tune \texttt{MLIRL} is employing forward RL with the learning reward function. Unfortunately, we find it performs highly suboptimally, as demonstrated in \cref{fig:recoverd_reward}. This is because the reward extrapolation error leads to spurious rewards in out-of-distribution environments. Of note, despite being an IRL method, \texttt{CSIL} can alleviate this issue to some extent by further refining the reward function during online fine-tuning.

\clearpage
\subsubsection{Undiscounted Experiments} 

\begin{figure*}[ht]
    \centering
    {\includegraphics[width=\textwidth]{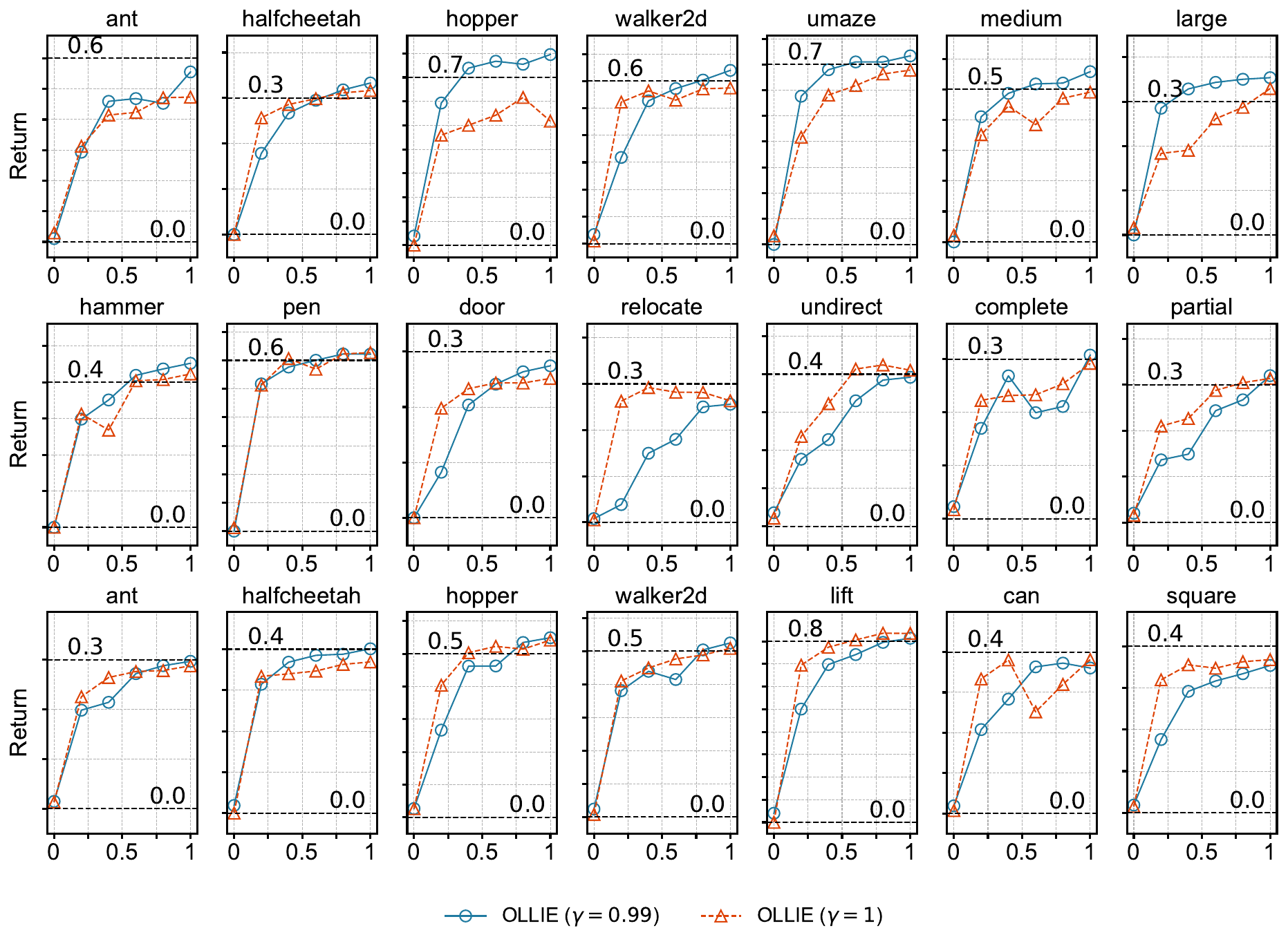}}
    \caption{\centering Performance of discounted and discounted OLLIE in offline IL. The results are averaged over 5 random seeds. The last row comprises vision-based tasks.}
    \label{fig:non_discount_1}
\end{figure*}

As discussed in \cref{sec:undiscounted}, \texttt{OLLIE} can extend to undiscounted problems ($\gamma = 1$). In \cref{fig:non_discount_1}, we compare between discounted and undiscounted \texttt{OLLIE}. There is not a significant discrepancy between them across employed benchmarks. 

\clearpage
\subsubsection{Forward and Reverse Policy Extraction}
\label{sec:comparison_forward_reverse}

We examine the performance of the \textit{forward} and \textit{reverse} policy extraction methods introduced in \cref{sec:policy_extraction}. The setup remains the same as \cref{tab:online_d_1,tab:online_d_10,tab:online_d_25} with limited expert demonstrations and low-quality complementary data. They both work well in the benchmarks, of which the final performance is environment-dependent. We find the forward method enjoys better convergence speed and stability in most environments.

\begin{figure*}[ht]
	\centering
	{\includegraphics[width=\textwidth]{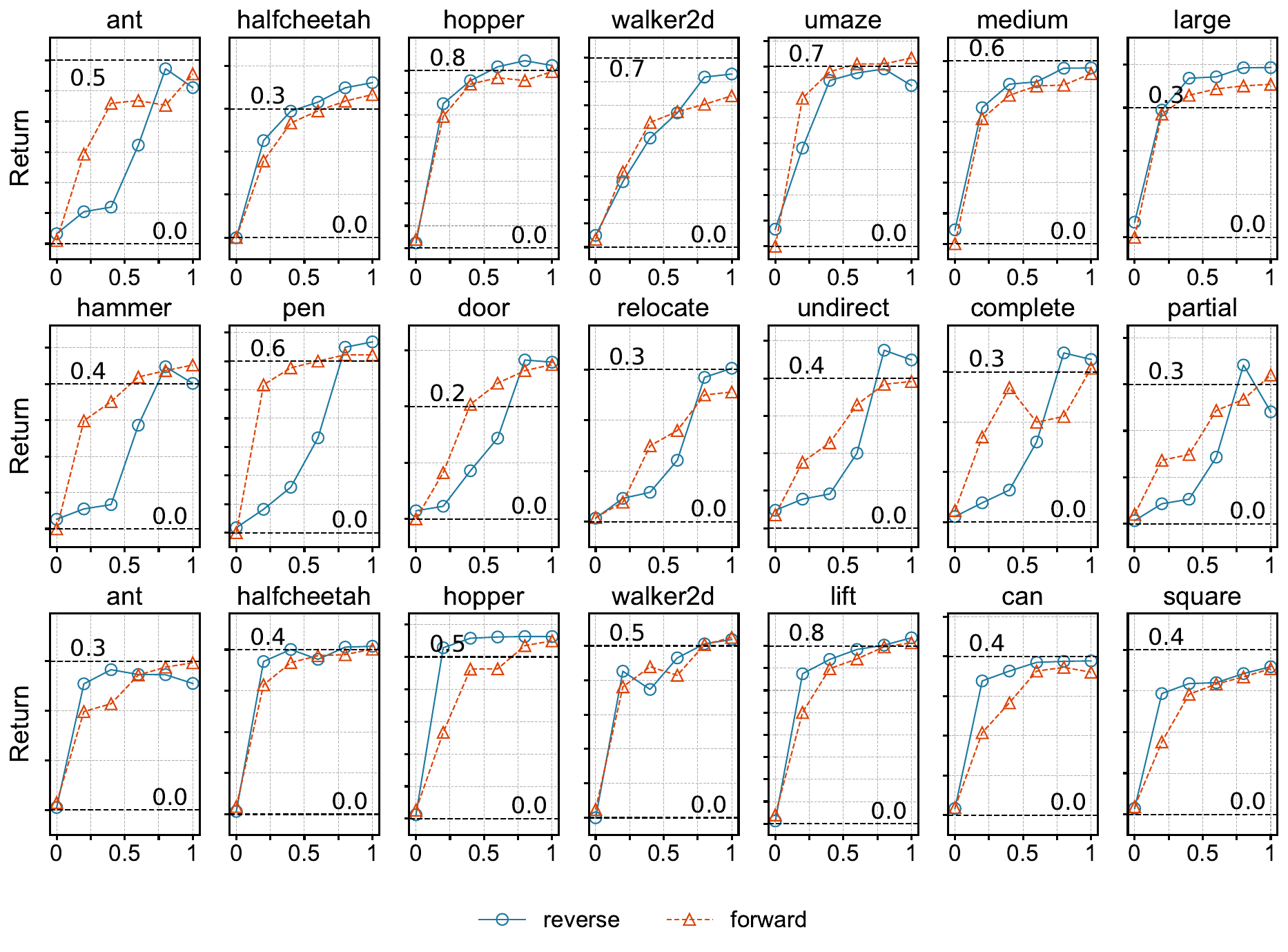}}
	\caption{Comparison between the forward and reverse policy extraction. Uncertainty intervals depict standard deviation over five seeds. The forward method enjoys better convergence speed and stability in most environments. Their final performance is environment-dependent. }
	\label{fig:reverse_forward}
\end{figure*}

\clearpage
\subsubsection{Discriminator Alignment}
\label{sec:motivating_full}

\begin{figure*}[!ht] 
 \centering
	{\includegraphics[width=\textwidth]{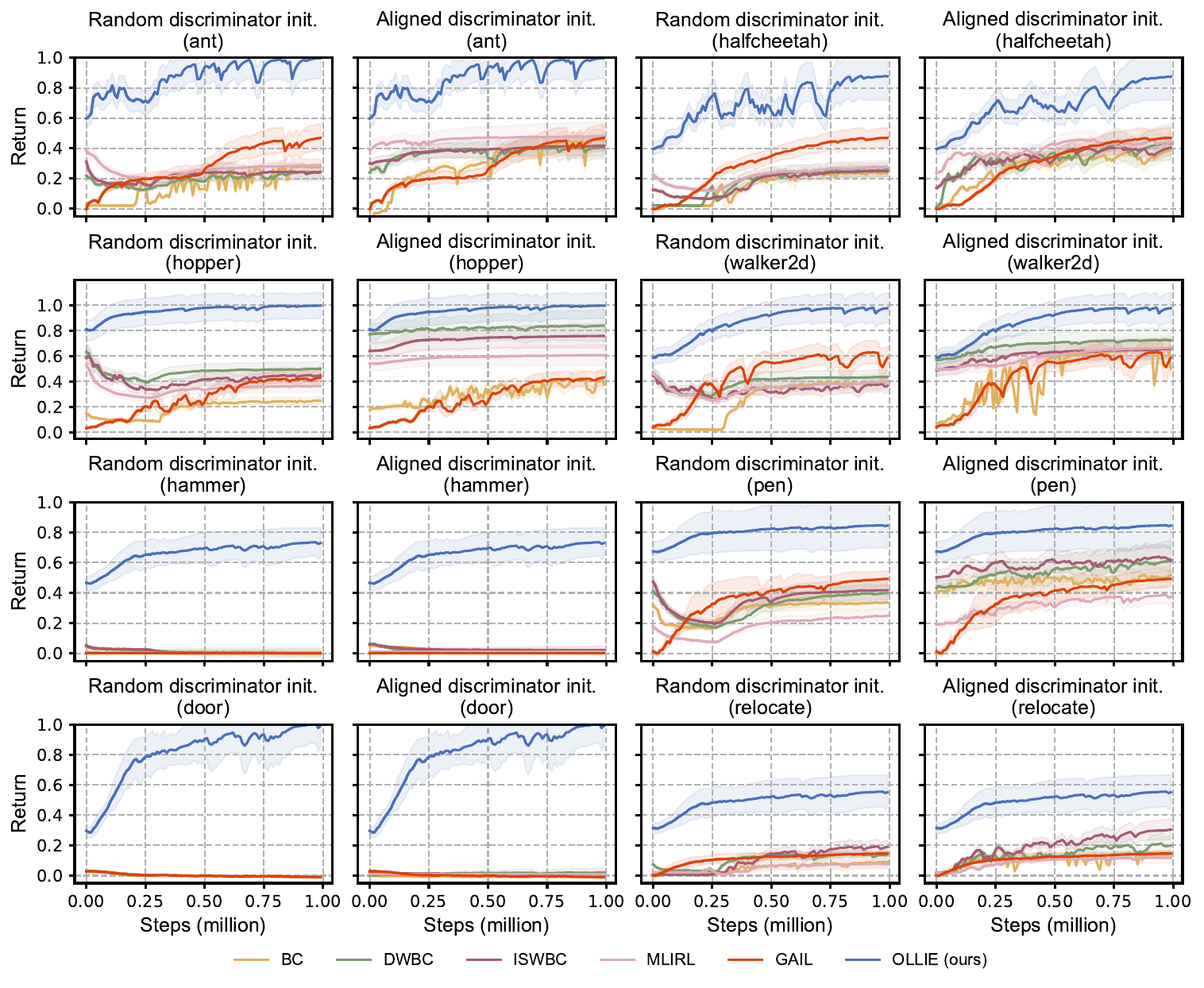}}
 \caption{The effect of discriminator alignment. Uncertainty intervals depict standard deviation over five random seeds. In each task, the left figure depicts the performance of finetuning the policies learned by existing methods using \texttt{GAIL} with a random discriminator initialization. The right figure shows the performance after aligning the discriminator using 100 trajectories generated by the initialized policies. In MuJoCo, the policies are pretrained with 1 \texttt{expert} trajectory and 1000 \texttt{random} trajectories. In Adroit, the policies are pretrained with 10 \texttt{expert} trajectory and 1000 \texttt{cloned} trajectories. }
 \label{fig:motivation_all}
\end{figure*}




\end{document}